\documentclass[twoside]{article}
\usepackage{ecj,palatino,epsfig,latexsym,natbib}

\usepackage{amssymb}
\usepackage{amsmath, amsthm}
\usepackage[ruled,vlined]{algorithm2e}
\usepackage{stmaryrd}
\newcommand{\argmax}{\mathop{\rm arg~max}\limits}

\newcommand{\R}{\mathbb{R}}
\newcommand{\T}{\mathrm{T}}
\newcommand{\N}{\mathrm{N}}
\newcommand{\E}{\mathbb{E}}

\theoremstyle{definition}
\newtheorem{theorem}{Theorem}
\newtheorem{lemma}{Lemma}
\newtheorem{corollary}{Corollary}

\newcommand{\KVAL}{f_{\,\textsc{kv}}}
\newcommand{\ONE}{f_{\,\textsc{com}}}

\newcommand*{\target}[1][true]{\ifthenelse{\boolean{#1}}{ccGA\ }{ccGA}}

\newcommand{\event}[1][t]{E^{(#1)}}
\newcommand{\bevent}[1][t]{\bar{E}^{(#1)}}

\newcommand{\param}[1][t]{\theta^{(#1)}}
\newcommand{\sample}[1][1]{x^{\langle #1 \rangle}}

\newcommand{\F}[1][t]{\mathcal{F}^{(#1)}}
\newcommand{\X}[1][t]{X^{(#1)}}
\newcommand{\tX}[1][t]{\tilde{X}^{(#1)}}
\newcommand{\Y}[1][t]{Y^{(#1)}}
\newcommand{\indic}[1][E^{(t)}]{\mathbf{1}_{\{{#1}\}}}
\newcommand{\intrange}[2]{{\llbracket  #1, #2 \rrbracket}}

\newcommand{\cone}[1][1]{c_{#1}}

\newcommand{\ckv}[1][1]{c_{#1}}

\usepackage{xcolor}
\usepackage{ifthen}
\newcommand{\markupdraft}[2]{
    \ifthenelse{\equal{#1}{display}}{#2}{}
    \ifthenelse{\equal{#1}{color}}{\color{#2}}{}
}
\newcommand{\notecolored}[3][]{\markupdraft{display}{{\color{#2}\noindent[Note (#1): #3]}}}
\newcommand{\newcolored}[3][]{{\markupdraft{color}{#2}#3}
    \ifthenelse{\equal{#1}{}}{}{\markupdraft{display}{{\color{yellow!70!black}[#1]}}}}
\newcommand{\del}[2][]{{\markupdraft{display}{{\color{orange}[removed: ``#2''[#1]]}}}} 
\newcommand{\new}[2][]{\newcolored[#1]{blue}{#2}}
\newcommand{\note}[2][]{\notecolored[#1]{green}{#2}}
\newcommand{\calc}[2][]{{\markupdraft{display}{{\color{purple} Derived as follows: {#2}[#1]}}}} 

\renewcommand{\del}[2]{}  
\renewcommand{\calc}[2]{}  
\renewcommand{\markupdraft}[2]{}  
\newcommand{\yohe}[1]{\note[Youhei]{\color{cyan} #1}}

\newcommand{\kento}[1]{\note[Kento]{\color{brown} #1}}

\renewcommand{\paragraph}[1]{\textbf{#1}\ }

\usepackage{amsmath}

\usepackage[colorlinks=false,pdfborder={0 0 0}]{hyperref}
\usepackage{cleveref}

\usepackage{mathtools}
\newcommand{\rev}[1]{\textcolor{red}{#1}}
\newcommand{\revdelsec}[1]{\textcolor{orange}{[removed: #1]}}
\newcommand{\revdel}[1]{\mathtoolsset{showonlyrefs=true}\textcolor{orange}{[removed: #1]}\mathtoolsset{showonlyrefs=false}}
\newcommand{\rrev}[1]{\textcolor{blue}{#1}}
\newcommand{\rrevdel}[1]{\mathtoolsset{showonlyrefs=true}\textcolor{pink}{[removed: #1]}\mathtoolsset{showonlyrefs=false}}
\newcommand{\rrrev}[1]{\textcolor{red}{#1}}
\renewcommand{\rev}[1]{#1}
\renewcommand{\revdelsec}[1]{}
\renewcommand{\revdel}[1]{}
\renewcommand{\rrevdel}[1]{}
\renewcommand{\rrev}[1]{#1}
\renewcommand{\rrrev}[1]{#1}

\begin{document}

\ecjHeader{x}{x}{xxx-xxx}{202X}{
Tail Bound\rev{s} on \rrev{the} Runtime of ccGA
}{R. Hamano et al.}
\title{\bf Tail Bound\rev{s} on \rrev{the} Runtime of Categorical Compact Genetic Algorithm}  

\author{\name{\bf Ryoki Hamano} \hfill \addr{hamano\_ryoki\_xa@cyberagent.co.jp}\\
\addr{CyberAgent, Inc., Tokyo, Japan}
\AND
        \name{\bf Kento Uchida} \hfill \addr{uchida-kento-fz@ynu.ac.jp}\\
        \name{\bf Shinichi Shirakawa} \hfill \addr{shirakawa-shinichi-bg@ynu.ac.jp}\\
        \addr{Yokohama National University, Kanagawa, Japan}
\AND
        \name{\bf Daiki Morinaga} \hfill \addr{morinaga@bbo.cs.tsukuba.ac.jp}\\
       \name{\bf Youhei Akimoto} \hfill \addr{akimoto@cs.tsukuba.ac.jp}\\
        \addr{University of Tsukuba, Ibaraki, Japan}\\
        \addr{RIKEN Center for Advanced Intelligence Project, Tokyo, Japan}
}

\maketitle

\begin{abstract}

\rrevdel{Major}\rrev{The majority of} theoretical analyses of evolutionary algorithms in the discrete domain focus on binary optimization algorithms, even though black-box optimization on the categorical domain has a lot of practical applications. In this paper, we consider a probabilistic model-based algorithm using the family of categorical distributions as its underlying distribution and set the sample size as two. We term this specific algorithm the categorical compact genetic algorithm (ccGA). The ccGA can be considered as an extension of the compact genetic algorithm (cGA)\revdel{ that}{}\rev{, which} is an efficient binary optimization algorithm. We theoretically analyze the dependency of the number of possible categories $K$, the number of dimensions $D$, and the learning rate $\eta$ \revdel{to}{}\rev{on} the runtime. We investigate the tail bound of the runtime on two typical linear functions on the categorical domain\revdel{, the}{}\rev{:} categorical OneMax (COM) and \textsc{KVal}. We derive that the runtimes on COM and \textsc{KVal} are $O(\sqrt{D} \ln (D K) / \eta)$ and $\Theta(D \ln K/ \eta)$ with high probability, respectively.
Our analysis is a generalization for that of the cGA on the binary domain. 

\end{abstract}


\section{Introduction} \label{sec:intro}
\rrevdel{Real-world applications often include discrete domain black-box optimization.}\rrev{Discrete domain black-box optimization is often involved in real-world applications.} Several \rev{discrete} black-box optimization methods are developed to obtain the optimal solution with \revdel{fewer}\rev{few} \rrevdel{calls}\rrev{evaluations} of the objective function. 
Among these methods, some well-performing methods are categorized as probabilistic model-based algorithms.
The probabilistic model-based algorithms repeat generating multiple samples from the underlying probability distribution and updating the distribution parameter\rev{s}.
Particularly, the compact genetic algorithm (cGA)~\citep{CGA:1999} has been widely used as a probabilistic model-based algorithm for \rrevdel{the }binary optimization and has shown efficient search performance in several situations~\citep{cGA:application:1, cGA:application:2, cGA:application:3}.

Information geometric optimization (IGO)~\citep{IGO:2017} has been proposed as a unified framework of probabilistic model-based algorithms to reveal the key factor of the well-known probabilistic model-based algorithms.
According to the IGO framework, cGA can be explained as a stochastic natural gradient ascent \revdel{for}{}\rev{with respect to (w.r.t.)} the parameter of \rev{the }Bernoulli distribution\rev{,} which is the underlying distribution of samples. 
\del{The}{}\new{An} instance of IGO contains two hyperparameters, the learning rate\footnote{The inverse of the learning rate is called the population size in the context of cGA. Following the IGO framework, we consider the learning rate as a hyperparameter instead of the population size. \rev{We also distinguish the terms ``population size'' and ``sample size," where the later is the number of samples generated form the underlying distribution in each iteration.}} $\eta$ and the sample size $\lambda$, whose setting significantly affects the search efficiency. To obtain their recommended setting, several theoretical analyses investigate the relations between the search efficiency, the setting of hyperparameters, and the information of the search space, such as the number of dimensions $D$~\citep{Droste:2006, igo:analysis:2, igo:analysis:1}.

The search efficiencies of discrete evolutionary algorithms are often measured by the runtime, also called the first hitting time\del{, on representative benchmark functions}{}. In this context, the runtime is the number of iterations required until the optimal solution is obtained for the first time. 
\revdel{The major analyses on}{}\rev{The majority of the analyses in the} discrete domain investigated binary optimization methods even though runtime analyses of probabilistic model-based evolutionary algorithms are actively studied. However, the applications of stochastic algorithms are not limited to binary optimization. For example, \cite{asng:2019} \revdel{applied the family of categorical distributions to the IGO framework}derived \rrev{the} IGO algorithm \rrevdel{with}\rrev{for} the family of categorical distributions and developed an efficient neural architecture search for machine learning tasks. The optimization of the categorical domain can also be \revdel{applied to}\rev{utilized in} several applications, including graph optimization~\rev{\citep{graph-optimization1,graph-optimization2}} and genetic programming~\rev{\citep{gp:application}}. Investigation of the runtime of the optimization method on the categorical domain \rrevdel{has not yet been conducted}\rrev{has not been actively conducted}, regardless of their significant impact.\rrev{\footnote{\rrev{While our study was under review, we became aware of a recently published study by \cite{Multival:2023} that tackles a similar topic. We have chosen to retain our original notation to maintain consistency and clarity within our paper. The differences between this recent study and ours are discussed in Section~\ref{sec:conclusion}.}}}
The dependence of the search efficiency on the number of dimensions $D$ and categories $K$ indicates the proper settings for such applications and the ways of performance improvement.\rev{\footnote{\rev{We note that the variable $K$ is often used as population size in the context of cGA, while it is used as the number of categories in this paper.}}}

\subsection{Runtime Analysis of cGA} \label{sec:analysis-cga}
The works most related to our analysis are the runtime analyses of cGA.
In those analyses, OneMax and BinVal have been popularly investigated as representative linear pseudo-Boolean functions~\rev{\citep{Droste:2006, Domino:2018, Jump:2018, Sudholt:2019, Lengler:2021, Jump:2021}}\revdel{\citep{Droste:2006, Domino:2018, Jump:2018, Medium:2018, Sudholt:2019, Jump:2021}}.
OneMax counts the number of ones in a binary variable as its evaluation value, while BinVal represents the value of a binary variable regarded as a binary unsigned integer.

The first analysis of cGA on these functions was done by \cite{Droste:2006}.
This work investigated cGA without the border, which is often introduced to \rev{correct the distribution parameters to be within a certain range so that they are not fixed to $0$ or $1$.}\revdel{prevent the distribution parameters from being fixed to $0$ or $1$.}
The theoretical analysis provides the upper bound $O(D / \eta)$ and the lower bound $\Omega(\sqrt{D} / \eta)$ of the expected first hitting time (under the condition of finite first hitting time) and basic tail bounds $O(D / \eta)$ and $\Omega(\sqrt{D} / \ln D / \eta)$ on any linear function, where the conditions for $\eta$ and the tail probabilities are also provided. 
Additionally, this paper demonstrates that the expectations and tail bounds of the runtime on OneMax and BinVal are different.
More precisely, on OneMax, the expected first hitting time is $O(\sqrt{D}/\eta)$ and the probability of the runtime being $O(\sqrt{D}/\eta)$ is at least $1/2$ with the condition of $\eta = o(D^{-\frac{1}{2}})$.
On BinVal, the expected first hitting time is $\Omega(D/\eta)$ and the probability of the runtime being $O(D/\eta)$ is at least $1 - \exp(- \Theta(\eta^{-1}))$.


These results were refined later, owing to two observations: the first was that the bounds and the conditions for $\eta$ \revdel{are}\rev{were} improved. \cite{Sudholt:2019} analyzed the cGA with borders $\{D^{-1}, 1-D^{-1}\}$ and observed that the lower and upper bounds of the expected first hitting time were $\Omega(\sqrt{D}/\eta + D \ln D)$ with $\eta^{-1} = \Omega(D^{c})$ for some $c>0$ and $O(\sqrt{D}/\eta)$ with $\eta^{-1} = \Omega(\sqrt{D} \ln D)$, respectively.
\revdel{The second observation was that, for the analysis of the tail bound, the tail probability was given by greater than $1 - D^{-c}$ with some constant $c>0$.}{}
\rev{The second observation was that the tail bound has a probability of at least $1 - D^{-c}$ for some $c>0$.}
Referring to \citep{Domino:2018}, the probability of an event is said \revdel{as}{}\rev{to occur with} \revdel{{\it high probability}}\rev{high probability} when the event occurs with the probability of at least $1 - D^{-c}$. 
\revdel{For BinVal, }{}\cite{Domino:2018} showed that the runtime on BinVal was $\Theta(D/\eta)$ with a high probability when $\eta^{-1} = \Omega( D \ln D )$, and $\eta = D^{-c}$ for some $c>0$. 
\subsection{Contributions} \label{sec:contribution}
In this paper, we analyze a probabilistic model-based evolutionary algorithm \rrevdel{using}\rrev{that uses} the family of categorical distributions as its underlying distribution with sample size $\lambda = 2$.
We term this algorithm the categorical compact genetic algorithm (ccGA).
We select two typical linear functions on the categorical domain, categorical OneMax (COM) and \textsc{KVal}, as representative examples of simple objective functions and difficult objective functions. COM and \textsc{KVal} are the extensions of OneMax and BinVal on the binary domain, respectively.
As described in Section \ref{sec:analysis-cga}, the expected runtimes and \rrev{the} tail bounds of the runtime on OneMax and BinVal are different \revdel{with respect to (}w.r.t.\revdel{) } $D$.
Our analysis investigates how the number of categories $K$ affects the runtime of ccGA and the conditions on the learning rate to realize an efficient search on COM and \textsc{KVal}.
In the categorical domain, we say \rrev{that} an event occurs with high probability when it occurs with a probability of at least $1 - D^{-c} - K^{-c}$ for some $c>0$.
Moreover, our analysis shows the following results under a common assumption $\eta^{-1} = O((DK)^{c'})$ for some $c'>0$.
\begin{itemize}
    \item With a high probability, the runtime of ccGA is $O(\sqrt{D} \ln (DK) / \eta)$ when $\eta^{-1} = \Omega( \sqrt{D} K \ln(DK))$ and is $\Omega((\sqrt{D} + \ln K) / \eta)$ when $\eta^{-1} = \Omega(\sqrt{D} \ln D)$ on COM\rev{~(Theorem~\ref{thm:runtime-upper-one} and  Theorem~\ref{thm:runtime-lower-onemax})}.
    \item With a high probability, the runtime of ccGA is $\Theta(D \ln (DK) / \eta)$ when $\eta^{-1} = \Omega( D K^2 \ln K \ln(DK))$ on \textsc{KVal}\rev{~(Theorem~\ref{thm:runtime-upper-kval} and  Theorem~\ref{thm:runtime-lower-kval})}.
\end{itemize}
Considering the order w.r.t. $K$, while we obtain the same tail bound $\Theta(\ln K / \eta)$ on \rrev{both} COM and \textsc{KVal}, the conditions for the learning rate are different as $\eta^{-1} = \Omega(K \ln K)$ for the upper tail bound on COM and $\eta^{-1} = \Omega( K^2 (\ln K)^2)$ on \textsc{KVal}. 

The detailed proofs of the non-trivial drift theorems are another contribution of this study.
The drift theorem provides the tail bound of the first hitting time or the bound of the expected first hitting time of the target stochastic process $\{ \X \}$ based on the upper or lower bound of the conditional expectation of a single transition $\E[ \X[t+1] - \X \mid \mathcal{F}^{(t)}]$ under the filtration $\{\mathcal{F}^{(t)}\}$, termed as {\it drift}.
We extend the existing drift theorems in two ways to make our analysis rigorous.
The first way is to consider the $\mathcal{F}^{(t)}$-measurable event where the bound of the drift is significantly larger or smaller than the bound under the complementary event. \rrevdel{Tighter}\rrev{A tighter} tail bound of the first hitting time is achieved compared to that of the existing drift theorem when the complementary event rarely occurs. The second way is to skip the iteration where the process stays \rev{at }the same value, i.e., $\X[t+1] = \X$. If such a skip occurs frequently, our drift theorem shows \revdel{the}{}\rev{a} tighter tail bound. We first illustrate the explicit proofs of these drift theorems even though they were implicitly used in the runtime \revdel{analysis}\rev{analyses} of \rev{$(1,\lambda)$ EA~\citep{negativedrift:skipping} and} cGA~\citep{Friedrich:2017}. 

\subsection{Organization of Paper}
The rest of this paper is organized as follows.
In Section~\ref{sec:ccga}, we introduce the update rule and the background of ccGA.
In Section~\revdel{\ref{sec:general-drift-theorem}}\rev{\ref{sec:first-hitting-time}}, we define the first hitting time and \rrevdel{show}\rrev{present} \revdel{the useful lemmas for}\rev{lemmas that simplify} our analysis.
In Section~\revdel{\ref{sec:first-hitting-time}}\rev{\ref{sec:general-drift-theorem}}, we establish novel drift theorems that consider the conditional drift and \rrev{the} skipping process.
In Sections~\ref{sec:onemax} and \ref{sec:kval}, we investigate the upper and lower tail bounds of the first hitting times on COM and \textsc{KVal}, respectively.
In Section~\ref{sec:experiment}, we confirm our analysis results \revdel{can explain the actual behavior of ccGA}\rev{hold even when the practical constants are applied} \revdel{in the}\rev{by} numerical simulation\rev{s}. Finally, we conclude our paper in Section~\ref{sec:conclusion}.


\section{Categorical Compact Genetic Algorithm (ccGA)} \label{sec:ccga}

In this paper, we focus on the optimization problems of $D$-dimensional categorical variable $x = (x_1, \cdots, x_D)^\T$ whose $d$-th element $x_d$ is represented by \rev{a }$K$-dimensional one-hot vector. The $k$-th element of $x_d$ is denoted as $x_{d,k}$. The domain of $x$ is denoted as 
\begin{equation}
    \mathbb{K}^D_K = \left\{ x \in \{0, 1\}^{D \times K} \biggm\vert \revdel{{}^\forall d \in \llbracket 1,D \rrbracket,} \sum_{k=1}^{K} x_{d,k} = 1 \rev{\quad \text{for all} \enspace d \in \llbracket 1,D \rrbracket} \right\} \enspace,
\end{equation}
where $\llbracket a, b \rrbracket = \{\lceil a \rceil, \lceil a \rceil + 1, \cdots, \lfloor b \rfloor -1, \lfloor b \rfloor \} \subset \mathbb{N}$ for $a, b \in \mathbb{R}$ satisfying $a < b$.
\rev{We consider finding the maximizer of a given objective function $f: \mathbb{K}^D_K \to \R $ as
\begin{align}
    x^\ast = \argmax_{x \in \mathbb{K}^D_K} f(x) \rrev{\enspace.}
\end{align}}

We define \target[] as the optimization method for $\mathbb{K}^D_K$; the method is a probabilistic model-based algorithm with the family of categorical distributions $\{P_\theta\}$ as its underlying distribution. The probability mass function of the categorical distribution is given by
\begin{align}
P_\theta(x) = \prod^D_{d=1} \prod^K_{k=1} ( \theta_{d,k} )^{x_{d,k}} \enspace,
\end{align}
where $\theta \in \Theta$ is \rev{a }distribution parameter. The element of $\theta$ represents the probability of the corresponding category to be selected, i.e., $ \theta_{d,k} = \Pr(x_{d,k} = 1)$. The domain $\Theta$ is defined as
\begin{align}
\Theta = \left\{ \theta \in [0, 1]^{D \times K} \biggm\vert \revdel{{}^\forall d \in \llbracket 1, D \rrbracket, \enspace} \sum^K_{k=1} \theta_{d,k} = 1 \rev{\quad \text{for all} \enspace d \in \llbracket 1, D \rrbracket} \right\} \enspace.
\end{align}
The distribution parameter is initialized as
\begin{align}
\revdel{{}^\forall d \in \llbracket 1, D \rrbracket, \, {}^\forall k \in \llbracket 1, K \rrbracket, \quad} \param[0]_{d,k} = \frac{1}{K} \rev{\quad \text{for all} \enspace d \in \llbracket 1, D \rrbracket \enspace \text{and} \enspace k \in \llbracket 1, K \rrbracket} \enspace. \label{eq:InitialState}
\end{align}

The ccGA algorithm is described as follows:
\rrev{similar to} the cGA, two samples $x, x'$ are \revdel{sampled}\rev{generated} from $P_{\param}$ in each iteration. 
The superior sample $\sample[1]$ and inferior sample $\sample[2]$ are selected from $\{x, x'\}$, based on the evaluation value, which satisfies $f(\sample[1]) \geq f(\sample[2])$. 
In the case of a tie, $x$ and $x'$ are assigned to $\sample[1]$ and $\sample[2]$, respectively, which \rrevdel{works as}\rrev{is the} same as the random selection. 
The update rule of \target is summarized as
\begin{align}
\param[t+1] = \param + \eta (\sample[1] - \sample[2]) \enspace, \label{eq:Update}
\end{align}
where $\eta > 0$ is the learning rate.
The update procedure of \target is summarized in Algorithm~\ref{alg:ccga}. \revdel{Note that the update rule~\eqref{eq:Update} corresponds to the update rule of the cGA when $K = 2$. }
\revdel{We will investigate \target with the update rule~\eqref{eq:Update}.}

The update rule of ccGA is introduced in \citep{asng:2019} as a stochastic natural gradient method using the family of categorical distributions.
\rrev{Contrary to the update rule detailed in~\citep{asng:2019}, we do not restrict the domain of each element of $\theta$ to be $[\ell, u]$ for some $0 < \ell < u < 1$.}
Instead, we assume that $\eta$ satisfies $(\eta K)^{-1} \in \mathbb{N}$. We note that all elements of the distribution parameter are initialized to $1/K$ and they are not changed after they become \rrevdel{$1$ or $0$}\rrev{$0$ or $1$} because, in such cases, the corresponding elements of the samples $x, x'$ are always the same. Therefore, the distribution parameter $\param_{d,k}$ is not updated \rrevdel{above $1$ or below $0$}\rrev{below $0$ or above $1$} without modification of the update rule in \eqref{eq:Update}. This assumption eases the difficulty of the analysis.

To explain the justification \rrev{of} the update rule of ccGA, we mention that the\revdel{The} relation between cGA and \target can be explained using the information geometric optimization (IGO)~\citep{IGO:2017}. The IGO is a unified framework of stochastic search algorithms, and the IGO algorithm is instantiated by applying a family of probability distributions. The update in the IGO algorithm is a kind of stochastic natural gradient ascent~\citep{Amari:1998}.
\rrev{The} IGO framework recovers several well-known algorithms, and the update rules of cGA and \target are derived by applying the family of Bernoulli distributions and the family of categorical distributions with the sample size of two, respectively. \rrevdel{Note}\rrev{It should be noted} that \rrev{when $K = 2$,} the update rule~\eqref{eq:Update} corresponds to the update rule of the cGA. This background may help to understand our analysis results.

\begin{algorithm}[t]
\caption{\target maximizing $f$}
\label{alg:ccga}
\revdel{$t \leftarrow 0$, and $\param[0]_{d,k} \leftarrow 1 / K$ for all $d \in \llbracket 1, D \rrbracket, k \in \llbracket 1, K \rrbracket$ \\}
\rev{
Initialize the iterator as $t = 0$. \\
Initialize the distribution parameters as $\param[0]_{d,k} = 1 / K$ for $d \in \llbracket 1, D \rrbracket, k \in \llbracket 1, K \rrbracket$} \\
\While {terminate conditions are not fulfilled}{
Generate \del{$x_t, x'_t$}{}\new{$x^{(t)}, (x')^{(t)}$} from \rev{the }categorical distribution $P_{\param}$ \\
\del{
\lIf{$f(x_t) \geq f(x'_t)$}{Set $\sample[1] \leftarrow x_t$ and $\sample[2] \leftarrow x'_t$ }
\lElse{Set $\sample[1] \leftarrow x'_t$ and $\sample[2] \leftarrow x_t$ }
}{}
\rrev{
\If{$f(x^{(t)}) \geq f((x')^{(t)})$}{Set $\sample[1] \leftarrow x^{(t)}$ and $\sample[2] \leftarrow (x')^{(t)}$ }
\Else{Set $\sample[1] \leftarrow (x')^{(t)}$ and $\sample[2] \leftarrow x^{(t)}$ }
}
\revdel{$\param[t+1] \leftarrow \param + \eta (\sample[1] - \sample[2])$}\rev{Update the distribution parameter as $\param[t+1] = \param + \eta (\sample[1] - \sample[2])$} \\
$t \leftarrow t + 1$
}
\end{algorithm}

\section{Lemmas Related to First Hitting Time\rev{s}} \label{sec:first-hitting-time}
We investigate the \emph{first hitting time}, that is, the number of iterations $T_\mathrm{Hit}$ necessary for ccGA to generate the optimal solution $x^\ast$ of a given objective function for the first time. 
It is defined as
\begin{align}
T_\mathrm{Hit} = \min\left\{ t \in \mathbb{N}_0: x^\ast \in \{ \del{x_t, x'_t}{}\new{x^{(t)}, (x')^{(t)}} \} \right\} \enspace,
\end{align}
where \del{$x_t$ and $x'_t$}{}\new{$x^{(t)}$ and $(x')^{(t)}$} are samples described in Algorithm~\ref{alg:ccga} \new{(we denote them as $x$ and $x'$ in the following when the value of the iterator $t$ is \revdel{trivial}{}\rev{clear from context})}\footnote{We define the first hitting times introduced in this manuscript with the convention that \rrev{$\min \emptyset = \infty$}.}.
Our analysis focuses on objective functions with unique optimal solutions and investigates the tail bound of $T_\mathrm{Hit}$ on them. We consider the lower and upper tail bounds $T_\mathrm{lower}$ and $T_\mathrm{upper}$ in our analysis, which are given by functions of $D$, $K$ and $\eta$ and satisfy that the event $T_\mathrm{lower} < T_\mathrm{Hit} < T_\mathrm{upper}$ occurs with high probability.
We assume, without loss of generality (w.l.o.g.), that \rrev{the} categories of the optimal solution are \rev{the }first categories in all dimensions, i.e., $x^\ast_{d,1} = 1$ for all $d \in \llbracket 1, D\rrbracket$ as \target is invariant to permutation of categories. We \revdel{termed}{}\rev{term} the first category in each dimension \revdel{as }{}the optimal category. 

For \target[], the first hitting time and the transition of the distribution parameter are related.
In our analysis, the transition of a function of the distribution parameter, called a potential function, is investigated to derive the tail bound of $T_\mathrm{Hit}$.
Our potential functions relate the product and the summation of the elements of the distribution parameter $\param_{d,1}$ corresponding to the optimal categories.
The following lemmas provide a useful relationship between the tail bound of $T_\mathrm{Hit}$ and the tail bounds of the iteration where the product and the summation reach a given target value for the first time.

\begin{lemma} \label{lem:general-tail-upper}
Consider the update of \target[]. Assume the objective function has a unique optimal solution $x^\ast$ whose categories are the first categories in all dimensions. Let $\alpha \in [0,1]$ and 
\begin{align}
T_1 &= \min \left\{ t \in \mathbb{N}_0 : \prod^D_{d=1} \param_{d,1} \geq \alpha \right\} \\
T_2 &= \min \left\{ t \in \mathbb{N}_0 : \sum^D_{d=1} \param_{d,1} \geq D - 1 +\alpha \right\} \enspace.
\end{align}
Then, for any \new{$s, u \geq 0$} satisfying $u < \alpha / (D \eta)$, it holds \rrev{$T_1 \leq T_2$,}
\begin{align}
\Pr( T_\mathrm{Hit} \leq s + u ) &> 1 - \Pr( T_1 \geq s )  - (1 - p)^{2u} \enspace \text{and} \label{eq:upper-tail-1} \\
\Pr( T_\mathrm{Hit} \leq s + u ) &> 1 - \Pr( T_2 \geq s )  - (1 - p)^{2u} \enspace, \label{eq:upper-tail-2}
\end{align}
where $p = \alpha - \eta D u > 0$.
\end{lemma}

\begin{lemma} \label{lem:general-tail-lower}
Consider the update of \target[]. Assume the objective function has a unique optimal solution $x^\ast$ whose categories are the first categories in all dimensions. Let $\alpha \in [0,1]$ and 
\begin{align}
T_1 &= \min \left\{ t \in \mathbb{N}_0 : \prod^D_{d=1} \param_{d,1} \geq \alpha^D \right\} \\
T_2 &= \min \left\{ t \in \mathbb{N}_0 : \sum^D_{d=1} \param_{d,1} \geq \alpha D \right\} \enspace.
\end{align}
Then, for any \new{$s \geq 0$}, it holds \rrev{$T_1 \geq T_2$,}
\begin{align}
\Pr( T_\mathrm{Hit} \geq s ) &\geq 1 - \Pr( T_1 < s )  - 2 \new{(s+1)} \alpha^D \enspace \text{and} \label{eq:lower-tail-1} \\
\Pr( T_\mathrm{Hit} \geq s ) &\geq 1 - \Pr( T_2 < s )  - 2 \new{(s+1)} \alpha^D \enspace. \label{eq:lower-tail-2}
\end{align}
\end{lemma}

The proofs of the lemmas are shown in \revdel{the supplementary material}\rev{Section~\ref{apdx:sec-lemma-proof} in the appendix}. 

\section{Generalization of Drift Theorems} \label{sec:general-drift-theorem}


Our analysis is based on the drift theorems that are used in previous research on runtime analyses of evolutionary algorithms~\citep{Domino:2018,Multiplicative:2010}. 
The drift is the conditional expectation of a single-step difference of a target stochastic process with respect to \del{the}{}\rev{a }filtration\rev{\footnote{\rev{We denote a stochastic process $(X^{(t)})_{t \in \mathbb{N}_0}$ \rrevdel{adopted}\rrev{adapted} to a filtration $(\mathcal{F}^{(t)})_{t \in \mathbb{N}_0}$ as $(X^{(t)}, \mathcal{F}^{(t)})_{t \in \mathbb{N}_0}$. In this manuscript, we use the $\sigma$-algebra generated by $(\theta^{(0)}, \cdots \theta^{(t)})$ as $\mathcal{F}^{(t)}$.}}}.
The drift theorem provides the tail bound and the bound of the expectation of the first hitting time using the upper or lower bound of the drift. 
\revdel{The}{}\rev{A} tighter bound of the drift generally provides a \revdel{more strict}{}\rev{stricter} bound of the first hitting time.
In the following, we generalize the well-known drift theorems to allow different kinds of bounds of the drift and obtain a \rrevdel{more strict}\rrev{stricter} bound than the existing drift theorems \new{for analyzing ccGA}.

\subsection{Conditional Drift Theorem\rev{s}}
The drift changes significantly depending on events measurable with respect to the filtration in some cases, including our analysis.
This makes the upper or lower bound of the drift we can obtain too small or too large to derive the tight tail bound of \rev{the }runtime. To deal with this problem, we introduce the conditional drift theorems, which consider a sequence of (likely) events $(\event)_{t \in \mathbb{N}_0 }$ where the bounds of \rev{the }drift are not significantly large or small.
We note that such an approach appears in previous research~\citep{Domino:2018,Friedrich:2017}. Nevertheless, to our best knowledge, \rrevdel{a proof with an explicit explanation does not exist}\rrev{there is no proof with an explicit explanation}. The conditional drift theorems lead to a rigorous proof for this approach.

\rev{We provide three kinds of conditional drift theorems that assume different drift bounds. Theorem~\ref{theo:additive-d-lower} requires an upper bound of the drift under the event $\event$ to show an upper tail bound of the first hitting time, while Theorem~\ref{theo:additive-d-upper} requires a lower bound of the drift to show a lower tail bound. Theorem~\ref{theo:multiplicative-d} requires a negative multiplicative upper bound under the event $\event$ to show a lower tail bound.}
\revdel{The}\rev{These} conditional drift theorems are derived from existing drift theorems. We introduce the conditional drift theorems based on the additive drift theorem~\cite[Theorem\rrev{s}~8, 9]{Kotzing:2014} and multiplicative drift theorem~\cite[Theorem~1]{Multiplicative:2010} as follows:

\begin{theorem} \label{theo:additive-d-lower}
Consider a stochastic process $(\X, \F)_{t \in \mathbb{N}_0}$ with $\X[0] \leq 0$. Consider a sequence of events $(\event)_{t \in \mathbb{N}_0 }$, where $\event \in \F$ and $\event \supseteq \event[t+1]$. Let $m \in \mathbb{R}_{>0}$ and
\begin{align}
T = \min\{ t \in \mathbb{N}_0 : \X \geq m \} \enspace. \label{eq:additive-drift-d-lower-T}
\end{align}
Then, for constants $c > 0$ and $ \varepsilon \in [0, c]$,  if it holds, for all $t < T$,
\begin{align}
\E[ \X[t+1] - \X \mid \F ] \indic &\leq \varepsilon \indic \\
|\X[t+1] - \X | \indic &< c \enspace,
\end{align}
then it holds, for all $n \leq m / (2 \varepsilon)$,
\begin{align}
\Pr \left( T < n \right) \leq \exp \left( - \frac{ m^2 }{8 c^2 n} \right) + \Pr( \bevent[n-1] ) \enspace. \label{eq:additive-d-lower-prob-upper}
\end{align}
\end{theorem}

\begin{theorem} \label{theo:additive-d-upper}
Consider a stochastic process $(\X, \F)_{t \in \mathbb{N}_0}$ with $\X[0] \geq 0$. Consider a sequence of events $( \event )_{t \in \mathbb{N}_0 }$, where $\event \in \F$ and $\event \supseteq \event[t+1]$. Let $m \in \mathbb{R}_{>0}$ and
\begin{align}
T = \min\{ t \in \mathbb{N}_0 : \X \geq m \} \enspace. \label{eq:additive-drift-d-upper-T}
\end{align}
Then, for constants $c > 0$ and $ \varepsilon \in [0, c]$,  if it holds, for all $t < T$,
\begin{align}
\E[ \X[t+1] - \X \mid \F ] \indic &\geq \varepsilon \indic \\
|\X[t+1] - \X | \indic &< c \enspace,
\end{align}
then it holds, for all $n \geq 2 m / \varepsilon$,
\begin{align}
\Pr \left( T \geq n \right) \leq \exp \left( - \frac{n \varepsilon^2}{8 c^2} \right) + \Pr( \bevent[n-1] ) \enspace. \label{eq:additive-d-upper-prob-upper}
\end{align}
\end{theorem}

\begin{theorem} \label{theo:multiplicative-d}
Consider a stochastic process $(\X, \F)_{t \in \mathbb{N}_0}$ and a sequence of events $( \event )_{t \in \mathbb{N}_0 }$, where $\event \in \F$ and $\event \supseteq \event[t+1]$.
Assume $\X$ takes the value on $\{0\} \cup [x_{\min}, x_{\max}]$ for all $t \in \mathbb{N}$, where $0 < x_{\min} < x_{\max} < \infty$, and $\X$ remains zero after \revdel{once $\X$}{}\rev{$\X$ once} reaches zero. Let
\begin{align}
T = \min\{ t \in \mathbb{N}_0 : \X = 0 \} \enspace.
\end{align}
Then, for a constant $\varepsilon \in [0, 1]$, if it holds
\begin{align}
\E[ \X[t+1] - \X \mid \F ] \indic \leq - \varepsilon \X \indic \enspace \label{eq:multiplicative-d-drift-cond}
\end{align}
for all $t < T$, it holds, for any $r > 0$,
\begin{align}
\Pr \left( T > \frac{r + \ln( \X[0] / x_{\min})}{\varepsilon} \right) \leq \exp \left( - r \right) + \Pr( \bevent[n-1] ) \enspace, \label{eq:multiplicative-d-prob-upper}
\end{align}
where $n = \lceil (r + \ln (\X[0] / x_{\min}) ) / \varepsilon \rceil$.
\end{theorem}

The proofs of theorems are shown in \revdel{the supplementary material}\rev{Section~\ref{apdx:sec-proof} in the appendix}. We note that if the event $\event$ occurs with probability (w.p.) 1 for all $t \in \mathbb{N}_0$, Theorem\rrev{s}~\ref{theo:additive-d-lower}, \ref{theo:additive-d-upper} and Theorem~\ref{theo:multiplicative-d} recover \cite[Theorem~1]{Kotzing:2016}, \cite[Theorem~2]{Kotzing:2016} and a part of \cite[Theorem~1]{Multiplicative:2010}, respectively
\footnote{\rev{This is shown by the fact that, for the process $\tX$ and stopping time $\tilde{T}$ w.r.t. $\tX$ defined in each proof of Theorem\rrev{s}~\ref{theo:additive-d-lower}, \ref{theo:additive-d-upper} and \ref{theo:multiplicative-d}, $\tX = \X$ and $\tilde{T} = T$ almost everywhere, which shows $\E[ \X[t+1] - \X \mid \F] = \E[ \tX[t+1] - \tX \mid \F]$ and $\Pr(T < n) = \Pr(\tilde{T} < n)$ for all $n \in \mathbb{N}$, respectively.}}. 


\subsection{Drift Theorem for Skipping Process\rev{es}}
Next, we demonstrate an extension of the negative drift theorem~\revdel{\citep{Kotzing:2016}}\rrevdel{\rev{\cite{Kotzing:2016, Oliveto:2011:simplified, Oliveto:2012:erratum}}}\rrev{\citep{Kotzing:2016, Oliveto:2011:simplified, Oliveto:2012:erratum}} for \rrevdel{the }stochastic process\rev{es} which often stay\revdel{s}{} in the same state. By introducing the conditional probability of staying, we establish a novel drift theorem based on a bound\revdel{s}{} of the drift \revdel{which}{}\rev{that} is given by the probability multiplied by a constant. We can obtain more detailed information about the drift compared \revdel{with}{}\rev{to} the original negative drift theorem, which bounds the drift by a constant value. This leads to a strict tail bound of the first hitting time in our analysis. 

The derived drift theorem is as follows.
\begin{theorem} \label{theo:negative-skip}
Consider a stochastic process $(\X, \F)_{t \in \mathbb{N}_0}$ with $\X[0] \leq 0$. Let $T$ be a stopping time defined as
\begin{align}
T = \min \left\{ t \in \mathbb{N} : \X \geq m \right\} \enspace.
\end{align}
Assume that, for all $t$, there are constants $0 < \varepsilon < m / 2$ and $0 < c < m$ satisfying
\begin{align}
\E[ \X[t+1] - \X \mid \F] &\leq - \varepsilon \Pr( \X[t+1] \neq \X \mid \F) \\
| \X[t+1] - \X | &\leq c \qquad \text{w.p. 1} \enspace.
\end{align}
Then, for all $n \geq 0$,
\begin{align}
\Pr( T \leq n ) \leq \frac{2m n}{\varepsilon} \exp \left( - \frac{m \varepsilon}{4 c^2} \right)  \enspace.
\end{align}
\end{theorem}
The proof of Theorem~\ref{theo:negative-skip} \revdel{are}{}\rev{is} shown in \revdel{the supplementary material}\rev{Section~\ref{apdx:sec-proof-skip} in the appendix}. Some existing works~\citep{negativedrift:skipping, Friedrich:2017} reconstruct a Markov chain by ignoring the transition to the same state in the given Markov chain, called self-loops, and applied the negative drift theorem to the modified Markov chain. Although it seems \rrev{intuitively} correct, \rrevdel{the}\rrev{an} explicit explanation was not provided. In contrast, we reformulate the theorem in~\citep{Fan:2012}, which was applied to derive the negative drift theorem, to \rrevdel{show}\rrev{provide} an unambiguous proof.

\section{Runtime Analysis on Categorical OneMax}
\label{sec:onemax}

In the theoretical analysis of evolutionary algorithms for \revdel{the }{}binary optimization, OneMax is often picked up as one of the simplest but most widely used benchmark functions to measure the search efficiency \citep{Droste:2006,Sudholt:2019}. 
In this section, we introduce a generalization of OneMax, termed Categorical OneMax (COM),  defined in the categorical domain as
\begin{align}
\ONE(x) = \sum_{d=1}^D x_{d,1} \enspace.
\end{align}
The function value of COM is the number of the optimal categories in the inputted categorical variable. Therefore, the distinction of the categories except for the optimal categories is not necessary for the analysis on COM because the function value is determined by whether the optimal categories are selected or not. This implies \rrev{that} the behavior of \target has some common properties \rrevdel{to}\rrev{with} cGA optimizing OneMax. The next lemma shows the basic properties of the number of the optimal categories in samples, which are derived in a similar way \rrevdel{as}\rrev{to} the analysis of cGA.

\begin{lemma} \label{thm:Delta_bound}
Consider $D$-dimensional categorical variables $x, x' \in \mathbb{K}^D_K$ generated from the categorical distribution $P_{\theta}$ with distribution parameter $\theta \in \Theta$.
Let us denote $\delta := \sum_{d =1}^D ( x_{d,1} - x'_{d,1} ) $. Then, it satisfies $\Pr(\delta = 0) \geq ( 4 \sqrt{D} )^{-1}$ and $\E_{P_{\theta}}[ |\delta| ] \leq \sqrt{D / 2}$.
\end{lemma}

\begin{proof}[Proof of Lemma~\ref{thm:Delta_bound}]
Let $x_{\mathrm{first}} := (x_{1,1}, \cdots, x_{D,1})^\T$ and $x'_{\mathrm{first}} := (x'_{1,1}, \cdots, x'_{D,1})^\T$ be vectors generated from $x$ and $x'$ whose elements correspond to the first category, respectively. The expectation of $| \delta |$ is determined by the law of $x_{\mathrm{first}}$ and $x'_{\mathrm{first}}$ since $| \delta |$ is determined by $x_{\mathrm{first}}$ and $x'_{\mathrm{first}}$. We note \rev{that }$x_{\mathrm{first}}$ and $x'_{\mathrm{first}}$ are independent and identically distributed, whose law is given by the Bernoulli distribution with the probability mass function 
\begin{align}
p_\theta(x_{\mathrm{first}}) = \prod^D_{d=1} (\theta_{d,1})^{x_{d,1}} (1 - \theta_{d,1})^{( 1 - x_{d,1})} \enspace.
\end{align}
Then, the lower bound of $\Pr(\delta = 0)$ is obtained by applying \cite[Lemma 3]{Friedrich:2017} \rev{as $1 / ( 4 \sqrt{\sum_{d=1}^D \indic[{\theta_{d,1} > 0}]} )$, which is larger than $( 4 \sqrt{D} )^{-1}$}. Moreover, \cite[Lemma 4]{Droste:2006} shows the upper bound of $\E_{P_{\theta}}[ |\delta| ]$ as $\sqrt{2 \sum^D_{d=1}\theta_{d,1} (1 - \theta_{d,1} )}$. Then, considering the inequality $\theta_{d,1} (1 - \theta_{d,1} ) \leq 1 / 4$ finishes the proof.
\end{proof}

\rev{The statements of \cite[Lemma 3]{Friedrich:2017} and \cite[Lemma 4]{Droste:2006} are found in Appendix~\ref{apdx:sec:existing}.}
In the next subsections, we will investigate the upper and lower tail bounds of ccGA on COM, respectively.

\subsection{Upper Tail Bound of \rev{the }Runtime on COM}

\subsubsection{\revdelsec{Outline and Main Analysis Result}\rev{Main Result and Outline of the Proof}}
We firstly show the outline of the derivation for the upper tail bound of the runtime of the ccGA on COM. 
There are two \revdel{difficulties}\rev{main points} in our analysis. The first is {\it genetic drift}, the situation where $\param_{d,1}$ gets too small~\rev{\citep{Doerr:2020:tevc}}. In genetic drift, the expected increase $\E[\param[t+1]_{d,1} - \param[t]_{d,1} \mid \F]$ also gets too small and the runtime tends to be large.
Similar to OneMax on the binary domain~\cite[Lemma~4]{Sudholt:2019}, a situation where $\param_{d,1}$ becomes severely small compared \rrevdel{with}\rrev{to} the initial value rarely occurs on COM when the learning rate is small enough. In Lemma~\ref{thm:theta-lower-onemax}, we will show that $\param_{d,1}$ is maintained above $1/(2K)$ with high probability with \rrev{a} small learning rate using Theorem~\ref{theo:negative-skip}.

Designing \revdel{the}\rev{a} suitable potential function to apply the drift theorem is another \revdel{challenge}\rev{main point}.
We consider the potential function \revdel{, which}{}\rev{that} reflects the optimization process and has tractable properties to obtain the upper bound as close as possible using the drift theorem.
The \revdel{possible choices of}\rev{established} potentials are $1 - \param[t]_{d,1}$ or $\sum_{d=1}^D ( 1 - \param[t]_{d,1})$, which are used in the runtime analyses of cGA in~\rev{\citep{Sudholt:2019, Lengler:2021}}\revdel{\citep{Medium:2018, Sudholt:2019}}.
However, \rrevdel{as}\rrev{since} the distribution parameters are initialized by $1 / K$, one may need to tackle the intractable drift when $\theta_{d,1}$ is small. More precisely, the lower bounds of the drifts of those choices become too small to get a reasonable analysis result.

To deal with this problem, we define the potential $\X_d$ for $d \in \llbracket 1, D \rrbracket$ as
\begin{align}
\X[t]_d =
\begin{cases}
\dfrac{1 - \eta}{\eta} &\enspace \text{if} \enspace 0 \leq \param[t]_{d,1} < \eta \\
\dfrac{1 - \param[t]_{d,1}}{\param[t]_{d,1}} &\enspace \text{if} \enspace \eta \leq \param[t]_{d,1} \leq 1
\end{cases}
\enspace. \label{eq:onemax-potential}
\end{align}
The potential value is given by $( 1- \param_{d,1}) / \param_{d,1}$ until $\param_{d,1}$ is updated to zero unless $\param_{d,1}$ is zero, as we assume $(\eta K)^{-1} \in \mathbb{N}$ and $\param[0]_{d,1} = 1 / K$ for all $d \in \intrange{1}{D}$. 
\rev{Intuitively, this potential function is designed so that \rrevdel{its dynamics is enhanced when $\param_{d,1}$ is small and makes a similar dynamics of the existing potential $1- \param_{d,1}$ when $\param_{d,1}$ is relatively large.}\rrev{the drifts divided by current potential value $\E[\X[t+1]_d - \X[t]_d \mid \F] / \X[t]_d$ are almost the same value for both small and large $\param_{d,1}$, differently from the existing potential $1 - \param[t]_{d,1}$\footnote{\rrev{We have a lower bound of the drift of $Y_d^{(t)} := 1 - \param[t]_{d,1}$ proportional to $\param_{d,1} Y_d^{(t)}$ by considering $\E[\param[t]_{d,1} - \param[t+1]_{d,1}\mid \F] \geq - \eta \Pr( x_{d,1} \neq x'_{d,1} \mid \F) = - 2 \eta \param_{d,1} (1 - \param_{d,1})$.}}.}}
\rrev{This property allows us to apply Theorem~\ref{theo:multiplicative-d} in our analysis.}
We consider the case $\param[t]_{d,1} = 0$ as a special case \rrevdel{to prevent the potential value taking \revdel{infinity}\rev{undefined value}}\rrev{to avoid division by zero}.

\begin{figure}
  \begin{center}
    \includegraphics[width=\linewidth]{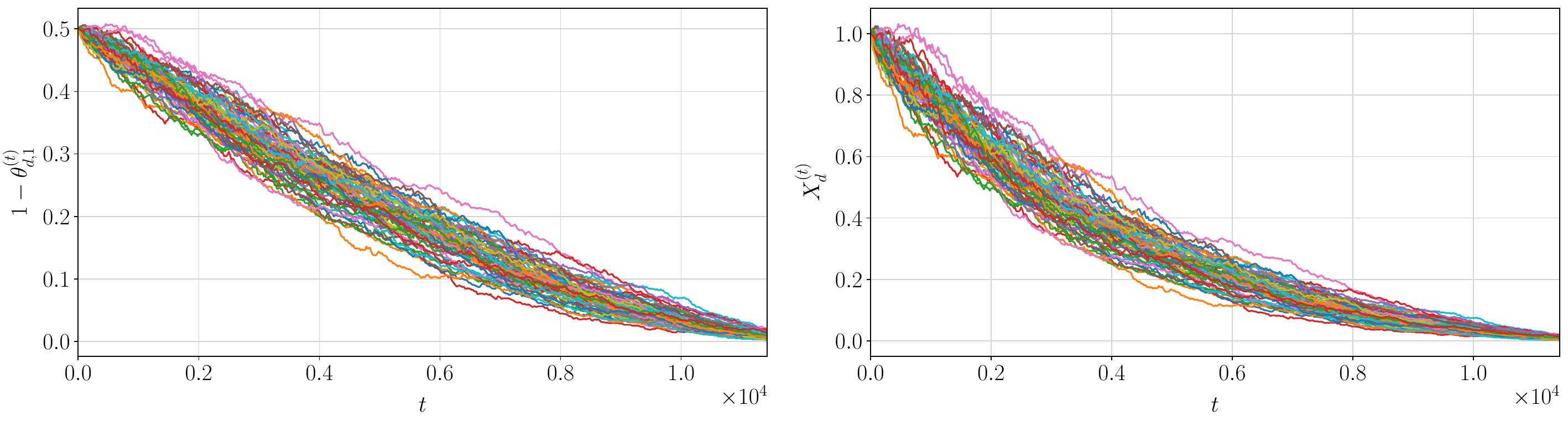}
    \vspace{-4mm}
    \caption{The \rrev{values} of $1 - \param[t]_{d,1}$~(left) and $\X_d$ \rev{defined }in \eqref{eq:onemax-potential}~(right) \rrev{in one typical trial of optimizing} COM with $D=64$ and $K=2$. \rrev{The values for each dimension are plotted as separate lines.}}
    \label{fig:potential_one_k2}
  \end{center}
\end{figure}

\begin{figure}
  \begin{center}
    \includegraphics[width=\linewidth]{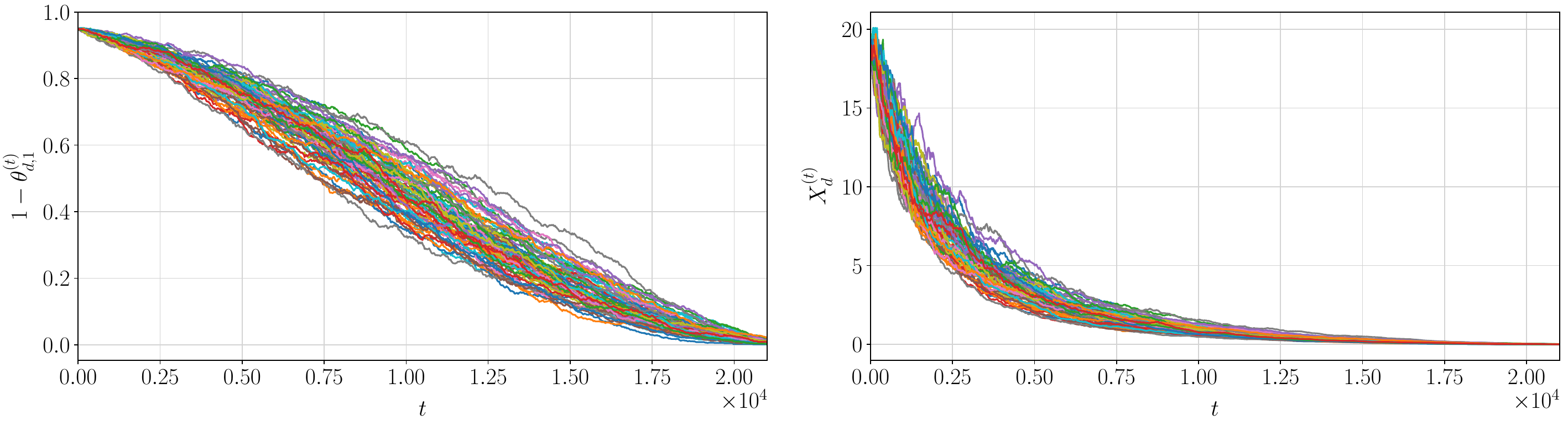}
    \vspace{-4mm}
    \caption{The \rrev{values} of $1 - \param[t]_{d,1}$~(left) and $\X_d$ \rev{defined }in \eqref{eq:onemax-potential}~(right) \rrev{in one typical trial of optimizing} COM with $D=64$ and $K=20$. \rrev{The values for each dimension are plotted as separate lines.}}
    \label{fig:potential_one}
  \end{center}
\end{figure}

\del{Figure~\ref{fig:potential_one} shows the transition of $1 - \param[t]_{d,1}$ and $\X_d$ on COM with $D=64$ and $K=20$. Focusing on the transition of $1 - \param[t]_{d,1}$, \del{when the number of categories $K$ is large, }{}$1 - \param[t]_{d,1}$ changes slowly in the early stage of optimization. On the other hand, the decrease speed of our potential $\X_d$ in the early stage is almost proportional to $\X_d$ itself because the transition is almost linear when plotting $\X_d$ in log-scale as shown in Figure~\ref{fig:potential_one}. This property becomes more significant the larger $K$ is. Figure~\ref{fig:potential_one_k2} shows the transition of $1 - \param[t]_{d,1}$ and $\X_d$ on COM with $D=64$ and $K=2$. The transition of $1 - \param[t]_{d,1}$ in the early stage is not so slow.
Therefore, this potential function is less effective in the binary domain but more effective in the categorical domain.
\del{This property is tractable}{} It is especially effective when applying Theorem~\ref{theo:multiplicative-d} to show the tight upper tail bound of the iteration when our potential function reaches $0$, which means $\param_{d,1}=1$.}{}
\new{
To display the importance of our potential, we show the \rrevdel{transitions}\rrev{values} of $1 - \param[t]_{d,1}$ and $\X_d$ on COM with $D=64$ and $K=2$ in Figure~\ref{fig:potential_one_k2}, and the \rrevdel{transitions}\rrev{values} with $D=64$ and $K=20$ in Figure~\ref{fig:potential_one}. 
\rev{The learning rate was set as $\eta^{-1} = \lceil \sqrt{D} \ln (DK) \rceil K$ as used in the experiment in Section~\ref{sec:experiment}.}
Focusing on the case of $K=2$, i.e., in binary search space, the \rrevdel{shapes of transition curves}\rrev{changes of two potential values} are almost the same. On the other hand, $1 - \param[t]_{d,1}$ hardly moves in the early stage of optimization when $K=20$, while our potential $\X_d$ decreases, \rrevdel{as well as}\rrev{as in} the case of $K=2$. In the early stage, the setting of the initial distribution parameter $1/K$ significantly affects the dynamics of \rev{the }potential, and the potential whose dynamics are almost the same when \rrevdel{setting $K$ differently}\rrev{$K$ is set differently} is tractable in our analysis. The decreased speed of $\X_d$ \revdel{in the early stage }is almost proportional to $\X_d$ itself because \rrevdel{the transition}\rrev{it} is \revdel{almost linear when plotting $\X_d$ in the log scale}\rev{almost exponential\rrev{ly} decay\rrev{ing}}. The dynamics of our potential \rrevdel{are \revdel{tractable}{}\rev{feasible} to applying Theorem~\ref{theo:multiplicative-d} to show \revdel{the}}\rrev{allow us to apply Theorem~\ref{theo:multiplicative-d} in a way that yields} \rev{a} tight upper tail bound of the iteration when \rrevdel{our potential function}\rrev{$\X_d$} reaches $0$, even when $K$ is large.}
When $\X_d = 0$, which means $\param_{d,1}=1$, is satisfied for all $d \in \intrange{1}{D}$, ccGA generates the optimal solution w.p. $1$.

The main analysis result is given by the next theorem.
\begin{theorem} \label{thm:runtime-upper-one}
Consider the update of ccGA on COM. \new{Assume $\eta := \eta(D,K)$ is given by a function of $D$ and $K$ and there exists \rev{a} strictly positive constant value $\cone[2] > 0$ which \revdel{satisfy}\rev{satisfies} $(\eta K)^{-1} \in \mathbb{N}$ and $\cone[1] \sqrt{D} K \ln (DK) \leq \eta^{-1} \leq (DK)^{\cone[2]}$ for any $D \in \mathbb{N}$, $K \in \mathbb{N}$ and $\cone[1] = 32 ( 7/2 + \new{2}\cone[2])$}. 
Then, 
the following \revdel{is held}{}\rev{holds} for all $D \geq D_1$: For the runtime $T_\mathrm{Hit}$,
\begin{align}
\Pr\left( T_\mathrm{Hit} \leq \frac{ \cone[3] \sqrt{D} \ln (DK) }{ \eta } \right) \geq 1 - (DK)^{- \cone[4]}  \enspace, \label{eq:runtime-upper-one}
\end{align}
\new{where $c_3 = 4(c_2 + 3), c_4 = 1/2$ and $D_1 = (4c_3+1)^2$.}
\end{theorem}
\kento{The main reason why we show the upper tail bound of runtime on COM is to compare the (lower) tail bound of runtime on \textsc{KVal}. The comparison of result for cGA on onemax is described below the proof.}



\subsubsection{Details of \rev{the }Proofs}

In the next lemma, we \revdel{show}\rev{derive a lower bound of} the probability that $\param_{d,1}$ drops below $1/(2K)$ within $u$ iterations in the optimization of COM.

\begin{lemma} \label{thm:theta-lower-onemax}
Consider the update of \target on COM. Let $d \in \llbracket 1,D \rrbracket$ be arbitrary. Let $T_d = \min\{ t \in \mathbb{N}_0 : \param_{d, 1} \leq 1 / (2K) \}$. Assume $D \geq 2$. Then, for any $u \in \mathbb{N}_0$,
\begin{align}
\Pr( T_d > u ) \geq 1 - \frac{4 u \sqrt{D-1}}{K \eta} \exp \left( - \frac{1 }{32 \eta K \sqrt{D-1}} \right) \enspace.
\end{align}
\end{lemma}

\begin{proof}

We will show first a lower bound of the drift of $\param_{d,1}$ as
\begin{align}
\E[ \param[t+1]_{d,1} - \param_{d,1} \mid  \F ] \geq \eta \frac{\Pr( \theta^{(t+1)}_{d,1} \neq \theta^{(t)}_{d,1} \mid \F ) }{4 \sqrt{D-1}} \enspace. \label{eq:onemax-theta-drift-lower} 
\end{align}
Given the samples $x$ and $x'$ generated in the $t$-th iteration, we consider $\delta^{(t+1)}_d$ defined and transformed as
\begin{align}
\delta^{(t+1)}_d &:= \sample[1]_{d,1} - \sample[2]_{d,1} \\
&= (x_{d,1} - x'_{d,1}) \indic[{ \ONE(x) \geq \ONE(x') }] - (x_{d,1} - x'_{d,1}) \indic[{ \ONE(x) < \ONE(x') }] \enspace.
\end{align}
Let us define $\alpha = x_{d,1} - x'_{d,1}$ and $\beta = \sum_{d' \neq d} x_{d',1} - x'_{d',1}$. Note that $\ONE(x)-\ONE(x') = \alpha + \beta$. Then we have
\begin{align}
\delta^{(t+1)}_d &= \alpha \indic[{ \alpha + \beta \geq 0 }] - \alpha \indic[{ \alpha + \beta < 0 }] \enspace. \label{eq:onemax-delta-trans}
\end{align}
Since $\alpha \in \{ -1, 0, 1\}$, we can transform the indicators $\indic[{ \alpha + \beta \geq 0 }]$ and $\indic[{ \alpha + \beta < 0 }]$ as
\begin{align}
&\indic[{ \alpha + \beta \geq 0 }] = \indic[{ \beta \geq 1 }] + \indic[{ \alpha \geq 0 }] \indic[{ \beta = 0 }] + \indic[{ \alpha = 1 }] \indic[{ \beta = -1 }] \\
&\indic[{ \alpha + \beta < 0 }] = \indic[{ \beta < -1 }] + \indic[{ \alpha \leq 0 }] \indic[{ \beta = -1 }] + \indic[{ \alpha = -1 }] \indic[{ \beta = 0 }] \enspace.
\end{align}
Altogether,
\begin{align}
\delta^{(t+1)}_d &= \alpha \indic[{ \beta \geq 1 }] + \alpha \indic[{ \alpha \geq 0 }] \indic[{ \beta = 0 }] + \alpha \indic[{ \alpha = 1 }] \indic[{ \beta = -1 }] \notag \\
&\qquad  - \alpha \indic[{ \beta < -1 }] - \alpha \indic[{ \alpha \leq 0 }] \indic[{ \beta = -1 }] - \alpha \indic[{ \alpha = -1 }] \indic[{ \beta = 0 }] \\
&= \alpha \indic[{ \beta \geq 1 }] + \indic[{ \alpha = 1 }] \indic[{ \beta = 0 }] + \indic[{ \alpha = 1 }] \indic[{ \beta = -1 }] \notag \\
&\qquad  - \alpha \indic[{ \beta < -1 }] + \indic[{ \alpha = -1 }] \indic[{ \beta = -1 }] +  \indic[{ \alpha = -1 }] \indic[{ \beta = 0 }] \\
&= \alpha \left( \indic[{ \beta \geq 1 }] - \indic[{ \beta < -1 }] \right) + \indic[{ \alpha \neq 0 }] \indic[{ \beta \in \{ -1, 0\} }] 
\end{align}
Since $\alpha$ and $\beta$ are conditionally independent given $\param$, we have
\begin{align}
&\E[ \delta^{(t+1)}_d \mid \F] \notag \\
&\quad = \E[ \alpha \mid \F] \E[ \indic[{ \beta \geq 1 }] - \indic[{ \beta < -1 }] \mid \F] + \E[ \indic[{ \alpha \neq 0 }] \mid \F ] \E[ \indic[{ \beta \in \{ -1, 0\} }] \mid \F] \\
&\quad = \Pr( \alpha \neq 0 \mid \F ) \Pr( \beta \in \{ -1, 0\} \mid \F) \\
&\quad \geq \Pr( \alpha \neq 0 \mid \F ) \Pr( \beta = 0 \mid \F) \enspace,
\end{align}
where the second transformation uses $\E[ \alpha \mid \F] = 0$. The event $\alpha \neq 0$ is equivalent to $\delta^{(t+1)}_d \neq 0$. Moreover, since $\beta$ is obtained from two samples from \revdel{$D-1$ }{}\rrev{the }\rev{$(D-1)$-}dimensional categorical distribution, the lower bound of $\Pr(\beta = 0 \mid \F)$ is obtained as $(4 \sqrt{D-1})^{-1}$ from Lemma~\ref{thm:Delta_bound}. Therefore, 
\begin{align}
\E[ \delta^{(t+1)}_d \mid \F] 
& \geq \frac{ \Pr( \delta^{(t+1)}_d \neq 0 \mid \F ) }{ 4 \sqrt{D-1} } \enspace.
\end{align}
The inequality in \eqref{eq:onemax-theta-drift-lower} is proved as $\param[t+1]_{d,1} - \param_{d,1} = \eta \delta^{(t+1)}_d$.

\del{
For estimation of the upper bound of $T_d$, we ignore the self-loops, which is in the iterations where $\param[t+1]_{d,1} = \param_{d,1} $. Formally, we consider the restriction of $\param_{d,1}$ to the iterations where $\param[t+1]_{d,1} \neq \param_{d,1}$ until it reaches $0$ or $1$ and, after it reaches $0$ or $1$ once, it increases by $\eta$ in each iteration. We denote the process as $\{ \hat{\theta}^{(s)}_{d,1} : s\in \mathbb{N}_0 \}$. Note the initial state is given by $\hat{\theta}^{(0)}_{d,1} = 1 / K$. We also define $\hat{T}_d = \min\{ s \in \mathbb{N}_0 : \hat{\theta}^{(s)}_{d,1} \leq 1 / (2 K) \}$. The random variable $\hat{T}_d$ cannot be greater than $T_d$ because the hitting time cannot be greater when removing the self-loops and $\param_{d,1}$ does not decrease after $\param_{d,1}$ reaches one. 
Considering the conditional expectation of $\delta^{(t+1)}_d$ conditioned on $\delta^{(t+1)}_d \neq 0$ shows
\begin{align}
\E[ \hat{\theta}^{(s+1)}_{d,1} - \hat{\theta}^{(s)}_{d,1} \mid  \hat{\mathcal{F}}^{(s)}_d ] &\geq \frac{ \eta }{4 \sqrt{D-1}} \enspace
\end{align}
for $s < \hat{T}_d$, where $\hat{\mathcal{F}}_d$ and $\event[s]$ are the $\sigma$-algebra associated to $\hat{\theta}^{(s')}$ for $s' \in \intrange{1}{s}$ and the event $\hat{\theta}^{(s)}_{d,1} \notin\{0,1\}$, respectively.

Since $| \hat{\theta}^{(s+1)}_{d,1} - \hat{\theta}^{(s)}_{d,1} | = \eta < \sqrt{2} \eta$, applying \cite[Theorem~3]{Kotzing:2016} with the stochastic process $\{ 1/K - \hat{\theta}^{(s)}_{d,1} : s \in \mathbb{N}_0 \}$ shows
\begin{align}
\Pr(\hat{T}_d \leq u) &\leq u \exp\left( - \frac{1}{32 \eta K} \frac{1}{\sqrt{D-1}} \right) \enspace
\end{align}
for any $u \in \mathbb{N}_0$. This is end of the proof.}{}

\new{
Finally, since $|\param[t+1]_{d,1} - \param_{d,1}| \leq \eta$, Theorem~\ref{theo:negative-skip} shows \revdel{the}\rev{an} upper bound of $\Pr(T_d \leq u)$ as
\begin{align}
&\frac{2 u}{2K} \left( \frac{\eta}{4 \sqrt{D-1}} \right)^{-1} \exp\left( - \frac{1}{4 \cdot 2K \eta^2} \frac{\eta}{4 \sqrt{D-1}} \right) \notag \\
&\qquad\qquad\qquad = \frac{4 u \sqrt{D-1}}{K \eta} \exp\left( - \frac{1}{ 32 K \eta \sqrt{D-1}} \right) \enspace.
\end{align}
\new{We note the values $\varepsilon, c, m$ and $n$ in Theorem~\ref{theo:negative-skip} are set as
\begin{align}
    \varepsilon = \frac{\eta}{4 \sqrt{D-1}} \enspace, \quad
    c = \eta \enspace, \quad
    m = \frac{1}{2 K} \enspace, \quad
    n = u \enspace.
\end{align}
}
\revdel{This is the end of the proof.}{}\rev{This concludes the proof.}
}
\end{proof}

The proof of Theorem~\ref{thm:runtime-upper-one} is as follows.

\begin{proof}[Proof of Theorem~\ref{thm:runtime-upper-one}]
Let us denote $u = \cone[3] \sqrt{D} \ln (D K) / \eta$ for short.
First, we consider the number of iterations $\new{T_d'}$ necessary for each $\param_{d,1}$ to reach $1$\del{ for the first time}{}, i.e.,
\begin{align}
\new{T_d'} = \min \left\{ t \in \mathbb{N}_0 : \param_{d,1} = 1 \right\} \enspace.
\end{align}
Then we consider the events $\event_d$ for $d \in \intrange{1}{D}$ defined as
\begin{align}
\event_d &: \min_{s \in \llbracket 0, t \rrbracket } \param[s]_{d,1} > \frac{ 1 }{ 2 K } \enspace. \label{eq:onemax_upper_event_1}
\end{align}
We note $\event[t+1]_d$ does not occur when $\event_d$ does not.
\new{We note that $\Pr(E_d^{(\lfloor u \rfloor)})$ is same as $\Pr(T_d > u)$ provided by Lemma~\ref{thm:theta-lower-onemax}, where $T_d = \min \{ t \in \mathbb{N}_0 : \param_{d,1} \leq 1 / (2K) \}$. Then,}\del{Applying Lemma~\ref{thm:theta-lower-onemax} shows that}{} the probability of the complementary event of $\event[\lfloor u \rfloor]_d$ is bounded from above as
\begin{align}
    \Pr(\bar{E}_d^{(\lfloor u \rfloor)}) &\leq \frac{4 \lfloor u \rfloor \sqrt{D-1}}{K \eta} \exp\left( - \frac{1}{ 32 K \eta \sqrt{D-1}} \right) \\
    &\leq \frac{4 \cone[3] \sqrt{D(D-1)} \ln (DK) }{K \eta^2} \exp\left( - \frac{1}{ 32 K \eta \sqrt{D-1}} \right) \enspace, \label{eq:onemax_upper_event_1_prob_1}
\end{align}
where the last inequality is held since $\lfloor u \rfloor \leq u = \cone[3] \sqrt{D} \ln (D K) / \eta$.
Because $\cone[1] \sqrt{D}K \ln (DK) \leq \eta^{-1} \leq (DK)^{\cone[2]}$, the upper bound of \eqref{eq:onemax_upper_event_1_prob_1} is given by
\begin{align}
    \frac{4 \cone[3] \sqrt{D(D-1)} \ln (DK) }{K} (DK)^{2\cone[2]} \exp\left( - \frac{\cone[1] \sqrt{D}K \ln (DK)}{32K \sqrt{D-1}}\right) \enspace. \label{eq:onemax_upper_event_1_prob_2}
\end{align}
Since $\ln (DK) \leq \sqrt{DK}$, we can calculate the upper bound of \eqref{eq:onemax_upper_event_1_prob_2} as $4 \cone[3] D K^{-1} \sqrt{DK} (DK)^{2\cone[2]} (DK)^{-\frac{\cone[1]}{32}}$.
Thus, using $\cone[1]/32 - \new{2}\cone[2] - 3/2 = 2$ and $\cone[1]/32 - \new{2}\cone[2] + 1/2 = 4$, we obtain the inequality $\Pr(\bar{E}_d^{(\lfloor u \rfloor)}) \leq 4\cone[3] D^{-2} K^{-4}$.

Next, we consider \revdel{the}\rev{a} lower bound of \rev{the} drift of $\X_d$ in \eqref{eq:onemax-potential} conditioned on $\event_d$. Let us denote
\begin{align}
p_{d,1}^{(t)} &:= \Pr (\param[t+1]_{d,1} = \param[t]_{d,1} + \eta \mid \F[t]) \\
q_{d,1}^{(t)} &:= \Pr (\param[t+1]_{d,1} = \param[t]_{d,1} - \eta \mid \F[t]) \enspace.
\end{align}
Note that since $\eta \leq 1/(4K)$, $\param[t] > 1/(2K) \geq 2\eta$ and $\param[t+1] > 1/(2K) - \eta \geq \eta$ hold under the condition of $\event_d$. Then, the \rev{negative of the} drift of $\X[t]_d$ conditioned on $\event_d$ is derived as\del{ follows.}{}
\begin{align}
&\E [ \X[t]_d - \X[t+1]_d \mid \F[t]]  \indic[\event_d] \\
&= \E \left[ \frac{\param[t+1]_{d,1} - \param[t]_{d,1}}{\param[t]_{d,1}\param[t+1]_{d,1}} \mid \F[t] \right] \indic[\event_d] \\
&= \frac{1}{\param[t]_{d,1}}\left( \frac{ \eta }{ \param[t]_{d,1} + \eta } p_{d,1}^{(t)} - \frac{ \eta }{ \param[t]_{d,1} - \eta } q_{d,1}^{(t)} \right) \indic[\event_d] \\
&= \frac{ 1 }{ \param[t]_{d,1} ((\param[t]_{d,1})^2 - \eta^2)  } \left( \param[t]_{d,1}\E [ \param[t+1]_{d,1} - \param[t]_{d,1} | \F[t] ] - \eta^2 \Pr(x_{d,1} \neq x'_{d,1} \mid \F[t]) \right) \indic[\event_d] \enspace,
\end{align}
where we use the following relations
\begin{align}
\eta (p_{d,1}^{(t)} - q_{d,1}^{(t)}) &= \E [ \param[t+1]_{d,1} - \param[t]_{d,1} | \F[t] ] \\
p_{d,1}^{(t)} + q_{d,1}^{(t)} &= \Pr(x_{d,1} \neq x'_{d,1} \mid \F[t]) \enspace.
\end{align}
Because of \eqref{eq:onemax-theta-drift-lower} in the proof of Lemma~\ref{thm:theta-lower-onemax}, we have
\begin{align}
\frac{ \E [ \param[t+1]_{d,1} - \param[t]_{d,1} | \F[t] ] }{(\param[t]_{d,1})^2 - \eta^2 } \geq \eta \frac{2\param[t]_{d,1} (1-\param[t]_{d,1})}{4 \sqrt{D-1}} \frac{1}{(\param[t]_{d,1})^2 - \eta^2} \label{eq:onemax-cond-drift_1}
\end{align}
under $\param[t]_{d,1} > \eta$, which is satisfied if $\event_d$ occurs as $\eta \leq 1 / (4K)$. Then, \revdel{the}\rev{a} lower bound of \eqref{eq:onemax-cond-drift_1} can be calculated by
\begin{align}
    \eta \frac{2\param_{d,1} (1-\param_{d,1})}{4 \sqrt{D}} \frac{1}{(\param_{d,1})^2} = \eta \frac{\X_d}{2\sqrt{D}} \enspace.
\end{align}
Moreover, it holds $(\param_{d,1})^2 - \eta^2 \geq (\param_{d,1})^2 / 2$ when $\param_{d,1} > 1/ (2K)$\del{ since $\param_{d,1} + \eta \geq \param_{d,1}$ and $\param_{d,1} - \eta \geq \param_{d,1}/2$}{}. Then, we have
\begin{align}
\frac{ \eta^2 \Pr(x_{d,1} \neq x'_{d,1} \mid \F[t]) }{\param_{d,1} ((\param[t]_{d,1})^2 - \eta^2) } &= \frac{ 2 \eta^2 \param_{d,1} ( 1 - \param_{d,1}) }{\param_{d,1} ((\param[t]_{d,1})^2 - \eta^2) }  \leq \frac{ 2 \eta^2 \param_{d,1} ( 1 - \param_{d,1}) }{\param_{d,1} \cdot \frac{1}{2} ( \param_{d,1})^2 } = 4\eta \X_d \cdot \frac{\eta}{\param_{d,1}} \label{eq:onemax-cond-drift_2}
\end{align}
under $\event_d$. Since $\param_{d,1} \geq 1/(2K)$ under $\event_d$ and $\eta \leq 1/(\cone[1] \sqrt{D} K \ln (DK))$, \revdel{the}\rev{an} upper bound of \eqref{eq:onemax-cond-drift_2} can be given by
\begin{align}
    4 \eta \X_d \cdot 2K/(\cone[1] \sqrt{D} K \ln (DK)) \enspace.
\end{align}
Moreover, $2K/(\cone[1] \sqrt{D} K \ln (DK)) \leq 1/(16\sqrt{D})$ holds when $D \geq \exp(\new{32} / c_1) / 2$, which is satisfied when $D \geq 1$ since $\ckv[1] = 32(7/2 + \new{2}c_2) \geq \new{32} / (\ln 2)$.
Therefore, we have \revdel{the}\rev{a} lower bound of \rev{the negative of the} drift as
\begin{align}
\E [ \X[t]_d - \X[t+1]_d \mid \F[t]]  \indic[\event_d] \geq \frac{\eta}{4 \sqrt{D}} \X[t]_d \indic[\event_d] \enspace. \label{eq:expected_drift_under}
\end{align}
\new{We note $\X[0]_d = K-1$. To apply Theorem~\ref{theo:multiplicative-d}, let us define
\begin{align}
u' = \frac{4 \sqrt{D}}{\eta} \left( 2 \ln (DK) + \ln \left( \frac{ K-1 }{\eta / (1 - \eta)} \right) \right) \enspace.
\end{align}
Since $\ln ( (K - 1) (1 - \eta) / \eta ) \leq \ln (K/\eta) \leq \ln (K (DK)^{\cone[2]} ) \leq (1 + \cone[2]) \ln (DK)$\calc{
\begin{align*}
\ln \left( \frac{ K-1 }{\eta / (1 - \eta)} \right) &\leq \ln \left( \frac{ K }{\eta} \right) \leq \ln \left( D^{c_2} K^{c_2 + 1} \right) \leq (c_2 + 1) \ln \left( DK \right)
\end{align*}
}{}, $u'$ is bounded from above as
\begin{align}
u' \leq 4 \left(\cone[2] + 3 \right) \frac{ \sqrt{D} \ln (DK)}{\eta} \enspace.
\end{align}
Thus, since $\cone[3] = 4(\cone[2] + 3)$, we have $u' \leq u$ and $\Pr(\bar{E}_d^{(\lceil u' \rceil - 1)}) \leq \Pr(\bar{E}_d^{(\lfloor u \rfloor)})$.
Then, Theorem~\ref{theo:multiplicative-d} shows
\begin{align}
\Pr( T_d' \geq u' ) \leq (DK)^{- 2} + 4 \cone[3] D^{- 2} K^{- 4} \enspace, \label{eq:onemax_upper_TPd} 
\end{align}
\new{where the values in Theorem~\ref{theo:multiplicative-d} are set as $\varepsilon = \eta / (4 \sqrt{D})$ and $r = 2 \ln (DK)$.}
When $D \geq D_1$, the probability of \eqref{eq:onemax_upper_TPd} is bounded from above by $D^{-3/2}K^{-2}$.
}
Finally, we consider the probability that $T_\mathrm{max} = \max_{d \in \llbracket 1,D \rrbracket} \new{T_d'}$ is at most $u$. According to \eqref{eq:onemax_upper_TPd} and applying a union bound for $d \in \llbracket 1, D \rrbracket$, we derived \revdel{as}{}
\begin{align}
\Pr\left( T_\mathrm{max} < u \right) &\geq 1 - \sum_{d=1}^D \Pr\left( \new{T_d'} \geq u' \right) \geq 1 - D^{-\frac12} K^{-2}  \geq 1 - (DK)^{-\frac12} \enspace.
\end{align}
Since the probability of generating the optimal solution in \rev{the }$T_\mathrm{max}$-th iteration is one, we note $\Pr\left( T_\mathrm{max} < u \right) = \Pr\left( T_\mathrm{max} \leq u -1 \right) \leq \Pr\left( T_\mathrm{Hit} \leq u \right)$. \revdel{This is the end of the proof.}{}\rev{This concludes the proof.}
\end{proof}

\del{ \subsubsection{Discussion}
Theorem~\ref{thm:runtime-upper-one} shows the runtime of ccGA on COM as $O(\sqrt{D} \ln (DK) / \eta)$ with high probability when the learning rate satisfies the conditions $(\eta K)^{-1} \in \mathbb{N}$ and $\cone[1] \sqrt{D} K \ln (DK) \leq \eta^{-1} \leq (DK)^{\cone[2]}$, where the later can be expressed as $\eta \in \Omega( ( \sqrt{D} K \linebreak \ln (DK) )^{-1} )$ and $\eta \in O( (DK)^c )$ for some constant $c > 1/2$ with order notation. We note that the condition for the learning rate w.r.t. $K$ is more severe than the condition w.r.t. $D$. This implies the learning rate is recommended to be set carefully when optimizing the problem with a large number of categories. The upper tail bound of the runtime is $O(D K (\ln (DK) )^2 )$ when $\eta \in \Theta(( \sqrt{D} K \ln (DK) )^{-1} )$. Therefore, whereas the order of the learning rate in $D$ and $K$ differ, the order of the upper tail bound of the runtime $D$ and $K$ is same.

Considering the case of $K=2$, we can see that Theorem~\ref{thm:runtime-upper-one} recovers the analysis result of the cGA optimizing OneMax on the binary domain $\mathbb{K}^D_2$, which is analyzed in \cite[Theorem~5]{Sudholt:2019}. However, the target of the analysis in~\citep{Sudholt:2019} is the runtime of cGA imposed borders $\{1/D, 1-1/D\}$ on. 
In practice, their upper bound is $O(\sqrt{D} / \eta)$, which is smaller than ours with $K=2$ by a factor of $\Theta(\ln D)$. 
The analysis of the runtime which recovers this analysis result is one of our future works.
Even when $K > 2$, the behavior of ccGA on COM coincides with the behavoir of cGA on OneMax on the binary domain with the initial distribution parameter as $\param[0]_{d,1} = 1/K$ and $\param[0]_{d,2} = (K - 1) / K$. One may consider, therefore, our analysis is almost same as the analysis of cGA.
To our best knowledge, however, the runtime analysis of cGA with biased initial distribution parameter has not been performed.
This analysis result is useful since comparing the analysis of \textsc{KVal} described in Section~\ref{sec:kval} shows the relation of $K$ with the runtime may be the same between the different linear functions on the categorical domain.


When $K$ is given by the polynomial in $D$, the upper tail bound of runtime on COM is given by $O(\sqrt{D} \ln D / \eta)$ because $\ln (DK)$ becomes $O(\ln D)$. 
In contrast, when $K$ is super-polynomial in $D$, $K$ affects the upper tail bound of runtime with order notation. 
In the next subsection, we will show that the affection of $K$ is remained in the lower tail bound in such cases. }{}

\subsection{Lower Tail Bound of \rrev{the }Runtime on COM}

\subsubsection{\revdelsec{Outline and Main Analysis Result}{}\rev{Main Result and Outline of the Proof}}
In the analysis of \rev{the }lower tail bound, we focus on the dynamics of $\sum^D_{d=1} \param_{d,1}$ and consider two phases in the optimization process. 
The first phase is $\Theta(\ln K / \eta)$ iterations and the second phase is the next $\Theta( \sqrt{D} / \eta)$ iterations. In the first phase, we consider the event where $\sum^D_{d=1} \param_{d,1}$ is \revdel{not updated over}{}\rrevdel{kept at most}\rrev{never exceeds} $2D/3$ \revdel{from}\rev{with} the initial state $D/K$. Then, in the second phase, we consider the event where $\sum^D_{d=1} \param_{d,1}$ is not updated over $5D/6$ from the end of the first phase. Separating the optimization process allows us to use different potential functions and different drift theorems, which eases the difficulty of the analysis. These potential functions are designed as functions of $\sum^D_{d=1} \param_{d,1}$. Applying drift theorems twice (Theorem~\ref{theo:additive-d-lower} and Theorem~\ref{thm:general-drift}) shows that the product event of the above two events occurs with high probability. Finally, the summation of lower tail bounds $\Theta(\ln K / \eta)$ and $\Theta( \sqrt{D} / \eta)$ leads \rev{to} the lower tail bound of the runtime of ccGA on COM.
The main analysis result is given by the following theorem.

\begin{theorem} \label{thm:runtime-lower-onemax}
Consider the update of ccGA on COM.
\new{Assume $\eta := \eta(D,K)$ is given by a function of $D$ and $K$, and there are strictly positive constant values $c_1, c_2, c_3$ \revdel{which}{}\rev{that} satisfy $\cone[1] < 96 \sqrt{2}$ and $c_1\sqrt{D} \ln D \leq \eta^{-1} \leq D^{c_2} K^{c_3}$ for any $D \in \mathbb{N}$ and $K \in \mathbb{N}$}. Then, \new{given a strictly positive constant $c_4 > 0$ satisfying $1 - 6 c_3 c_4 > 0$}, 
\new{the following \revdel{is held}\rev{holds} for any $D \in \mathbb{N}$ and $K \in \mathbb{N}$ satisfying $K \geq 2$ and $\ln K \leq c_4 D$}:
\begin{align}
\Pr\left( T_{\mathrm{Hit}} \geq \frac{ c_5 (\sqrt{D} + \ln K) }{ \eta } \right) \geq 1 - 4 D^{-c_6} - K^{-c_7} \enspace, \label{eq:runtime-lower-onemax}
\end{align}
\new{where $c_5, c_6$ and $c_7$ are strictly positive constant values defined as
\begin{align}
c_5 &= \min \left\{ \frac{c_1}{2 \cdot 576}, \frac{1}{2(c_4 + 1)} \left( \frac{e(1 - 6 c_3 c_4)}{12 (c_2 + 1)}\right)^{c_2 + 1} \right\} \\
c_6 &= \min \left\{ \frac{c_1}{576 c_5}, \frac{1 - 6 c_3 c_4}{12}, \frac{1}{6} \right\} \\
c_7 &= \frac{1}{4} - 2 c_5 \enspace.
\end{align}
}
\end{theorem}

\subsubsection{Details of \rev{the }Proofs}
We first introduce two drift theorems applied twice in our derivation, a variant of general drift theorem~\rev{\cite[Theorem~3.2,~(iv)]{Variable:2021}}\revdel{\cite[Theorem~2,~(iv)]{Variable:2013}}. \rev{The statement of the general drift theorem can be found in \rrev{the} appendix.}
\revdel{
\begin{theorem}[Theorem~2,~(iv) in \citep{Variable:2013}] 
Consider a stochastic process $(\X, \F)_{t \in \mathbb{N}_0}$ over some state space $S \in \mathbb{R}$. Let $T$ be a stopping time defined as
\begin{align}
T = \min \left\{ t \in \mathbb{N} : \X \leq a \right\} \enspace.
\end{align}
for some $a \geq 0$
let $g: S \to \mathbb{R}_{\geq 0}$ be a function such that $g(0) = 0$ and $g(x) \geq g(a)$ for all $x > a$. Assume $S \cap \{ x \mid x \leq a \}$ is absorbing. Then, if there exists $\lambda > 0$ and a function $\beta_l: \mathbb{N}_0 \to \mathbb{R}_{>0}$ such that
\begin{align}
    \E \left[ e^{ \lambda ( g(\X) - g(\X[t+1])) } - \beta_l(t) \mid \F \right] \indic[{ \X \geq a}] \leq 0
\end{align}
for all $t \in \mathbb{N}_0$. Then, it holds, for $n > 0$ and $\X[0] > a$,
\begin{align}
    \Pr( T \leq n \mid \F[0] ) \leq \left( \prod^{n-1}_{t=0} \beta_l(t) \right) e^{ \lambda ( g(a) - g(\X[0])) } \enspace.
\end{align}
\end{theorem}
}


The proof of Theorem~\ref{thm:runtime-lower-onemax} is as follows.

\begin{proof}[Proof of Theorem~\ref{thm:runtime-lower-onemax}]
\new{
First, we will show that $c_5, c_6$ and $c_7$ are strictly positive. By the assumption $1 - 6 c_3 c_4 > 0$, $c_5$ and $c_6$ are obviously strictly positive. The strict positivity of $c_7$ is held since $c_5 \leq c_1 / (2 \cdot 576)$ and $c_1 < 96 \sqrt{2}$, which shows $c_5 < \sqrt{2} / 12 < 1 / 8$.

Next, we will show the inequality \eqref{eq:runtime-lower-onemax}.}
Let $u_1 := c_5 \ln K / \eta$, $u_2 := c_5 \sqrt{D} / \eta$ and $u = u_1 + u_2$ for short. We also define
\begin{align}
T_1 &= \min \left\{ t \in \mathbb{N}_0 : \sum^D_{d=1} \param_{d,1} \geq \frac{5 D}{6} \right\} \enspace\text{and} \\
T_2 &= \min \left\{ t \in \mathbb{N}_0 : \sum^D_{d=1} \param_{d,1} \geq \frac{2 D}{3} \right\} \enspace.
\end{align}
Then, Lemma~\ref{lem:general-tail-lower} leads \rev{to}
\begin{align}
&\Pr\left( T_{\mathrm{Hit}} \geq u \right) \geq 1 - \Pr( T_1 < u) - 2 (u+1) \left( \frac{5}{6} \right)^D \enspace\revdel{,}\rev{.}
\label{eq:onemax-lower-decomp}
\end{align}
In the following, we will consider upper bound\rrev{s} of \rev{the} second \revdel{,}\rev{and} third terms in \eqref{eq:onemax-lower-decomp} to derive the lower bound of $\Pr\left( T_{\mathrm{Hit}} \geq u \right)$.

First, we consider the second term in \eqref{eq:onemax-lower-decomp} by applying Theorem~\ref{theo:additive-d-upper} with the event $T_2 \leq u_1$. We introduce the stochastic process $(\X[s]_1, \F[s]_1)_{s \in \mathbb{N}_0}$ which starts \revdel{at}\rev{in} \rev{the} $u_1$-th iteration and is defined as
\begin{align}
\X[s]_1 = \sum^D_{d=1} \param[s + u_1]_{d,1} - \frac{2D}{3} \quad \text{and} \quad \F[s]_1 = \F[s + u_1] \enspace.
\end{align}
Note that the initial state $\X[0]_1$ is negative when \revdel{$T_2 < u_1$}\rev{$T_2 > u_1$}.
From Lemma~\ref{thm:Delta_bound}, we have
\begin{align}
\E[ \X[s+1]_1 - \X[s]_1 \mid \F[s]_1 ] \leq \eta \sqrt{ \frac{D}{2} } \enspace.
\end{align}
Moreover, $|\X[s+1]_1 - \X[s]_1| \leq \eta D < \sqrt{2} \eta D$ w.p. 1. We note the event $T_2 < u_1$ is $\F[s]_1$-measurable for all $s$. Then, \revdel{since $c_5 < \sqrt{2} / 12$, applying Theorem~\ref{theo:additive-d-lower} setting $\event$}\rev{setting $\event[s]$} as $T_2 < u_1$ shows
\rev{
\begin{align}
    \E[ \X[s+1]_1 - \X[s]_1 \mid \F[s]_1 ] \indic[{\event[s]}] & \leq \eta \sqrt{ \frac{D}{2} } \indic[{\event[s]}] \rrev{\quad \text{and}} \\
    |\X[s+1]_1 - \X[s]_1| \indic[{\event[s]}] & < \sqrt{2} \eta D \rrev{\enspace.}
\end{align}
Since $c_5 < \sqrt{2} / 12$, applying Theorem~\ref{theo:additive-d-lower} setting \rrevdel{$n = u_2$}\rrev{$n = u_2 = c_5 \sqrt{D} / \eta$} and $m = 5D/6 - 2D/3 = D/6$, where $n$ and $m$ are symbols in the statement of Theorem~\ref{theo:additive-d-lower}, shows
}
\rrevdel{\begin{align}
\Pr \left( T_1 < u \right) \leq D^{ - c_8 } + \Pr(T_2 \geq u_1) \enspace, 
\end{align}}
\rrev{\begin{align}
\Pr \left( T_1 < u \right) &\leq \exp \left( - \frac{ m^2 }{8 c^2 n} \right) + \Pr( \bevent[n-1] ) \\
&= \exp \left( - \frac{ (D / 6)^2 }{8 \cdot (\sqrt{2}\eta D)^2 \cdot (c_5 \sqrt{D} / \eta)} \right) + \Pr( \bevent[n-1] ) \\
&= \exp \left( - \frac{ 1 }{576c_5} \cdot \frac{1}{\eta \sqrt{D}} \right) + \Pr( \bevent[n-1] ) \\
&\leq \exp \left( - \frac{ 1 }{576c_5} \cdot \frac{c_1 \sqrt{D} \ln D}{\sqrt{D}} \right) + \Pr( \bevent[n-1] ) \\
&= D^{ - c_8 } + \Pr(T_2 \geq u_1) \enspace, \label{eq:onemax-lower-tail2}
\end{align}}
where $c_8 = {c_1} / ({576 c_5})$. 

Next, we will derive \revdel{the}\rev{an} upper bound of $\Pr(T_2 \geq u_1)$ by applying Theorem~\ref{thm:general-drift}.
We introduce the stochastic process $\X_2$ for $t \leq u_1$ defined as
\begin{align}
\X_2 &= - \ln \left( \sum^D_{d=1} \param_{d,1} \right) + \ln\left( \frac{2D}{3} \right) \enspace.
\end{align} 
Note that $\X[0]_2 = \ln (2K / 3)$ is positive. To apply Theorem~\ref{thm:general-drift}, we consider the expectation of $\exp\left( \X_2 - \X[t+1]_2 \right)$. We have
\begin{align}
\exp\left( \X_2 - \X[t+1]_2 \right) &= \exp\left( \ln \left( \sum^D_{d=1} \param[t+1]_{d,1} \right) - \ln \left( \sum^D_{d=1} \param_{d,1} \right) \right) \\
&= 1 + \frac{ \sum^D_{d=1} ( \param[t+1]_{d,1} - \param_{d,1} ) }{ \sum^D_{d=1} \param_{d,1} } \enspace.
\end{align}
\new{Note that
\begin{multline}
\param[t+1]_{d,1} - \param_{d,1} = \eta \left( \indic[{ \param[t+1]_{d,1} > \param_{d,1} }] - \indic[{ \param[t+1]_{d,1} < \param_{d,1} }] \right) \\
\leq \eta \indic[{ \param[t+1]_{d,1} \neq \param_{d,1} }] = \eta \indic[{ x_{d,1} \neq x'_{d,1} }] \enspace.
\end{multline}
}
Therefore, the conditional expectation of $\param[t+1]_{d,1} - \param_{d,1}$ w.r.t. $\F$ is bounded from above by the probability $\Pr( x_{d,1} \neq x'_{d,1} \mid \F) = 2 \param_{d,1} (1 -  \param_{d,1})$ multiplied by $\eta$. Then, we have
\begin{align}
\E\left[ \sum^D_{d=1} \param[t+1]_{d,1} - \param_{d,1} \mid \F \right] &\leq 2 \eta \sum^D_{d=1} \param_{d,1} (1 -  \param_{d,1}) \leq 2 \eta \sum^D_{d=1} \param_{d,1} \enspace.
\end{align}
Therefore, we have
\begin{align}
\E\left[ \exp\left( \X_2 - \X[t+1]_2 \right) \mid \F \right] \leq 1 + 2 \eta \enspace.
\end{align}
Moreover,
\begin{align}
\exp\left( - \left( \X[0]_2 - 0 \right) \right) = \frac{3}{2K} \enspace.
\end{align}
Furthermore, considering the update rule, \revdel{$\X$}\rev{$\X_2$} is monotonically decreasing on \del{\textsc{OneMax}}{}\new{COM} because 
\begin{align}
\sum^D_{d=1} \param[t+1]_{d,1} - \sum^D_{d=1} \param_{d,1} &= \eta \left( \sum^D_{d=1} \sample[1]_{d,1} - \sum^D_{d=1} \sample[2]_{d,1} \right) = \eta \left( f(\sample[1]) - f(\sample[2]) \right) \geq 0 \enspace,
\end{align}
which means that the states $\{ \X_2 \leq 0 \}$ \revdel{is}\rev{are} absorbing. Therefore, applying Theorem~\ref{thm:general-drift} with the variables $a = 0, \lambda = 1$ and the function $g(x) = x$ leads \rev{to}
\begin{align}
\Pr( T_2 \geq u_1) &= \Pr( \X[u_1]_2 \leq 0 ) \leq \left(1 + 2\eta \right)^\frac{c_5 \ln K}{\eta} \cdot \frac{3}{2 K} \leq K^{2 c_5 } \cdot \frac{3}{2 K} \enspace.
\end{align}
Since $3/2 < 2^{3/4}\leq K^{3/4}$, we have
\begin{align}
\Pr( T_2 \geq u_1) \leq K^{2 c_5 - \frac{1}{4}} = K^{-c_7} \enspace. \label{eq:onemax-lower-tail1}
\end{align}

Finally, we will consider the upper bound of the third term in \eqref{eq:onemax-lower-decomp}. To apply Lemma~\ref{lem:general-tail-lower}, we consider an upper bound of $2 (u_1 + u_2 + 1) (5/6)^D$ as
\begin{align}
&2\left( \frac{\revdel{2} c_5 (\sqrt{D} + \ln K)}{\eta} + 1 \right) \left( \frac{5}{6} \right)^D \notag\\
&\qquad \leq 2 c_5 (\sqrt{D} + \ln K) D^{c_2} K^{c_3} \left( \frac{5}{6} \right)^D + 2 \left( \frac{5}{6} \right)^D \\
&\qquad \leq 2 c_5 (c_4 + 1) D^{c_2 + 1} \exp\left( c_3 c_4 D \right) \left( 1 - \frac{1}{6} \right)^D + 2 \left( 1 - \frac{1}{6} \right)^D \\
&\qquad \leq \exp \left( \ln \left( 2 c_5 (c_4 + 1) \right) + (1 + c_2) \ln D - \left( \frac{1}{6} - c_3 c_4 \right) D \right) + 2 \exp\left( - \frac{D}{6} \right) \\
&\qquad \leq \exp \left( c_{9} + c_{10} \ln D - c_{11} D \right) + 2 \exp\left( - \frac{D}{6} \right) \enspace, 
\end{align}
where $c_{9} = \ln \left( 2 c_5 (c_4 + 1) \right), c_{10} = c_2 + 1$ and $c_{11} = 1/6 - c_3 c_4$. Note that $c_{9}, c_{10}$ and $c_{11}$ are positive. 
Since the maximizer of $g(x) = c_{9} + c_{10} \ln x - c_{11} x / 2$ is given by $x = 2 c_{10} / c_{11}$,
an elementary calculation shows that $c_{9} + c_{10} \ln D - c_{11} D / 2$ is negative for all $D$ when 
\begin{align}
  c_5 \leq \frac{1}{2(c_4 + 1)} \left( \frac{e(1 - 6 c_3 c_4)}{12 (c_2 + 1)}\right)^{c_2 + 1} \enspace. \label{eq:onemax-lower-const-upper}
\end{align}
Thus, we have
\begin{align}
c_{9} + c_{10} \ln D - c_{11} D \leq - \frac{c_{11}}{2} D \label{eq:onemax-lower-tail3-coef}
\end{align}
for all $D$.
\calc{
We denote
\begin{align*}
g(x) := c_{9} + c_{10} \ln x - \frac{c_{11}}{2} x \enspace.
\end{align*}
The differential of $g(x)$ is calculated as follows.
\begin{align*}
g'(x) = \frac{c_{10}}{x} - \frac{c_{11}}{2}
\end{align*}
When $x=2c_{10}/c_{11}$, $g(x)$ is maximized.
\begin{align*}
g\left( \frac{2c_{10}}{c_{11}} \right) &= c_{9} + c_{10} \ln \left( \frac{2c_{10}}{c_{11}} \right) - \frac{c_{11}}{2} \frac{2c_{10}}{c_{11}} \\
&= c_{9} - c_{10} \ln \left( \frac{e c_{11}}{2 c_{10}}\right) \\
&= \ln (2 c_5 (c_4 + 1)) - \ln \left( \frac{e c_{11}}{2 c_{10}}\right)^{c_{10}}
\end{align*}
Then, when
\begin{align*}
c_5 \leq \frac{1}{2(c_4 + 1)} \left( \frac{e c_{11}}{2 c_{10}}\right)^{c_{10}} \enspace, 
\end{align*}
$g(x)$ is always negative for all $x > 0$.
}{}
As $D \geq \ln D$, 
we obtain a lower bound of $\Pr\left( T_{\mathrm{Hit}} \geq u \right)$ from \eqref{eq:onemax-lower-decomp}, \eqref{eq:onemax-lower-tail2}, \eqref{eq:onemax-lower-tail1} and \eqref{eq:onemax-lower-tail3-coef} as
\begin{align}
\Pr\left( T_{\mathrm{Hit}} \geq u \right) & \geq 1 - K^{-c_7} - D^{-c_8} - \exp \left( -\frac{c_{11}}{2} D \right) - 2 \exp\left( - \frac{D}{6} \right)\\
& \geq 1 - K^{-c_7} - D^{-c_8} - D^{-\frac{c_{11}}{2}} - 2 D^{-\frac{1}{6}} \\
&\geq 1 - K^{-c_7} - 4 D^{-c_{6}} \enspace.
\end{align}
\revdel{This is the end of the proof.}{}\rev{This concludes the proof.}
\end{proof}

\del{ \subsubsection{Discussion}
From Theorem~\ref{thm:runtime-upper-one} and Theorem~\ref{thm:runtime-lower-onemax}, the runtime of ccGA on \textsc{OneMax} is shown as $\Theta(\ln K / \eta)$ w.r.t. $K$. When $K$ is given by \revdel{the}\rev{a} polynomial in $D$, the upper bound and the lower bound are respectively $O(\sqrt{D}\ln D / \eta)$ and $\Omega(\sqrt{D} / \eta)$ since $\ln (DK)$ and $\sqrt{D} + \ln K$ become $O(\ln D)$ and $\Omega(\sqrt{D})$, respectively. 
Comparing the upper bound and the lower bound w.r.t. $D$, there is a difference of $\Theta(\ln D)$. As discussed in Theorem~\ref{thm:runtime-upper-one}, there is room for improvement for the analysis of the upper bound, when compared to the analysis of cGA~\citep{Sudholt:2019}. 

In addition, by taking $K=2$ for Theorem~\ref{thm:runtime-lower-onemax}, our result also applies to the cGA optimizing OneMax on the binary domain, which is analyzed in \cite[Theorem~8]{Sudholt:2019}. However, our result does not provide the upper bound of $\eta$ for optimizing \textsc{OneMax} successfully. This analysis is left as future work. }{}

\subsection{Discussion}
The conditions of the learning rate in both of Theorem~\ref{thm:runtime-upper-one} and Theorem~\ref{thm:runtime-lower-onemax} are satisfied when $(\eta K)^{-1} \in \mathbb{N}$ and
\begin{align}
    c \sqrt{D} K \ln (DK) \leq \eta^{-1} \leq (DK)^{c'} \label{eq:one-lr-cond}
\end{align}
for properly chosen constants $c$ and $c'$. We note that the condition for the learning rate w.r.t. $K$ is more severe than the condition w.r.t. $D$. \del{This implies the learning rate is recommended to be set carefully when optimizing the problem with a large number of categories.}{}\new{This implies that, for the problems in a categorical domain, the order notations of the suitable learning rate settings in $D$ and $K$ may be different.} With the condition~\eqref{eq:one-lr-cond}, Theorem~\ref{thm:runtime-upper-one} and Theorem~\ref{thm:runtime-lower-onemax} show that the runtime of ccGA on COM is \rrev{with high probability} $O(\sqrt{D} \ln (DK) / \eta)$ and $\Omega( (\sqrt{D} + \ln K) / \eta)$, respectively. The upper and lower tail bounds of the runtime become $O(D K (\ln (DK) )^2 )$ and $\Omega(( \sqrt{D} + \ln K) \sqrt{D} K \ln (DK) )$ when the learning rate is set as large as possible satisfying~\eqref{eq:one-lr-cond}, i.e., $\eta^{-1} \in \Theta( \sqrt{D} K \ln (DK) )$.

Considering the case of $K=2$, Theorem~\ref{thm:runtime-upper-one} recovers the analysis result of the cGA optimizing OneMax on the binary domain $\mathbb{K}^D_2$, which is analyzed in \cite[Theorem~5]{Sudholt:2019}. 
However, the target \rev{algorithm} of the analysis in~\citep{Sudholt:2019} is the runtime of cGA \revdel{imposed borders $\{1/D, 1-1/D\}$ on}\rev{with borders $\{1/D, 1-1/D\}$}. 
\del{In practice}{}\new{Unfortunately}, the upper bound in \citep{Sudholt:2019} is $O(\sqrt{D} / \eta)$, which is smaller than ours with $K=2$ by a factor of $\Theta(\ln D)$. 
The analysis of the runtime which recovers this analysis result is one of our future works. Considering the lower tail bound in Theorem~\ref{thm:runtime-lower-onemax} with $K=2$, our result is consistent with the lower tail bound of the first hitting time of the cGA optimizing OneMax on the binary domain, which is derived in \cite[Theorem~8]{Sudholt:2019}.

\del{
When $K$ is given by the polynomial in $D$, both the upper and lower tail bounds of runtime on COM are given by $\Theta(\sqrt{D} \ln D / \eta)$ because $\ln K \in \Theta(1)$. 
The difference between the upper and lower tail bound appears in order notation when $K$ is superpolynomial in $D$. When $\ln K \in \Theta(D^c)$ for some $c > 1/2$, the upper tail bound is $O(D^{c + 1/2} \ln D / \eta)$ while the lower tail bound is $\Omega(D^{c} \ln D / \eta)$.
}{}
\new{
When $K$ is given by \revdel{the}\rev{a} polynomial in $D$, the upper and lower tail bounds of \rev{the }runtime on COM are given by $\Theta(\sqrt{D} \ln D / \eta)$ and $\Theta(\sqrt{D} / \eta)$, respectively, in which the ratio is given by $\Theta(\ln D)$. In contrast, \revdel{When}\rev{when} $\ln K \in \Theta(D^c)$ for some $c > 1/2$, the upper tail bound is $O(D^{c + 1/2} \ln D / \eta)$ while the lower tail bound is $\Omega(D^{c} \ln D / \eta)$, in which the ratio is given by $\Theta(\sqrt{D})$. Therefore, the difference between the upper and lower tail bounds becomes large in order notation when $K$ is super-polynomial in $D$.
}

Even when $K > 2$, the behavior of ccGA on COM coincides with the behavior of cGA on OneMax on the binary domain with a biased initial distribution parameter as $\param[0]_{d,1} = 1/K$ and $\param[0]_{d,2} = (K - 1) / K$. One may consider, therefore, our analysis is almost the same as the analysis of cGA.
To the best of our knowledge, the runtime analysis of cGA with a biased initial distribution parameter has not been performed.
This analysis result is useful since \rrevdel{comparing}\rrev{a comparison with} the analysis of \textsc{KVal} described in Section~\ref{sec:kval} shows the \rrevdel{relation of $K$ with the runtime}\rrev{relationship between $K$ and the runtime} may be the same between the different linear functions on the categorical domain.

\section{Runtime Analysis on \textsc{KVal}} \label{sec:kval}

In the binary domain, the dependency of the number of dimensions $D$ on the runtime differs even among the linear functions~\citep{Droste:2006}. 
In the analyses of binary linear functions, BinVal is treated as a representative example of the comparatively difficult linear function and the first hitting time on BinVal is \revdel{widely }investigated~\citep{Droste:2006,Domino:2018}. 
In this section, we investigate an extension of BinVal to the categorical domain, which is termed \textsc{KVal} and defined as
\begin{align}
\KVAL(x) = \sum_{d=1}^D \sum_{k=1}^K (K-k) K^{D-d} x_{d,k} \enspace.
\end{align}
The same as BinVal, \textsc{KVal} interprets the input categorical variable as the representation of the function value $\KVAL(x)$ in base $K$.
\rev{In contrast to COM, which counts the number of the first categories, all categories affect the evaluation value differently on \textsc{KVal}.}
In comparison with the analysis on \del{\textsc{OneMax}}{}\new{COM}, the analysis of the runtime on \textsc{KVal} shows us how the dependence of $K$ on the search efficiency differs among the linear functions on the categorical domain.

\begin{figure}
  \begin{center}
    \includegraphics[width=\linewidth]{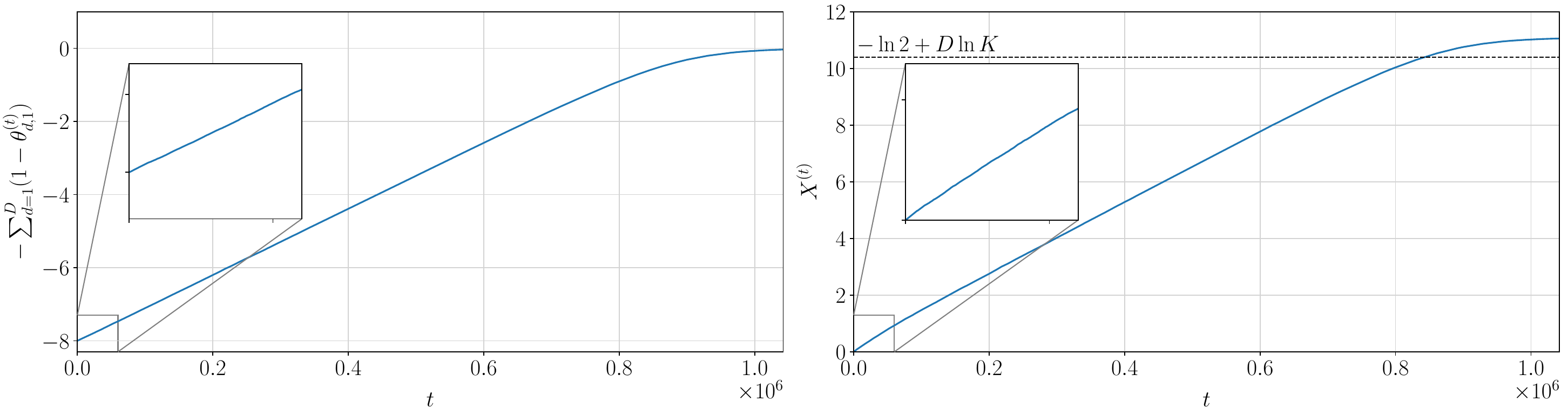}
    \vspace{-4mm}
    \caption{The \rrevdel{\rrev{values} of the negative of }\rrev{negative }potential \rrev{values} used in the analysis of cGA~\citep{Domino:2018}, $- \sum_{d=1}^D (1 - \param[t]_{d,1})$~(left), and \rrev{our potential values }$\X$ in \eqref{eq:kval-potential}~(right) \rrev{in one typical trial of optimizing} \textsc{KVal} with $D=16$ and $K=2$.}
    \label{fig:potential_kval_2k}
  \end{center}
\end{figure}

\begin{figure}
  \begin{center}
    \includegraphics[width=\linewidth]{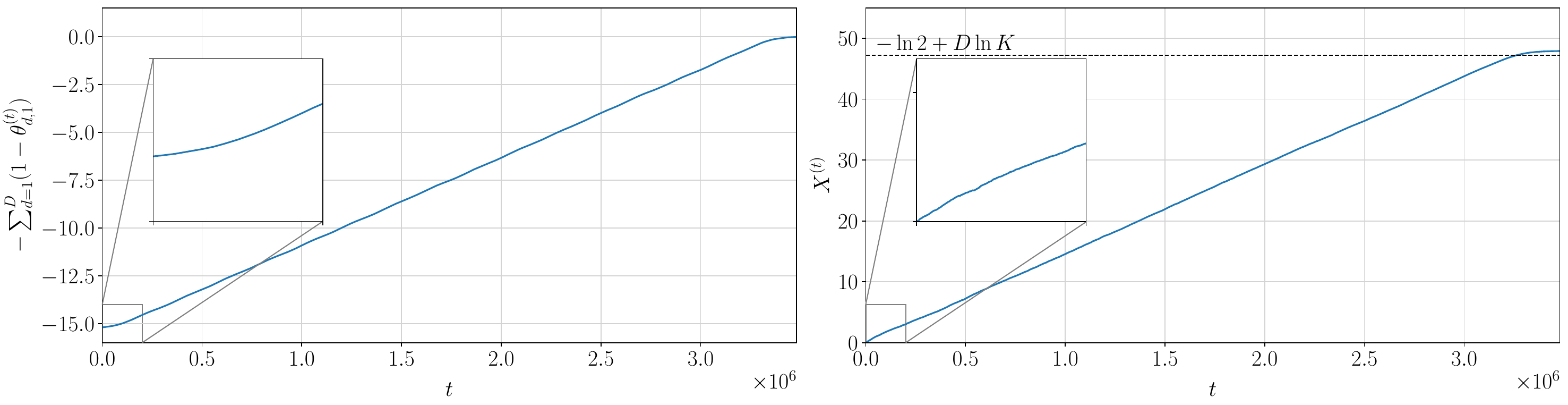}
    \vspace{-4mm}
    \caption{The \rrevdel{\rrev{values} of the negative of }\rrev{negative }potential \rrev{values }used in the analysis of cGA~\citep{Domino:2018}, $- \sum_{d=1}^D (1 - \param[t]_{d,1})$~(left), and \rrev{our potential values }$\X$ in \eqref{eq:kval-potential}~(right) \rrev{in one typical trial of optimizing} \textsc{KVal} with $D=16$ and $K=20$.}
    \label{fig:potential_kval}
  \end{center}
\end{figure}

\subsection{Upper Tail Bound of \rrev{the }Runtime on \textsc{KVal}}
\subsubsection{\revdelsec{Outline and Main Analysis Result}\rev{Main Result and Outline of the Proof}}

We consider a potential function specifically for the runtime analysis on \textsc{KVal}, which \revdel{relieves}\rev{reveals} the difficulty of our analysis as well as the runtime analysis on COM in Section~\ref{sec:onemax}.
For the runtime analysis of cGA on BinVal, \cite{Domino:2018} considers the potential function $\sum_{d=1}^D (1 - \param[t]_{d})$, where $\param[t]_d$ is the $d$-th element of distribution parameter of $D$-dimensional Bernoulli distribution, i.e., $\param[t]_d = \Pr( x_d = 1 )$. However, this potential function is intractable for the analysis of ccGA, especially in the early stage of the optimization, as well as the runtime analysis on COM.
Instead, we introduce the potential function $\X$ defined as
\begin{align}
&\X = \sum^D_{d=1} \X_d + D \ln K \enspace \label{eq:kval-potential}  \quad \text{where} \quad
\X[t]_d =
\begin{cases}
\ln \eta &\enspace \text{if} \enspace 0 \leq \param[t]_{d,1} < \eta \\
\ln \param[t]_{d,1} &\enspace \text{if} \enspace \eta \leq \param[t]_{d,1} \leq 1
\end{cases}
\enspace.
\end{align}
We note the initial state $\X[0]$ is given by zero. We consider the case $0 \leq \param[t]_{d,1} < \eta$ separately to prevent the potential \revdel{goes negative infinity}\rev{taking undefined value}\rrev{s}, similar to the potential function for COM. Otherwise, the potential function value is determined by the sum of $\ln \param[t]_{d,1} $ for all $d \in \llbracket 1, D \rrbracket$.

\del{
Figure~\ref{fig:potential_kval} shows the transitions of the negative of the potential in~\citep{Domino:2018}, $- \sum^D_{d=1} (1 - \param[t]_{d,1})$, and $\X$ on \textsc{KVal} with $D=16$ and $K=20$. Focusing the transition of $- \sum^D_{d=1} (1 - \param[t]_{d,1})$, it increases slowly in the early stage of optimization. On the other hand, our potential $\X$ increases at almost the same rate until it becomes above $-\ln 2 + D \ln K$. Similar to the potential function used in the analysis of COM, this potential function is practical when $K$ is large, as seen from the simulation result for $K=2$ in Figure~\ref{fig:potential_kval_2k}.
}{}
\new{
As \revdel{well as}\rev{for} the potential for COM, we demonstrate the importance of our potential function $\X$ by comparing the negative of a potential function $- \sum^D_{d=1} (1 - \param[t]_{d,1})$ used in the analysis of cGA on BinVal~\citep{Domino:2018}. We show the \rrevdel{transitions of the}\rrev{negative} potential function\rrevdel{s} \rrev{values} $- \sum^D_{d=1} (1 - \param[t]_{d,1})$ and \rrev{our potential values} $\X$ on \textsc{KVal} with $D=16$ and $K=2$ in Figure~\ref{fig:potential_kval_2k}, and the \rrevdel{transitions}\rrev{values} with $D=16$ and $K=20$ in Figure~\ref{fig:potential_kval}. 
\rev{The learning rate was set as $\eta^{-1} = \lceil D K \ln K \ln (DK) \rceil K$ as used in the experiment in Section~\ref{sec:experiment}.}
The \rrevdel{shapes of the transition curves}\rrev{changes of two potential values} are similar when $K=2$, which means the dynamics of cGA on BinVal. In contrast, the dynamics of $- \sum^D_{d=1} (1 - \param[t]_{d,1})$ and $\X$ differ in early stage of the optimization when $K=20$. Our potential $\X$ increases at almost the same rate until it \revdel{becomes}\rev{rises} above $-\ln 2 + D \ln K$, \rrevdel{as well as}\rrev{as in} the case of $K=2$, while $- \sum^D_{d=1} (1 - \param[t]_{d,1})$ increases slowly. As we discussed with the potential for COM, potential functions whose \revdel{the }shape of the \rrevdel{transitions}\rrev{changes} with different settings of $K$ are almost the same are tractable for our analysis. We construct our potential function $\X$ to satisfy this requirement.
}

The outline of the analysis is as follows. First, we show that all of $\param_{1,1}, \cdots, \param_{D,1}$ rarely reach below $1 / (2K)$ when the learning rate is \revdel{smaller}\rev{small} enough in Lemma~\ref{lem:frequency} to deal with the genetic drift. Next, we derive the lower bound $\eta / 8$ of the conditional drift of our potential function under the event $\param_{d,1} > 1 / (2K)$ for all $d \in \intrange{1}{D}$. Then, applying Theorem~\ref{theo:additive-d-upper} shows the upper tail bound of the iteration where $\X$ reaches $- \ln 2 + D \ln K$ for the first time, and applying Lemma~\ref{lem:general-tail-lower} shows the upper tail bound of the $T_\mathrm{Hit}$. The main analysis result is given by the next theorem.

\begin{theorem} \label{thm:runtime-upper-kval}
Consider the update of ccGA on \textsc{KVal}. \new{Assume $\eta := \eta(D,K)$ is given by a function of $D$ and $K$ fulfilling $(\eta K)^{-1} \in \mathbb{N}$ and $\cone[1] D K^2 \ln K \ln (DK) \leq \eta^{-1}$ for any $D \in \mathbb{N}$ and $K \in \mathbb{N}$ with $\cone[1] = 512$.}
\new{Then, the following \revdel{is held}\rev{holds} for all $D \geq 5$ and $K \geq 2$: For the runtime $T_\mathrm{Hit}$,}
\begin{align}
\Pr\left( T_\mathrm{Hit} \leq \frac{ \cone[2] D \ln K }{ \eta } \right) > 1 - (DK)^{- \cone[3]}  \enspace, \label{eq:runtime-upper-kval}
\end{align}
where $\cone[2] = 32$ and $\cone[3] = 1/2$.
\end{theorem}

\del{
\subsubsection{Proof Sketch}
We firstly introduce brief sketch of our derivation for the upper tail bound of the runtime of ccGA on \textsc{KVal}.
The basic idea is similar to the runtime analysis of cGA on BinVal in \citep{Domino:2018}. 
Before showing the upper tail bound of the runtime, in Lemma~\ref{lem:frequency}, we will give an upper bound of the probability that $\param_{d,1}$ reaches below $1 / (2K)$ within a given number of steps for some $d \in \llbracket 1, D \rrbracket$. 
This will be proved by the fact that $\param_{d,1}$ is sub-martingale, and applying~\cite[Corollary 4.1]{Wormald:1999} with $b=0$.
Lemma~\ref{lem:frequency} shows that $\param_{d,1}$ is maintained above $1 / (2K)$ within $O(D\ln K / \eta)$ steps with high probability when using a small learning rate satisfying $\eta^{-1} \in O(D K^2 \ln K \ln (DK))$.

Using Lemma~\ref{lem:frequency}, we will show that the runtime of ccGA on \textsc{KVal} is $O(D\ln K / \eta)$ in Theorem~\ref{thm:runtime-upper-kval}. 
In the proof of Theorem~\ref{thm:runtime-upper-kval}, we consider the number of iterations $T_1$ necessary for the potential function $\X[t]$ to reach $- \ln 2 + D \ln K$ for the first time.
To derive the upper tail bound of $T_1$, we will show the conditional drift of $\X[t]$ is $\Omega(\eta)$. Then, the conditional drift theorem shows, when $s \in O(D \ln K / \eta)$, $\Pr(T_1 \geq s)$ is significantly small.
Finally, applying Lemma~\ref{lem:general-tail-upper} with $u \in O((\eta D)^{-1})$ will shows the first hitting time is less than $s + u \in O(D \ln K / \eta)$ with high probability.
}{}

\subsubsection{Details of \rev{the }Proofs}
We first investigate the smallest value of $\param_{d,1}$ in the optimization of \textsc{KVal}. The following lemma shows the smallest value of $\param_{d,1}$ is greater than $1 / (2K)$ with high probability when the learning rate is \revdel{smaller}\rev{small} enough.

\begin{lemma} \label{lem:frequency}
Consider the update of \target on \textsc{KVal}. Let $T_1 = \min\{ t \in \mathbb{N}_0 : {}^\exists d \in \llbracket 1, D \rrbracket, \param_{d, 1} \leq 1 / (2K) \}$. Then, for $s \in \mathbb{N}$,
\begin{align}
\Pr( T_1 \leq s ) \leq D \exp \left( - \frac{ 1 }{ 16 K^2 \eta^2 s }  \right) \enspace.
\end{align}
\end{lemma}

\del{ 
The brief proof sketch of Lemma~\ref{lem:frequency} is as follows. 
Focusing an index $d \in \intrange{1}{D}$, we consider three cases as 1) there is $i < d$ such that the $i$-th categories in $x$ and $x$' differ 2) all of the first $d-1$ categories of $x$ and $x'$ agree and the $d$-th categories are different, 3) all of the first $d$ categories of $x$ and $x'$ agree.
Considering them individually, we show $\param_{d,1}$ does not move in expectation in the first and second cases and it increases by $\eta$ in the third case. 
To sum up, we obtain the drift of $\param_{d,1}$ as
\begin{align}
& \E[\param[t+1]_{d,1} - \param_{d,1} \mid \F ] = 2 \eta \param_{d,1} (1 - \param_{d,1}) \prod_{j=1}^{d-1} \left( \sum_{k=1}^K (\param_{j,k})^2 \right) \enspace.
\end{align}
Since $\param_{d,1}$ is super-martingale, applying~\cite[Corollary 4.1]{Wormald:1999} with $b=0$ shows the lower tail bound of $\param_{d,1}$. 
Finally, considering all indexes in $\intrange{1}{D}$, $\param_{d,1}$ finishes the proof.
}{}

\begin{proof}[Proof of Lemma~\ref{lem:frequency}]

First, we will derive 
\begin{align}
& \E[\param[t+1]_{d,1} - \param_{d,1} \mid \F ] = 2 \eta \param_{d,1} (1 - \param_{d,1}) \prod_{j=1}^{d-1} \left( \sum_{k=1}^K (\param_{j,k})^2 \right) \revdel{\enspace.}
\label{eq:exp-theta-diff-kval}
\end{align}
and show that $\param_{d,1}$ is \rev{a }sub-martingale. We note
\begin{align}
\E[\param[t+1]_{d,1} - \param[t]_{d,1} \mid \F] = \eta \E[\sample[1]_{d,1} - \sample[2]_{d,1} \mid \F] \enspace. \label{eq:kval_exp_theta_diff}
\end{align}
The difference $\sample[1]_{d,1} - \sample[2]_{d,1}$ can be transformed as
\begin{align}
\sample[1]_{d,1} - \sample[2]_{d,1} = (x_{d,1} - x'_{d,1}) \indic[\KVAL(x) \geq \KVAL(x') ] - (x_{d,1} - x'_{d,1}) \indic[\KVAL(x) < \KVAL(x') ] \enspace. \label{eq:kval-lema-decompo}
\end{align}
\del{
Moreover, the first term in \eqref{eq:kval-lema-decompo} can be transformed as
\begin{align}
& (x_{d,1} - x'_{d,1}) \indic[\KVAL(x) - \KVAL(x') \geq 0] \label{eq:kval_diff_geq_start} \\
\begin{split}
& \quad = (x_{d,1} - x'_{d,1}) \indic[\KVAL(x) - \KVAL(x') \geq 0] \\
& \quad \qquad \cdot \left( \prod_{i=1}^{D} (\indic[x_i = x'_i] + \indic[x_i \neq x'_i]) \right)
\end{split} \\
\begin{split}
& \quad = (x_{d,1} - x'_{d,1}) \indic[\KVAL(x) - \KVAL(x') \geq 0] \\
& \quad \qquad \cdot \left( \sum_{i=1}^D \indic[x_i \neq x'_i ] \prod_{j=1}^{i-1} \indic[x_{j} = x'_{j} ] + \prod_{j=1}^D \indic[x_{j} = x'_{j}] \right) \enspace. 
\end{split} \label{eq:kval_diff_samples1}
\end{align}
Since $(x_{d,1} - x'_{d,1}) \indic[x_d = x'_d] = 0$, \eqref{eq:kval_diff_samples1} is calculated as follows.
\begin{align}
(x_{d,1} - x'_{d,1}) \indic[\KVAL(x) - \KVAL(x') \geq 0] \sum_{i=1}^d \indic[x_i \neq x'_i ] \prod_{j=1}^{i-1} \indic[x_{j} = x'_{j} ] \label{eq:kval_diff_samples2}
\end{align}
Here, we define $\Delta_i^<$, $\delta_i$ and $\Delta_i^>$ for all $i \in \intrange{1}{D}$ as 
\begin{align}
\Delta_i^< &= \sum_{i'=1}^{i-1} \sum^K_{k=1} (K-k) K^{D-i'} (x_{i',k} - x'_{i',k}) \\
\delta_i &= \sum^K_{k=1} (K-k) K^{D-i} (x_{i,k} - x'_{i,k}) \label{eq:kval-delta} \\
\Delta_i^> &= \sum_{i'=i+1}^D \sum^K_{k=1} (K-k) K^{D-i'} (x_{i',k} - x'_{i',k}) \enspace. \label{eq:kval-Delta}
\end{align}
Note that $\KVAL(x) - \KVAL(x') = \Delta_i^< + \delta_i + \Delta_i^>$ for all $i \in \intrange{1}{D}$ and, if $\prod_{j=1}^{d-1} \indic[x_j = x'_j] = 1$, it holds $\Delta_d^< = 0$. Moreover, if $\prod_{j=1}^{d} \indic[x_j = x'_j] = 1$, it holds $x_{d,1} = x_{d,1}'$. Therefore, \eqref{eq:kval_diff_samples2} can be transformed as follows.
\begin{align}
(x_{d,1} - x'_{d,1}) \sum_{i=1}^d \indic[ \delta_i + \Delta_i^> \geq 0] \indic[x_i \neq x'_i ] \prod_{j=1}^{i-1} \indic[x_{j} = x'_{j} ] \label{eq:kval_diff_samples_geq}
\end{align}
Similar to the transformations in \eqref{eq:kval_diff_geq_start}-\eqref{eq:kval_diff_samples_geq}, the second term in \eqref{eq:kval-lema-decompo} can be transformed as
\begin{align}
& (x_{d,1} - x'_{d,1}) \indic[\KVAL(x) - \KVAL(x') < 0] \notag \\
& \quad = (x_{d,1} - x'_{d,1}) \sum_{i=1}^d \indic[ \delta_i + \Delta_i^> < 0] \indic[x_i \neq x'_i ] \prod_{j=1}^{i-1} \indic[x_{j} = x'_{j} ] \enspace. \label{eq:kval_diff_samples_leq}
\end{align}
By the definition of $\delta_{i}$ and $\Delta_{i}^>$, we have
\begin{align}
\delta_{i} &\in \left\{ n K^{D-i} : n \in \intrange{-(K-1)}{K-1} \right\} \enspace, \\
|\Delta_{i}^>| &\leq K^{D-i} - 1 \enspace.
\end{align}
As $|\Delta_{i}^>| < | \delta_{i} |$ when $\delta_{i} \neq 0$, it establishes
\begin{align}
\indic[\delta_i + \Delta_i^> \geq 0] &= \indic[\delta_i > 0] + \indic[\delta_i = 0] \indic[\Delta_i^> \geq 0] \\
\indic[\delta_i + \Delta_i^> < 0] &= \indic[\delta_i < 0] + \indic[\delta_i = 0] \indic[\Delta_i^> < 0] \enspace.
\end{align}
Therefore, since $\indic[x_i \neq x'_i] = \indic[\delta_i \neq 0]$, we obtain
\begin{align}
&\indic[\delta_i + \Delta_i^> \geq 0] \indic[x_i \neq x'_i] \notag\\
&\qquad = (\indic[\delta_i > 0] + \indic[\delta_i = 0] \indic[\Delta_i^> \geq 0]) \indic[x_i \neq x'_i] \notag \\
&\qquad = \indic[\delta_i > 0] \enspace, \\
&\indic[\delta_i + \Delta_i^> < 0] \indic[x_i \neq x'_i] \notag \\
&\qquad = (\indic[\delta_i < 0] + \indic[\delta_i = 0] \indic[\Delta_i^> < 0]) \indic[x_i \neq x'_i] \notag \\
&\qquad = \indic[\delta_i < 0] \enspace.
\end{align}
Then, according to \eqref{eq:kval_diff_samples_geq} and \eqref{eq:kval_diff_samples_leq}, we can transform each term in \eqref{eq:kval-lema-decompo} as
\begin{align} 
& (x_{d,1} - x'_{d,1}) \indic[\KVAL(x) - \KVAL(x') \geq 0] \notag \\
\begin{split}
& \qquad = (x_{d,1} - x'_{d,1}) \left( \sum_{i=1}^{d-1} \indic[\delta_i > 0] \prod_{j=1}^{i-1} \indic[x_j = x'_j] \right) \\
& \qquad \qquad + (x_{d,1} - x'_{d,1}) \indic[\delta_d > 0] \prod_{j=1}^{d-1} \indic[x_j = x'_j]
\end{split} \\
\begin{split}
& \qquad = (x_{d,1} - x'_{d,1}) \left( \sum_{i=1}^{d-1} \indic[\delta_i > 0] \prod_{j=1}^{i-1} \indic[x_j = x'_j] \right) \\
& \qquad \qquad + \indic[x_{d,1} = 1] \indic[x'_{d,1} = 0] \prod_{j=1}^{d-1} \indic[x_j = x'_j]
\end{split} \enspace, \\
& (x_{d,1} - x'_{d,1}) \indic[\KVAL(x) - \KVAL(x') < 0] \notag \\
\begin{split}
& \qquad = (x_{d,1} - x'_{d,1}) \left( \sum_{i=1}^{d-1} \indic[\delta_i < 0] \prod_{j=1}^{i-1} \indic[x_j = x'_j] \right) \\
& \qquad \qquad + (x_{d,1} - x'_{d,1}) \indic[\delta_d < 0] \prod_{j=1}^{d-1} \indic[x_j = x'_j]
\end{split} \\
\begin{split}
& \qquad = (x_{d,1} - x'_{d,1}) \left( \sum_{i=1}^{d-1} \indic[\delta_i < 0] \prod_{j=1}^{i-1} \indic[x_j = x'_j] \right) \\
& \qquad \qquad - \indic[x_{d,1} = 1] \indic[x'_{d,1} = 0] \prod_{j=1}^{d-1} \indic[x_j = x'_j]
\end{split} \enspace.
\end{align}
Note that $x_1, x'_1, \cdots x_D, x'_D$ are all mutually independent and $\delta_i$ is determined by $x_i$ and $x'_i$ only. Then, for $d \neq i$, we have
\begin{align}
& \E[ (x_{d,1} - x'_{d,1}) \indic[{\delta_i < 0}] \mid \F ] \notag \\
&\qquad = \E[ x_{d,1} - x'_{d,1} \mid \F ] \E[ \indic[{\delta_i < 0}] \mid \F ] \\
& \E[ (x_{d,1} - x'_{d,1}) \indic[{\delta_i > 0}] \mid \F ] \notag \\
&\qquad = \E[ x_{d,1} - x'_{d,1} \mid \F ] \E[ \indic[{\delta_i > 0}] \mid \F ] \\
& \E[ (x_{d,1} - x'_{d,1}) \indic[{x_i = x'_i}] \mid \F ] \notag \\
&\qquad = \E[ x_{d,1} - x'_{d,1} \mid \F ] \E[ \indic[{x_i = x'_i}] \mid \F ] \enspace.
\end{align}
Since $\E[x_{d,1} - x'_{d,1} \mid \F ] = 0$, the expectation of $\sample[1]_{d,1} - \sample[2]_{d,1}$ is derived as follows.
\begin{align}
& \E[\sample[1]_{d,1} - \sample[2]_{d,1} \mid \F ] \notag\\
\begin{split}
& = \E \left[ (x_{d,1} - x'_{d,1}) \sum_{i=1}^{d-1} (\indic[\delta_i > 0] - \indic[\delta_i < 0]) \prod_{j=1}^{i-1} \indic[x_j \neq x'_j] \right. \\
& \qquad\qquad + \left(\indic[x_{d,1} = 1] \indic[x'_{d,1} = 0] \right.\\
& \qquad\qquad\qquad \left. \left.  + \indic[x_{d,1} = 0] \indic[x'_{d,1} = 1] \right) \prod_{j=1}^{d-1} \indic[x_j = x'_j] \mid \F \right]
\end{split} \notag\\
& = 0 + 2 \param_{d,1} (1 - \param_{d,1}) \prod_{j=1}^{d-1} \left( \sum_{k=1}^K (\param_{j,k})^2 \right)
\end{align}
From \eqref{eq:kval_exp_theta_diff}, the conditional expectation of $\param[t+1]_{d,1} - \param[t]_{d,1}$ is obtained as \eqref{eq:exp-theta-diff-kval}, which shows $\param[t]_{d,1}$ is a sub-martingale.
}{}
\new{
Here, we define $\Delta_d^<$, $\delta_d$ and $\Delta_d^>$ for all $d \in \intrange{1}{D}$ as follows.
\begin{align}
\Delta_d^< &= \sum_{i=1}^{d-1} \sum^K_{k=1} (K-k) K^{D-i} (x_{i,k} - x'_{i,k}) \\
\delta_d &= \sum^K_{k=1} (K-k) K^{D-d} (x_{d,k} - x'_{d,k}) \label{eq:kval-delta} \\
\Delta_d^> &= \sum_{i=d+1}^D \sum^K_{k=1} (K-k) K^{D-i} (x_{i,k} - x'_{i,k}) \label{eq:kval-Delta}
\end{align}
Note that $\KVAL(x) - \KVAL(x') = \Delta_d^< + \delta_d + \Delta_d^>$ for all $d \in \intrange{1}{D}$ and
\begin{align}
    \indic[{ \KVAL(x) \geq \KVAL(x') }] &= \indic[{ \Delta_d^< > 0 }] + \indic[{ \Delta_d^< = 0 }] \indic[{ \delta_d > 0 }] + \indic[{ \Delta_d^< = 0 }] \indic[{ \delta_d = 0 }] \indic[{ \Delta_d^> \geq 0 }] 
    \label{eq:kval_diff_x_geq} \\
    \indic[{ \KVAL(x) < \KVAL(x') }] &= \indic[{ \Delta_d^< < 0 }] + \indic[{ \Delta_d^< = 0 }] \indic[{ \delta_d < 0 }] + \indic[{ \Delta_d^< = 0 }] \indic[{ \delta_d = 0 }] \indic[{ \Delta_d^> < 0 }] \enspace. \label{eq:kval_diff_x_leq}
\end{align}
Because 
\begin{align}
    (x_{d,1} - x'_{d,1}) \indic[{ \delta_d = 0 }] &= 0 \\
    (x_{d,1} - x'_{d,1}) \indic[{ \delta_d > 0 }] &= \indic[{ x_{d,1} = 1 }] \indic[{ x'_{d,1} = 0 }] \label{eq:kval-delta_d_eq} \\
    (x_{d,1} - x'_{d,1}) \indic[{ \delta_d < 0 }] &= - \indic[{ x_{d,1} = 0 }] \indic[{ x'_{d,1} = 1 }] \enspace,
\end{align}
we can transform each term of the RHS in \eqref{eq:kval-lema-decompo} as
\begin{align}
    & (x_{d,1} - x'_{d,1}) \indic[\KVAL(x) \geq \KVAL(x') ] \notag \\
    &\qquad = (x_{d,1} - x'_{d,1}) \left( \indic[{ \Delta_d^< > 0 }] + \indic[{ \Delta_d^< = 0 }] \indic[{ \delta_d > 0 }] \right) \\
    &\qquad = (x_{d,1} - x'_{d,1}) \indic[{ \Delta_d^< > 0 }] + \indic[{ \Delta_d^< = 0 }] \indic[{ x_{d,1} = 1 }] \indic[{ x'_{d,1} = 0 }] \\
    & (x_{d,1} - x'_{d,1}) \indic[\KVAL(x) < \KVAL(x') ] \notag \\
    &\qquad = (x_{d,1} - x'_{d,1}) \left( \indic[{ \Delta_d^< < 0 }] + \indic[{ \Delta_d^< = 0 }] \indic[{ \delta_d < 0 }] \right) \\
    &\qquad = (x_{d,1} - x'_{d,1}) \indic[{ \Delta_d^< < 0 }] - \indic[{ \Delta_d^< = 0 }] \indic[{ x_{d,1} = 0 }] \indic[{ x'_{d,1} = 1 }] \enspace.
\end{align}
Note that $x_1, x'_1, \cdots, x_D, x'_D$ are all mutually independent. Moreover, $\Delta_d^<$ is determined by $x_1, x'_1 \cdots x_{d-1}, x'_{d-1}$ while $\delta_d$ is determined by $x_d$ and $x'_d$ only. Then, considering $\E[ x_{d,1} - x'_{d,1} \mid \F ] = 0$, we have
\begin{align}
& \E[ (x_{d,1} - x'_{d,1}) \indic[{ \Delta_d^< > 0 }] \mid \F ] = \E[ x_{d,1} - x'_{d,1} \mid \F ] \E[ \indic[{ \Delta_d^< > 0 }] \mid \F ] = 0 \label{eq:kval_diff_indepndent-1} \\
& \E[ (x_{d,1} - x'_{d,1}) \indic[{ \Delta_d^< < 0 }] \mid \F ] = \E[ x_{d,1} - x'_{d,1} \mid \F ] \E[ \indic[{ \Delta_d^< < 0 }] \mid \F ] = 0 \label{eq:kval_diff_indepndent-2}\\
& \E[ \indic[{ \Delta_d^< = 0 }] \indic[{ x_{d,1} = 1 }] \indic[{ x'_{d,1} = 0 }] \mid \F ] = \Pr( \Delta_d^< = 0 \mid \F ) \E[ \indic[{ x_{d,1} = 1 }] \indic[{ x'_{d,1} = 0 }] \mid \F ]\label{eq:kval_diff_indepndent-3} \\
& \E[ \indic[{ \Delta_d^< = 0 }] \indic[{ x_{d,1} = 0 }] \indic[{ x'_{d,1} = 1 }] \mid \F ] = \Pr( \Delta_d^< = 0 \mid \F ) \E[ \indic[{ x_{d,1} = 0 }] \indic[{ x'_{d,1} = 1 }] \mid \F ] \enspace. \label{eq:kval_diff_indepndent-4}
\end{align}
Since
\begin{align}
    \E[ \indic[{ x_{d,1} = 0 }] \indic[{ x'_{d,1} = 1 }] \mid \F ] = \E[ \indic[{ x_{d,1} = 1 }] \indic[{ x'_{d,1} = 0 }] \mid \F ] = \param_{d,1} (1 - \param_{d,1}) 
\end{align}
and $\Pr( \Delta_d^< = 0 \mid \F )$ equals\revdel{ to}
\begin{align}
    \E\left[ \prod^{d-1}_{j=1} \indic[x_j = x'_j] \mid \F \right] = \prod^{d-1}_{j=1} \left( \sum_{k=1}^K (\param_{j,k})^2 \right) \enspace,
\end{align}
combining them with \eqref{eq:kval_diff_x_geq}, \eqref{eq:kval_diff_x_leq}, \eqref{eq:kval_diff_indepndent-1}--\eqref{eq:kval_diff_indepndent-4} shows
\begin{align}
    &\E \left[ (x_{d,1} - x'_{d,1}) \indic[\KVAL(x) \geq \KVAL(x') ] \mid \F \right] = \param_{d,1} (1 - \param_{d,1}) \prod^{d-1}_{j=1} \left( \sum_{k=1}^K (\param_{j,k})^2 \right) \\
    &\E \left[ (x_{d,1} - x'_{d,1}) \indic[\KVAL(x) < \KVAL(x') ] \mid \F \right] = - \param_{d,1} (1 - \param_{d,1}) \prod^{d-1}_{j=1} \left( \sum_{k=1}^K (\param_{j,k})^2 \right) \enspace.
\end{align}
Using \eqref{eq:kval_exp_theta_diff} and \eqref{eq:kval-lema-decompo}, we obtain \eqref{eq:exp-theta-diff-kval}.
}

Then, we define $T_d = \min\{ t \in \mathbb{N}_0 : \param_{d,1} \leq 1/(2K) \}$. We note $\param[0]_{d,1} = 1 / K$ and $|\param[t+1]_{d,1} - \param_{d,1}| \leq \eta < \sqrt{2} \eta$. As $(- \param_{d,1})$ is \rev{a }super-martingale\rev{, which is shown by \eqref{eq:exp-theta-diff-kval}}, applying~\cite[Corollary 4.1]{Wormald:1999} with $b=0$ shows
\begin{align}
\Pr( T_d \leq s ) &\leq \exp\left( - \left(\frac{1}{2K}\right)^2 \frac{1}{2 s (\sqrt{2} \eta)^2} \right) = \exp\left( -\frac{1}{16 K^2 \eta^2 s}\right) \enspace.
\end{align}
Finally, considering the union bound over all $d \in \llbracket 1, D \rrbracket$ completes the proof.
\end{proof}

The proof of Theorem~\ref{thm:runtime-upper-kval} is as follows.

\del{
The proof of Theorem~\ref{thm:runtime-upper-kval} is briefly summarized as follows.
First, we consider the first hitting time $T_1$ of the event $\prod^D_{d=1} \param_{d,1} \geq 1/2$, which equal to $\X[t] \geq - \ln 2 + D \ln K$. 
Then some transformations with the well-known inequality $\ln z \geq (z-1)/z$ with $z>0$, the condition of $\eta$ and \eqref{eq:exp-theta-diff-kval} show 
a lower bound of the conditional drift of $\X$ under the condition of $\param_{d,1} > 1 / (2K)$ as
\begin{align}
    \eta \left( \sum_{d=1}^D \left( 2(1 - \param[t]_{d,1}) \prod_{j=1}^{d-1} (\param[t]_{j,1})^2 \right) - \frac{1}{8} \right) \enspace. 
\end{align}
Considering the cases where an index $d$ satisfies $\prod_{j=1}^{d-1} (\param[t]_{j,1})^2 \geq 1/2$ and $\prod_{j=1}^{d} (\param[t]_{j,1})^2 \leq 1/2$ for every $d \in \intrange{1}{D}$ individually, we obtain the lower bound of the conditional drift as $\eta / 8$. 
Then, Theorem~\ref{theo:additive-d-upper} shows the upper bound of $\Pr(T_1 \leq u)$ with $u= c_2 D \ln K / (2 \eta)$.
Finally, applying Lemma~\ref{lem:general-tail-upper} finishes the proof.
}{}


\begin{proof}[Proof of Theorem~\ref{thm:runtime-upper-kval}]

To apply Lemma~\ref{lem:general-tail-upper}, we consider
\begin{align}
T_1 = \min \left\{ t \in \mathbb{N}_0 : \prod^D_{d=1} \param_{d,1} \geq \frac{1}{2} \right\} \enspace.
\end{align}
Let $u_1 = c_2 D \ln K / (2 \eta)$, $u_2 = (4 \eta D)^{-1}$ and $u = u_1 + u_2$.
Note that $c_2 D \ln K / \eta \geq u$ and $\Pr \left( T_\mathrm{Hit} \leq c_2 D \ln K / \eta \right) \geq \Pr \left( T_\mathrm{Hit} \leq u \right)$.
According to Lemma~\ref{lem:general-tail-upper}, we have
\begin{align}
\Pr \left( T_\mathrm{Hit} \leq u \right) > 1 - \Pr( T_1 \geq u_1) - \left( 1 - \frac{1}{4} \right)^{2 u_2} \enspace. \label{eq:kval-upper-tail-decompose}
\end{align}
In the following, we derive the upper bounds of the second term and third term in \eqref{eq:kval-upper-tail-decompose}.

First, we consider the second term $\Pr( T_1 \geq u_1 )$ in \eqref{eq:kval-upper-tail-decompose}. By the definition of $\X$ \rev{given in \eqref{eq:kval-potential}}, $T_1$ can be represented as 
\begin{align}
T_1 = \min \left\{ t \in \mathbb{N}_0 : \X[t] \geq - \ln 2 + D \ln K \right\} \enspace.
\end{align}
To apply Theorem~\ref{theo:additive-d-upper}, we consider the event $\event_1$ as
\begin{align}
\event_1 &: \min_{s \in \intrange{0}{t} } \min_{d \in \intrange{1}{D} } \param[s]_{d,1} \geq \frac{ 1 }{ 2 K } \enspace. \label{eq:kval_upper_event}
\end{align}
Then, Lemma~\ref{lem:frequency} shows that the probability of the complementary event of $\event[u_1 - 1]_1$ is bounded from above as
\begin{align}
\Pr(\bar{E}_1^{(u_1 - 1)}) &\leq \Pr(\bar{E}_1^{(u_1)}) \\
&\leq D \exp \left( - \frac{1}{16 K^2 \eta^2 u_1} \right) \\
& \leq D \exp\left( - \frac{1}{16 K^2} \cdot \frac{2 c_1 D K^2 \ln K \ln (DK)}{c_2 D \ln K} \right) \\
&= D (DK)^{- \frac{c_1}{8 c_2}} \leq (DK)^{- \ckv[5]} \enspace, \label{eq:kval-upper-event-1-prob}
\end{align}
where $\ckv[5] = \ckv[1] / (8 \ckv[2]) - 1 = 1$.

Next, we consider the drift of $\X$ in \eqref{eq:kval-potential}. As with the proof of Theorem~\ref{thm:runtime-upper-one}, we denote $p_{d,1}^{(t)} := \Pr (\param[t+1]_{d,1} = \param[t]_{d,1} + \eta \mid \F[t])$ and $q_{d,1}^{(t)} := \Pr (\param[t+1]_{d,1} = \param[t]_{d,1} - \eta \mid \F[t])$. 
Note \rev{that }$\param[t]_{d,1} - \eta$ is positive under $\event_1$ because $\eta \leq (c_1 D K^2 \ln K \ln (DK))^{-1} < 1 / (2K)$. Since $\ln z \geq (z-1)/z$ for $z>0$, the drift of $\X$ conditioned on $\event_1$ is bounded from below \revdel{as}\rev{by}
\begin{align}
\E [ \X[t+1] - \X[t] \mid \F[t] ] \indic[\event_1] &= \sum_{d=1}^D \E \left[ \ln \left( \frac{ \param[t+1]_{d,1} }{ \param[t]_{d,1} } \right) \mid \F[t] \right] \indic[\event_1] \\
&\geq \sum_{d=1}^D \E \left[ \frac{ \param[t+1]_{d,1} / \param[t]_{d,1} - 1 }{ \param[t+1]_{d,1} / \param[t]_{d,1} } \mid \F[t] \right] \indic[\event_1] \\
&= \sum_{d=1}^D \left( \frac{ \eta }{ \param[t]_{d,1} + \eta } p_{d,1}^{(t)} - \frac{ \eta }{ \param[t]_{d,1} - \eta } q_{d,1}^{(t)} \right) \indic[\event_1] \enspace.
\end{align}
We note $\eta (p_{d,1}^{(t)} + q_{d,1}^{(t)})$ \rev{is} equal to the drift of $\param_{d,1}$ in \eqref{eq:exp-theta-diff-kval} and $p_{d,1}^{(t)} + q_{d,1}^{(t)} = \Pr(x \neq x' \mid \F) = 2 \param_{d,1} ( 1 - \param_{d,1})$. Then, we have
\begin{align}
&\E [ \X[t+1] - \X[t] \mid \F[t] ] \indic[\event_1] \notag \\
&\qquad \geq \sum_{d=1}^D \Biggl( \frac{ 1 }{ (\param[t]_{d,1})^2 - \eta^2 } \left( \param[t]_{d,1}\E [ \param[t+1]_{d,1} - \param[t]_{d,1} | \F[t] ] - \eta^2 \Pr(x_{d,1} \neq x'_{d,1} \mid \F[t]) \right) \Biggr) \indic[\event_1] \notag \\
&\qquad = \eta \sum_{d=1}^D \Biggl( \frac{ 2 \param[t]_{d,1} (1 - \param[t]_{d,1}) }{ (\param[t]_{d,1})^2 - \eta^2 } \left( \param[t]_{d,1} \prod_{j=1}^{d-1} \left( \sum_{k=1}^K (\param[t]_{j,k})^2 \right) - \eta \right) \Biggr) \indic[\event_1] \enspace.
\end{align}
Note that 
\begin{align}
\eta\sum_{d=1}^D \frac{ 2 \param[t]_{d,1} (1 - \param[t]_{d,1}) }{ (\param[t]_{d,1})^2 - \eta^2 } \leq 8\eta DK  \leq \frac{ 8 }{ \ckv[1] K \ln K \ln (DK) } \leq \frac{1}{8}
\end{align}
when the event $\event_1$ occurs. Moreover, $\sum_{k=1}^K (\param[t]_{j,k})^2 \geq  (\param[t]_{j,1})^2$ and $(\param[t]_{d,1})^2 - \eta^2 \leq (\param[t]_{d,1})^2$.
Therefore, the drift conditioned on $\event_1$ can be calculated as
\begin{align}
\E [ \X[t+1] - \X[t] \mid \F[t] ] \indic[\event_1] \geq \eta \left( \sum_{d=1}^D \left( 2(1 - \param[t]_{d,1}) \prod_{j=1}^{d-1} (\param[t]_{j,1})^2 \right) - \frac{1}{8} \right) \indic[\event_1] \enspace. \label{eq:kval_drift_lower}
\end{align}

The lower bound of the drift in \eqref{eq:kval_drift_lower} contains the summation for $d \in \intrange{1}{D}$. We will ignore a part of \revdel{indexes}\rev{the indices} $\intrange{1}{D}$ in the summation to obtain the lower bound of \rev{the }drift. To select the indexes, we consider the event $\mathcal{E}_d^{(t)}$ defined as 
\begin{align}
\mathcal{E}_1^{(t)} &: \param[t]_{1,1} < \frac{1}{2} \\
\mathcal{E}_d^{(t)} &: \prod_{j=1}^{d-1} \param[t]_{j,1} \geq \frac12 \enspace \text{and} \enspace \prod_{j=1}^{d} \param[t]_{j,1} < \frac12 \enspace \enspace \text{for} \enspace d \in \intrange{2}{D} \enspace.
\end{align}
Note that those events do not have intersections, and one of the events occurs when $\X < - \ln 2 + D \ln K$.
When the event $\mathcal{E}_1^{(t)}$ occurs, it holds $\param_{1,1} < 1/2$. Then, ignoring the summation in \eqref{eq:kval_drift_lower} except for the case of $d=1$ shows
\begin{align}
\E[ \X[t+1] - \X[t]  \mid \F[t] ] \indic[\event_1] \indic[\mathcal{E}_1^{(t)}] &\geq \eta \left( 2(1 - \param[t]_{1,1} ) - \frac{1}{8} \right) \indic[\event_1] \indic[\mathcal{E}_1^{(t)}] \\
&> \eta \left( 2 \left(1 - \frac{1}{2} \right) - \frac{1}{8} \right) \indic[\event_1] \indic[\mathcal{E}_1^{(t)}] \\
&= \frac{7}{8} \eta \indic[\event_1] \indic[\mathcal{E}_1^{(t)}]  \enspace. \label{eq:kval_drift_lower_conditional_1}
\end{align}

Similarly, when the event $\mathcal{E}_d^{(t)}$ occurs, $\prod^{i-1}_{j=1} (\param_{j,1})^2 \geq 1/4$ for $i \in \intrange{1}{d}$.
Then, ignoring the summation in \eqref{eq:kval_drift_lower} with indexes larger than $d$ shows 
\begin{align}
&\E [ (\X[t+1] - \X[t]) \indic[\event_1] \indic[\mathcal{E}_d^{(t)}] \mid \F[t] ] \notag \\
&\qquad \geq \eta \left( 2(1 - \param[t]_{1,1}) + \sum_{i=2}^d \left( 2 (1 - \param[t]_{i,1}) \prod^{i-1}_{j=1} ( \param_{j,1} )^2 \right) - \frac{1}{8} \right) \indic[\event_1] \indic[\mathcal{E}_d^{(t)}] \\
&\qquad \geq \eta \left( 2(1 - \param[t]_{1,1}) + \sum_{i=2}^d \left( \frac{2}{2^2} (1 - \param[t]_{i,1}) \right) - \frac{1}{8} \right) \indic[\event_1] \indic[\mathcal{E}_d^{(t)}] \\
&\qquad = \eta \left( \frac{d}{2} - \frac12 \sum_{i=1}^d \param[t]_{i,1} + \frac{3}{2} (1 - \param[t]_{1,1}) - \frac{1}{8} \right) \indic[\event_1] \indic[\mathcal{E}_d^{(t)}] \\
&\qquad \geq \eta \left( \frac{d}{2} - \frac12 \sum_{i=1}^d \param[t]_{i,1} - \frac{1}{8} \right) \indic[\event_1] \indic[\mathcal{E}_d^{(t)}] \enspace. \label{eq:kval_drift_lower_conditional_3-1}
\end{align}

By \rev{the }Weierstrass product inequality, we have
\begin{align}
\prod_{i=1}^d \param[t]_{i,1} = \prod_{i=1}^d (1 - (1 - \param[t]_{i,1})) \geq \sum_{i=1}^d \param[t]_{i,1} - (d-1) \enspace.
\end{align}
Therefore, $\sum_{i=1}^d \param[t]_{i,1}  \leq d - 1/2$ under $\mathcal{E}_d^{(t)}$ and we obtain
\begin{align}
\E [ (\X[t+1] - \X[t]) \indic[\event_1] \indic[\mathcal{E}_d^{(t)}] \mid \F[t] ] &\geq \eta \left( \frac{d}{2} - \frac12 \left( d - \frac12 \right) - \frac{1}{8} \right) \indic[\event_1] \indic[\mathcal{E}_d^{(t)}] \\
&= \frac{\eta}{8} \indic[\event_1] \indic[\mathcal{E}_d^{(t)}] \enspace. \label{eq:kval_drift_lower_conditional_3}
\end{align}

Note that $\bigcup_{d=1}^D \mathcal{E}_d^{(t)}$ is equal to the events $\prod_{d=1}^D \param[t]_{d,1} < 1/2$ and $\X[t] <  - \ln 2 + D \ln K $.
Because $t < T_1$ satisfies $\X[t] < - \ln 2 + D \ln K$, we obtain
\begin{align}
\E [(\X[t+1] - \X[t]) \indic[\event_1] \indic[{t < T_1}] \mid \F[t]] \geq \frac{\eta}{8} \indic[\event_1] \indic[{t < T_1}] \enspace.
\end{align}
Considering 
\begin{align}
|\X[t] - \X[t+1] | \indic[\event_1] \leq D\ln\left( \frac{1}{2K} + \eta \right) - D\ln\left( \frac{1}{2K} \right) < 2\eta DK \enspace, 
\end{align}
and $u_1 \geq 2 (- \ln2 + D \ln K) / (\eta / 8)$, 
we obtain an upper bound of $\Pr\left( T_1 \geq u_1 \right)$ by Theorem~\ref{theo:additive-d-upper} as
\begin{align}
\Pr\left( T_1 \geq u_1 \right) & \leq \exp\left( - \frac{(c_2 D \ln K / (2\eta))}{8 (2 \eta DK)^2}  \left( \frac{\eta}{8} \right)^2 \right) + \Pr(\bar{E}_1^{(u_1 - 1)}) \\
& \leq \exp\left( - \frac{c_2 }{2^{12}}  \frac{D \ln K \cdot c_1 DK^2 \ln K \ln (DK)}{D^2 K^2} \right) + (DK)^{- \ckv[5]} \\
& \leq \exp\left( - \frac{ c_1 c_2}{2^{12}}  (\ln K)^2 \ln (DK) \right) + (DK)^{- \ckv[5]} \\
& \leq (DK)^{- \ckv[6]} + (DK)^{- \ckv[5]} \enspace, \label{eq:kval_t1_tail_bound}
\end{align}
where $\ckv[6] = \ckv[1] \ckv[2] (\ln 2)^2 / 2^{12} = 1.921 \cdots \geq 1$. 

Next, we consider the upper bound of the third term in \eqref{eq:kval-upper-tail-decompose}. Recalling $u_2 = (4\eta D)^{-1}$, we get
\begin{align}
\left( 1 - \frac{1}{4} \right)^{2 u_2} &= \exp\left( - 2 u_2 \ln (4/3) \right) \\
&\leq \exp\left( - \frac{\ln (4/3)}{2} \cdot \frac{c_1 DK^2 \ln K \ln (DK)}{D}  \right) \\
&\leq \exp\left( - \frac{\ln (4/3)}{2} \cdot c_1 4 \ln 2 \ln (DK)  \right) \\
&= (DK)^{-\ckv[7]}\enspace, \label{eq:kval-upper-third-upper}
\end{align}
where $\ckv[7] = 2(\ln 2) \ln (4/3) \ckv[1] = 204.9 \cdots \geq 1$.

Finally, we get the upper bound of $\Pr(T_\mathrm{Hit} > u)$ from \eqref{eq:kval-upper-tail-decompose}. From \eqref{eq:kval_t1_tail_bound} and \eqref{eq:kval-upper-third-upper}, when \new{$D \geq 3^2 / 2$},
\begin{align}
\Pr ( T_\mathrm{Hit} > u ) & \leq (DK)^{-c_5} + (DK)^{-c_6} + (DK)^{-c_7} \\
&\leq 3 (DK)^{-c_{8} - \frac{1}{2}} \\
&\leq (DK)^{-c_{8}} \enspace,
\end{align}
where $c_{8} = \min\{ c_5, c_6, c_7 \} - 1 / 2  = 1/2$. Recall that $\Pr (T_\mathrm{Hit} \leq c_2 D \ln K / \eta ) \geq \Pr ( T_\mathrm{Hit} \leq u )$,
\begin{align}
\Pr\left( T_\mathrm{Hit} \leq \frac{ \cone[2] D \ln K }{ \eta } \right) > 1 - (DK)^{- \cone[3]} \enspace.
\end{align}
\revdel{This is the end of the proof.}{}\rev{This concludes the proof.}
\end{proof}

\subsection{Lower Tail Bound of \rev{the }Runtime on \textsc{KVal}}

The lower tail bound is necessary to show \rev{that }the runtime\rev{s} on COM and \textsc{KVal} differ w.r.t. $K$ even though the upper tail bound derived in the previous section agrees with the experimental result, as shown in Section~\ref{sec:experiment}. In this section, we investigate the lower tail bound of the runtime on \textsc{KVal}.

\del{
\subsubsection{Proof Sketch}
Similar to the derivation of the upper tail bound on \textsc{KVal}, we will establish another lemma before showing the lower tail bound of the first hitting time. Compared with Lemma~\ref{lem:frequency}, which provides the lower tail bound of $\param_{d,1}$, new lemma shows the lower tail bound of the ratio between $\param_{d,1}$ and $\param_{d,k}$ for any $k \in \intrange{2}{K}$. More precisely, within $O(D \ln K / \eta)$ iterations, $2 \param_{d,1}$ is greater than $\param_{d,k}$ with high probability when the learning rate satisfies $\eta^{-1} \in O(D K^2 \ln K \ln (DK))$.

Using this relation, we will show the conditional drift of $\X$ is $O(\eta)$. Then, the conditional drift theorem shows the potential $\X[t]$ is maintained to be smaller than $ {D}\ln K / 2$ for $O(D \ln K / \eta)$ iterations with high probability. When $\X[t] <  {D}/{2} \ln K$, the probability of sampling optima is significantly low. Therefore, the lower tail bound of first hitting time on \textsc{KVal} is given by $O(D \ln K / \eta)$.

\subsubsection{Details of Proof}
}{}

\subsubsection{\revdelsec{Outline and Main Analysis Result}\rev{Main Result and Outline of the Proof}}

The domino convergence~\citep{Domino:2018} is one of the key points in the dynamics of the distribution parameters on the optimal categories $\{ \param_{d,1} \}_{d=1}^D$ on \textsc{KVal}.
On \textsc{KVal}, $\param_{d,1}$ behaves as a random walk until the probability that the same categories are generated in the samples $x$ and $x'$ on $d'$-th dimension for all $d' < d$ is too small. Then, $\param_{d,1}$ tends to increase after the probability becomes large enough. As a result, \revdel{the distribution parameters tend to be optimized by turn}\rev{the ccGA tends to optimize the distribution parameter} from the first element $\param_{1,1}$ to the last element $\param_{D,1}$. By the random walk update in the early phase, the distribution parameters corresponding to non-optimal categories $\param_{d,2}, \cdots, \param_{d,K}$ are possibly updated too large compared with $\param_{d,1}$, which delays the optimization of ccGA. 
We will show that such an event does not occur with high probability when the learning rate is small enough in Lemma~\ref{lemma:kval-theta-ratio}.

Then, we derive the upper bound of the conditional drift of the potential function in \eqref{eq:kval-potential} to apply Theorem~\ref{theo:additive-d-lower}. We consider two events, the event $\event_1$ where $\param_{d,1} \geq 1/(2K)$ for all $d \in \intrange{1}{D}$ and the event $\event_2$ where $2 \param_{d,1} \geq \param_{d,k}$ for all $d \in \intrange{1}{D}$ and $k \in \intrange{2}{K}$. We obtain the upper bound of the conditional drift of $\X$ as $3 \eta$ under the intersection of these events.
Then, Theorem~\ref{theo:additive-d-lower} shows the upper bound of $\Pr(T_1 < u)$ with $u= \Theta( D \ln K / \eta)$.
Finally, applying Lemma~\ref{lem:general-tail-lower} shows the main analysis result as follows.

\begin{theorem} \label{thm:runtime-lower-kval}
Consider the update of ccGA on \textsc{KVal}. 
\new{Assume $\eta := \eta(D,K)$ is given by a function of $D$ and $K$ and there exists \rev{a }strictly positive constant value $c_2 > 0$ fulfilling $(\eta K)^{-1} \in \mathbb{N}$ and $c_1 D K^2 \ln K \ln (DK) \leq \eta^{-1} \leq (D K)^{c_2}$ for any $D \in \mathbb{N}$ and $K \in \mathbb{N}$ with $c_1 = 32 / 3$.}
Then, there exist constant values $c_3, c_4$ and $D_1 > 0$ independent from $D$, $K$ and $\eta$ and satisfying the following for all $D \geq D_1$ and $K \geq 2$: For the runtime $T_\mathrm{Hit}$,
\begin{align}
\Pr\left( T_\mathrm{Hit} \geq \frac{ c_3 D \ln K }{ \eta } \right) \geq 1 - (DK)^{- c_4}  \enspace, \label{eq:runtime-lower-kval}
\end{align}
where $c_3 = 1/12$, $c_4 = 1/2$, and $D_1 = \del{\max\{11, 256 (c_2 + 1)^2\}}{}\new{256 (c_2 + 1)^2}$.
\end{theorem}

We note that the condition for the learning rate is the same as that in Theorem~\ref{thm:runtime-upper-kval}, except for the constant factor.

\subsubsection{Details of \rev{the }Proofs}
The next lemma shows that $\param_{d,1}$ does not become smaller than the halves of all of $\param_{d,2}, \cdots, \param_{d,K}$ with high probability when the learning rate is small enough.

\begin{lemma} \label{lemma:kval-theta-ratio}
Consider the update on \textsc{KVal}. Let $d \in \intrange{1}{D}$ and $k \in \intrange{2}{K}$ be arbitrary. Let $\X_{d,k} = 2 \param_{d,1} - \param_{d,k}$ and $T_{d,k} = \min\{ t \in \N_0 : \X_{d,k} \leq 0 \}$. Then, for $s \in \mathbb{N}$,
\begin{align}
\Pr( T_{d,k} \leq s ) \leq \exp \left( - \frac{1}{32 K^2 \eta^2 s} \right) \enspace. \label{eq:theta-ratio-prob}
\end{align}
\end{lemma}

\del{
We give a brief sketch of the proof of Lemma~\ref{lemma:kval-theta-ratio} as follows.
The one-step difference of $\X_{d,k}$ is decomposed as
\begin{align*}
\X[t+1]_{d,k} - \X_{d,k} = 2 (\param[t+1]_{d,1} - \param_{d,1}) - (\param[t+1]_{d,k} - \param_{d,k}) \enspace.
\end{align*}
The expectation of first term is given by \eqref{eq:exp-theta-diff-kval} multiplied by two.
As the category with smaller index has greater weight, $\param_{d,k}$ increases when one sample selects $k$-th category and another sample selects a category with greater index than $k$ on $d$-th dimension. Similarly, $\param_{d,k}$ decreases when one sample selects $k$-th category and another selects a category with smaller index than $k$. As a result, we obtain the drift of $\param_{d,k}$ as follows.
\begin{align*}
    \eta \left(-2\param_{d,1}\param_{d,k} + 2\param_{d,k}(1 - \param_{d,1} - \param_{d,k}) \right) \prod_{j=1}^{d-1} \left( \sum_{k=1}^K (\param_{j,k})^2 \right)
\end{align*}
Combining this with \eqref{eq:exp-theta-diff-kval} shows the drift of $\X_{d,k}$ is positive when $\X_{d,k} \geq 0$. 
Finally, applying~\cite[Corollary 4.1]{Wormald:1999} with $b=0$ finishes the proof.
}{}

\begin{proof}[Proof of Lemma~\ref{lemma:kval-theta-ratio}]
First, we will show \rev{that }the drift of $\X_{d,k}$ is positive until $\X_{d,k}$ becomes negative.
The one-step difference $\X[t+1]_{d,k} - \X_{d,k}$ can be decomposed as
\begin{align}
\X[t+1]_{d,k} - \X_{d,k} = 2 (\param[t+1]_{d,1} - \param_{d,1}) - (\param[t+1]_{d,k} - \param_{d,k}) \enspace.
\end{align}
As the expectation of $\param[t+1]_{d,1} - \param_{d,1}$ is derived in \eqref{eq:exp-theta-diff-kval}, we will consider the expectation of $\param[t+1]_{d,k} - \param_{d,k}$. We note that $\param[t+1]_{d,k} - \param_{d,k} = \eta(\sample[1]_{d,k} - \sample[2]_{d,k})$ and the difference $\sample[1]_{d,1} - \sample[2]_{d,1}$ can be transformed as
\begin{align}
\sample[1]_{d,k} - \sample[2]_{d,k} = (x_{d,k} - x'_{d,k}) \indic[\KVAL(x) - \KVAL(x') \geq 0] - (x_{d,k} - x'_{d,k}) \indic[\KVAL(x) - \KVAL(x') < 0] \enspace. \label{eq:kval-lemma-dk-diff}
\end{align}
Since $(x_{d,k} - x'_{d,k}) \indic[x_d = x'_d] = 0$, each term in \eqref{eq:kval-lemma-dk-diff} can be decomposed as
\begin{align} 
& (x_{d,k} - x'_{d,k}) \indic[\KVAL(x) - \KVAL(x') \geq 0] \notag \\
& \quad = (x_{d,k} - x'_{d,k}) \left( \sum_{i=1}^{d-1} \indic[\delta_i > 0] \prod_{j=1}^{i-1} \indic[x_j = x'_j] \right) + (x_{d,k} - x'_{d,k}) \indic[\delta_d > 0] \prod_{j=1}^{d-1} \indic[x_j = x'_j] \enspace, \label{eq:kval_k_diff_g} \\
& (x_{d,k} - x'_{d,k}) \indic[\KVAL(x) - \KVAL(x') < 0]  \notag \\
& \quad = (x_{d,k} - x'_{d,k}) \left( \sum_{i=1}^{d-1} \indic[\delta_i < 0] \prod_{j=1}^{i-1} \indic[x_j = x'_j] \right) + (x_{d,k} - x'_{d,k}) \indic[\delta_d < 0] \prod_{j=1}^{d-1} \indic[x_j = x'_j] \enspace, \label{eq:kval_k_diff_l}
\end{align}
where $\delta_d$ is defined in \revdel{\eqref{eq:kval-delta_d_eq}}\rev{\eqref{eq:kval-delta}}.
The expectations of the first terms in \eqref{eq:kval_k_diff_g} and \eqref{eq:kval_k_diff_l} are zero since $x_{d,k} - x'_{d,k}$ is mutually independent from the events $\delta_i > 0$, $\delta_i < 0$ and $x_j = x'_j$ for $d \neq i$ and $d \neq j$.
Here, we consider the difference between the second terms of \eqref{eq:kval_k_diff_g} and \eqref{eq:kval_k_diff_l}, and calculate it separately considering two cases $x_{d,1} + x'_{d,1} = 1$ and $x_{d,1} + x'_{d,1} \neq 1$. Then, we have
\begin{align}
&(x_{d,k} - x'_{d,k}) \left( \indic[\delta_d > 0] - \indic[\delta_d < 0]\right) \prod_{j=1}^{d-1} \indic[x_j = x'_j] \notag \\
&= (x_{d,k} - x'_{d,k}) \left( \indic[x_{d,1} + x'_{d,1} = 1] + \indic[x_{d,1} + x'_{d,1} \neq 1] \right) \left( \indic[\delta_{d} > 0] - \indic[\delta_{d} < 0] \right) \prod_{j=1}^{d-1} \indic[x_j = x'_j] \enspace. \label{eq:kval_k_diff_s}
\end{align}
We note that
\begin{align}
x_{d,k} - x'_{d,k} &= \indic[x_{d,k} = 1] \indic[x'_{d,k} = 0] - \indic[x_{d,k} = 0] \indic[x'_{d,k} = 1] \enspace, \\
\indic[x_{d,1} + x'_{d,1} = 1] &= \indic[x_{d,1} = 1] \indic[x'_{d,1} = 0] + \indic[x_{d,1} = 0] \indic[x'_{d,1} = 1] \enspace.
\end{align}
Since $\delta_d > 0$ when $x_{d,1} = 1$ and $x'_{d,1} = 0$, and since $\delta_d < 0$ when $x_{d,1} = 0$ and $x'_{d,1} = 1$, we get
\begin{align}
&\indic[x_{d,1} + x'_{d,1} = 1] (\indic[\delta_d > 0] - \indic[\delta_d < 0]) = \indic[x_{d,1} = 1] \indic[x'_{d,1} = 0] - \indic[x_{d,1} = 0] \indic[x'_{d,1} = 1] \enspace. 
\end{align}
Then, as $x_d$ and $x'_d$ are one-hot vector\rev{s}, we have
\begin{align}
&(x'_{d,k} - x'_{d,k}) \indic[x_{d,1} + x'_{d,1} = 1] (\indic[\delta_d > 0] - \indic[\delta_d < 0]) \notag \\
&= - \indic[x_{d,1} = 1] \indic[x'_{d,1} = 0] \indic[x_{d,k} = 0] \indic[x'_{d,k} = 1] - \indic[x_{d,1} = 0] \indic[x'_{d,1} = 1] \indic[x_{d,k} = 1] \indic[x'_{d,k} = 0] \\
&= -\indic[x_{d,1} + x'_{d,1} = 1] \indic[x_{d,k} + x'_{d,k} = 1] \enspace.
\end{align}
Moreover, when $x_{d,1} + x'_{d,1} \neq 1$, we obtain 
\begin{align}
(x'_{d,k} - x'_{d,k}) \indic[x_{d,1} + x'_{d,1} \neq 1] (\indic[\delta_d > 0] - \indic[\delta_d < 0]) &\leq (x'_{d,k} - x'_{d,k}) \indic[x_{d,1} + x'_{d,1} \neq 1] \\
&\leq \indic[x_{d,k} + x'_{d,k} = 1] \indic[x_{d,1} + x'_{d,1} = 0]
\end{align}
Therefore, \eqref{eq:kval_k_diff_s} is bounded from above as
\begin{align}
\left( -\indic[x_{d,k} + x'_{d,k} = 1] \indic[x_{d,1} + x'_{d,1} = 1] + \indic[x_{d,k} + x'_{d,k} = 1] \indic[x_{d,1} + x'_{d,1} = 0] \right) \prod_{j=1}^{d-1} \indic[x_j = x'_j] \enspace.
\end{align}
To sum up, the expectation of $\sample[1]_{d,k} - \sample[2]_{d,k}$ can be bounded from above as
\begin{align}
&\E[\sample[1]_{d,k} - \sample[2]_{d,k} \mid \F ] \notag \\
\begin{split}
&\qquad \leq \E \left[ (x_{d,k} - x'_{d,k}) \sum_{i=1}^{d-1} (\indic[\delta_i > 0] - \indic[\delta_i < 0]) \prod_{j=1}^{i-1} \indic[x_j \neq x'_j]  \right. \\
&\qquad\qquad\qquad + \left( -\indic[x_{d,k} + x'_{d,k} = 1] \indic[x_{d,1} + x'_{d,1} = 1] \right. \\
&\qquad\qquad\qquad\qquad \left. \left. + \indic[x_{d,k} + x'_{d,k} = 1] \indic[x_{d,1} + x'_{d,1} = 0] \right) \prod_{j=1}^{d-1} \indic[x_j = x'_j] \mid \F \right]
\end{split}  \\
&\qquad = 0 + \left(-2\param_{d,1}\param_{d,k} + 2\param_{d,k}(1 - \param_{d,1} - \param_{d,k}) \right) \prod_{j=1}^{d-1} \left( \sum_{k=1}^K (\param_{j,k})^2 \right) \enspace. \label{eq:exp-theta_k-diff-kval}
\end{align}
Combining \eqref{eq:exp-theta_k-diff-kval} with \eqref{eq:exp-theta-diff-kval} in the proof of Lemma~\ref{lem:frequency}, the expectation of $\X[t+1]_{d,k} - \X[t]_{d,k}$ is bounded from \revdel{lower as}\rev{below by}
\begin{align}
&\E[\X[t+1]_{d,k} - \X[t]_{d,k} \mid \F] \notag \\
&\qquad = 2\E[\param[t+1]_{d,1} - \param[t]_{d,1} \mid \F] - \E[\param[t+1]_{d,k} - \param[t]_{d,k} \mid \F] \\
&\qquad \geq 2\eta \left( 2\param_{d,1}(1 - \param_{d,1}) + \param_{d,1}\param_{d,k} - \param_{d,k}(1 - \param_{d,1} - \param_{d,k})\right) \prod_{d'=1}^{d-1} \left( \sum_{k=1}^K (\param_{d',k})^2 \right) \\
&\qquad = 2\eta \left(\X_{d,k}(1 - \param_{d,1} - \param_{d,k}) + 3\param_{d,1}\param_{d,k}\right) \prod_{d'=1}^{d-1} \left( \sum_{k=1}^K (\param_{d',k})^2 \right) \enspace.
\end{align}

Note $\E[ \X[t+1]_{d,k} - \X_{d,k} \mid \F] \geq 0$ when $\X_{d,k} \geq 0$. Therefore, $(- X_{d,k})$ is a super-martingale for $- \X_{d,k} \leq 0$. Moreover, $\X[0]_{d,k} = 1 / K$ and $| \X[t+1]_{d,k} - \X_{d,k} | < 4 \eta$. Finally, applying~\cite[Corollary 4.1]{Wormald:1999} with $b=0$ shows
\begin{align}
\Pr( T_{d,k} \leq s ) \leq \exp \left( - \left( \frac{1}{K} \right)^2 \frac{1}{ 2 (4 \eta)^2 s} \right) = \exp \left( - \frac{1}{32 K^2 \eta^2 s} \right) \enspace.
\end{align}
\revdel{This is the end of the proof.}{}\rev{This concludes the proof.}

\end{proof}



\rev{The statement of~\cite[Corollary 4.1]{Wormald:1999} is found in Appendix~\ref{apdx:sec:existing}.}
The proof of Theorem~\ref{thm:runtime-lower-kval} is as follows.

\del{
The brief proof sketch is as follows.
First, we consider the first hitting time $T_1$ of the event $\prod^D_{d=1} \param_{d,1} \geq K^{-\frac{D}{2}}$, which equal to $\X[t] \geq (D \ln K) / 2$. 
The well-known inequality $\ln z \leq z-1$ with $z>0$ and \eqref{eq:exp-theta-diff-kval} show the lower bound of the conditional drift of $\X$ under the condition of $\theta_{d,1} \geq 1/(2K)$ for all $d$ as
\begin{align}
    2 \eta \sum^D_{d=1} ( 1 - \param_{d,1} ) \prod^{d-1}_{i=1} \sum^K_{k=1} (\param_{i,k} )^2 \enspace.
\end{align}
When $2 \param_{d,1} \geq \param_{d,k}$ for all $k \in \intrange{2}{K}$, the summation $\sum^K_{k=1} (\param_{i,k} )^2$ is bounded from above by both of $a_i := 1 - (1 - \param_{i,1})^2$ and $b_i := 1 - 2 \param_{i,1} (1 - \param_{i,1})$.
Combining them shows
\begin{small}
\begin{align}
    2 \eta \left(\sum^D_{d=1} (1 - a_d) \prod^{d-1}_{i=1} a_i \right) + \eta \left(\sum^D_{d=1} (1 - b_d) \prod^{d-1}_{i=1} b_i \right)
\end{align}
\end{small}
and we obtain the upper bound of the conditional drift of $\X$ as $3 \eta$. 
Then, Theorem~\ref{theo:additive-d-lower} shows the upper bound of $\Pr(T_1 < u)$ with $u= c_3 D \ln K / \eta$.
Finally, applying Lemma~\ref{lem:general-tail-lower} finishes the proof.
}{}


\begin{proof}[Proof of Theorem~\ref{thm:runtime-lower-kval}]
Let us denote $u := c_3 D \ln K / \eta$ for short. We also define
\begin{align}
T_1 = \min \left\{ t \in \mathbb{N}_0 : \prod^D_{d=1} \param_{d,1} \geq K^{-\frac{D}{2}} \right\} \enspace.
\end{align}
Then, Lemma~\ref{lem:general-tail-lower} shows
\begin{align}
\Pr( T_\mathrm{Hit} \geq u) \geq 1 - \Pr( T_1 < u) - 2 (u+1) K^{-\frac{D}{2}} \enspace. \label{eq:kval-lower-tail-decompose}
\end{align}

In the following, we derive the upper bounds of the second term and the third term in \eqref{eq:kval-lower-tail-decompose}.

First, we consider the upper bound of the third term $2 (u+1) K^{\rev{-}\frac{D}{2}}$ in \eqref{eq:kval-lower-tail-decompose}. Since \del{$D \geq 11$, $\eta^{-1} \geq c_1 = 32 / 3$ and $c_3 = 1/12$}{}\new{$c_3 = 1/12$ and $D \geq D_1 \geq 2 / c_3 = 24$}, we have
\begin{align}
2 (u+1) K^{- \frac{D}{2}} = 2 \left( \frac{c_3 D \ln K}{ \eta } + 1 \right) K^{- \frac{D}{2}} \leq 3 c_3 (D K)^{c_2 + 1} K^{- \frac{D}{2}} \enspace.
\end{align}
Moreover, since $\ln D \leq 2 \sqrt{D}$ and $1/2 \leq \ln 2$, for $D \geq 256 (c_2 + 1)^2$, it holds
\begin{align}
K^{- \frac{D}{2}} = \exp\left( - \frac{D}{2} \ln K \right) 
&\leq \exp\left( - \frac{1}{2} \left( \frac{D}{2} \ln K + \frac{D}{4 \ln D} \ln D \right) \right) \\
&\leq \exp\left( - 2 (c_2 + 1) \ln (DK) \right) \\
&= (DK)^{- 2 (c_2 + 1)} \enspace.
\end{align}
To sum up, we have
\begin{align}
2(u+1) K^{- \frac{D}{2}} \leq 3 c_3 (D K)^{- (c_2 + 1)} \enspace.
\end{align}

Next, we derive the probability $\Pr( T_1 < u)$. Since $\ln z \leq z - 1$ for $z > 0$, it holds
\begin{align}
\X[t+1] - \X = \sum^D_{d=1} \left( \ln \param[t+1]_{d,1} - \ln \param_{d,1} \right) \leq \sum^D_{d=1} \frac{ \param[t+1]_{d,1} - \param_{d,1} }{ \param_{d,1} }
\end{align}
under $\event_1$, which is defined in \eqref{eq:kval_upper_event}.
Since the expectation of $\param[t+1]_{d,1} - \param_{d,1}$ is given by \eqref{eq:exp-theta-diff-kval}, the drift is bounded from above as follows.
\begin{align}
&\E[ \X[t+1] - \X \mid \F] \indic[ \event_1 ] \leq 2 \eta \sum^D_{d=1} ( 1 - \param_{d,1} ) \prod^{d-1}_{d'=1} \Biggl( \sum^K_{k=1} (\param_{d',k} )^2 \Biggr) \cdot \indic[ \event_1 ]
\end{align}
Here, we consider the event $\event_2$ defined as
\begin{align}
\event_2 &: \min_{s \in \intrange{0}{t} } \min_{d \in \intrange{1}{D} } \min_{k \in \intrange{2}{K} } 2 \param[s]_{d,1} - \param[s]_{d,k} \geq 0 \enspace. \label{eq:kval-lower-event-2}
\end{align}
Under $\event_2$, since $\sum^K_{k=2} \param_{d',k} = 1 - \param_{d',1}$, we have
\begin{align}
\sum^K_{k=1} (\param_{d',k} )^2 \leq ( \param_{d',1} )^2 + 2 \param_{d',1} \sum^K_{k=2} \param_{d',k} = 1 - (1 - \param_{d',1})^2 \enspace.
\end{align}
There is another upper bound of $\sum^K_{k=1} (\param_{d',k} )^2$ as
\begin{align}
\sum^K_{k=1} (\param_{d',k} )^2 \leq ( \param_{d',1} )^2 + (1 - \param_{d',1} )^2 = 1 - 2 \param_{d',1} (1 - \param_{d',1} ) \rev{\enspace.}
\end{align}
Note $1 - \param_{d',1}$ can be decomposed as
\begin{align}
1 - \param_{d',1} = \param_{d',1} (1 - \param_{d',1} ) + (1 - \param_{d',1})^2 \enspace.
\end{align}
Moreover, for any sequence of $N$ real values $(a_1, \cdots, a_N)$, we have
\begin{align}
1 - \prod^N_{i=1} (1 - a_i) = \sum^N_{i=1} a_i \prod^{i-1}_{j=1} (1 - a_j) \enspace.
\end{align}
Therefore, the drift conditioned on $\event = \event_1 \cap \event_2$ is bounded from above as
\begin{align}
&\E[ \X[t+1] - \X \mid \F] \indic[ \event ] \notag \\
&\qquad \leq \Biggl( \Biggr. 2 \eta \sum^D_{d=1} ( 1 - \param_{d,1} )^2 \prod^{d-1}_{d'=1} ( 1 - (1 - \param_{d',1})^2 ) \notag \\
&\qquad\qquad\qquad + \eta \sum^D_{d=1} 2 \param_{d,1} ( 1 - \param_{d,1} ) \prod^{d-1}_{d'=1} ( 1 - 2 \param_{d',1} (1 - \param_{d',1}) ) \Biggl. \Biggr) \indic[ \event ] \\
&\qquad = \Biggl( \Biggr. 2 \eta \left( 1 - \prod^D_{d=1} ( 1 - ( 1 - \param_{d,1} )^2 ) \right) + \eta \left( 1 - \prod^D_{d=1} ( 1 -2 \param_{d,1} ( 1 - \param_{d,1} ) ) \right) \Biggl. \Biggr) \indic[ \event ] \\
&\qquad \leq 3 \eta \indic[ \event ]
\enspace.
\end{align}
Recall $\X[0] \geq 0$ and $|\X[t+1] - \X| \indic[ \event_1 ] < 2 \eta DK$. As $c_3 = 1 / 12$, applying Theorem~\ref{theo:additive-d-lower} shows
\begin{align}
\Pr\left( T_1 < u \right) & \leq \exp\left( -  \frac{ ( D \ln K )^2 }{ 128 ( \eta D K)^2 } \frac{ \eta }{ c_3 D \ln K } \right) + \Pr( \bevent[u] ) \\
&\leq (DK)^{- c_4} + \Pr( \bevent[u]_1 ) + \Pr( \bevent[u]_2 ) \enspace,
\end{align}
where $c_4 = c_1 / ( 128 c_3) = 1$. The upper bound of $\Pr( \bevent[u]_2 )$ is derived from Lemma~\ref{lemma:kval-theta-ratio} as
\begin{align}
\Pr( \bevent[u]_2 ) &\leq D (K-1) \exp\left( - \frac{1}{32 K^2 \eta^2} \frac{\eta}{ c_3 D \ln K} \right) \leq (DK)^{1-c_5}
\end{align}
where $c_5 = c_1 / (32 c_3) = 4$. We can also derive the upper bound of $\Pr( \bevent[u]_1 )$ from Lemma~\ref{lem:frequency} as $\Pr( \bevent[u]_1 ) \leq (DK)^{1-c_5}$. Therefore, by \eqref{eq:kval-lower-tail-decompose}, 
\begin{align}
\Pr\left( T_\mathrm{Hit} \geq u \right) \geq 1 - (DK)^{- c_4} - 2 (DK)^{1-c_5} - 3 c_3 (DK)^{-(c_2+1)} \enspace.
\end{align}
\del{When}{}\new{Because} $D \geq \del{11}{}\new{D_1} \geq (3 c_3 + 3)^2$, we have
\begin{align}
\Pr\left( T_\mathrm{Hit} \geq u \right) \geq 1 - (3 c_3 + 3) (DK)^{-1} \geq 1 - (DK)^{- \frac{1}{2}} \enspace.
\end{align}
\revdel{This is the end of the proof.}{}\rev{This concludes the proof.}
\end{proof}


\subsection{Discussion}

From Theorem~\ref{thm:runtime-upper-kval} and Theorem~\ref{thm:runtime-lower-kval}, we can see that the runtime of ccGA on \textsc{KVal} is $\Theta( D \ln K / \eta)$ with high probability when the learning rate is set as $O(1 / ( D K^2 \ln K \ln (DK) ) )$.
When $K=2$, Theorem~\ref{thm:runtime-upper-kval} recovers the analysis result of \citep{Domino:2018}, the runtime analysis of cGA on BinVal.
We note the derived bounds of the runtime on \textsc{KVal} are in the same order w.r.t. $K$, $\Theta(\ln K / \eta)$, as \rrevdel{that}\rrev{those} on COM under the same learning rate setting.
However, when the learning rate is set as $\Theta(1 / ( D K^2 \ln K \ln (DK) ) )$, the upper bound can be expressed as $O(D^2 K^2 (\ln K)^2 \ln (DK))$, which is greater than that on COM.
This analysis result indicates that the dependency of $K$ to the suitable setting of the learning rate differs among the class of the linear functions on the categorical domain, as well as $D$.

However, the lower tail bound on \textsc{KVal} is still not enough to show the difference in performance between COM and \textsc{KVal} w.r.t. $K$ when using the ideal learning rate settings on each function. To prove this, we need to show that \revdel{too large a}\rev{a too large} learning rate leads \rev{to} the inefficient optimization performance or the failure of the optimization, such as \revdel{the stuck}\rev{early convergence} of the distribution parameter. This is discussed in \cite[Theorem~3.4]{Domino:2018} in the analysis of BinVal. The analysis of the condition for the learning rate is left as a future work.

In the analysis of cGA on BinVal, \cite[Lemma~3.2]{Domino:2018} shows $\param_{d}$ stays above a constant threshold $1/3$ for all $d \in \intrange{1}{D}$ with high probability.
Ignoring constant factor\rev{s} in the threshold, this claim can be represented by combining Lemma~\ref{lem:frequency} with $K=2$ and Lemma~\ref{lemma:kval-theta-ratio} with $K=k=2$.
From another \rrevdel{viewpoint}\rrev{perspective}, we extend \cite[Lemma~3.2]{Domino:2018} in two ways: one is to consider the lower tail bound of $\param_{d,1}$ as shown in Lemma~\ref{lem:frequency}, and the other is to consider the ratio between $\param_{d,1}$ and $\param_{d,k}$ for $k \in \llbracket 2, K \rrbracket$ as shown in Lemma~\ref{lemma:kval-theta-ratio}. \del{This implies that the genetic drift is prevented by multiple properties in the probabilistic model-based algorithms, and the novel algorithms derived from the existing algorithms should inherit them. To reveal such properties is one of the contribution of our research.}{}

\section{Experiments}
\label{sec:experiment}


\begin{figure*}[t]
  \begin{center}
    \begin{tabular}{c}

      \begin{minipage}{0.47\hsize}
        \begin{center}
          \includegraphics[width=1.0\linewidth]{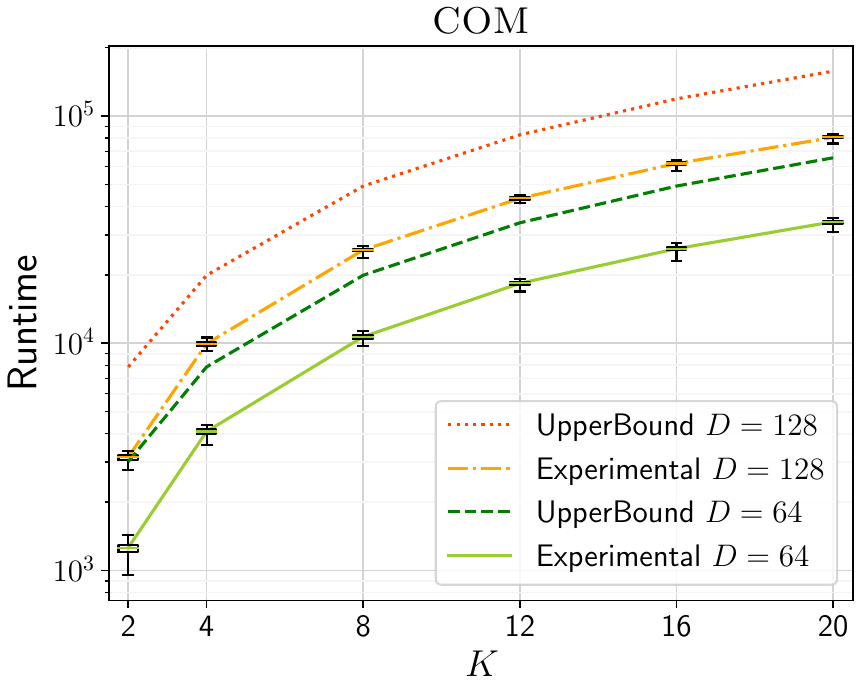}
        \end{center}
      \end{minipage} \hspace{3mm}

      \begin{minipage}{0.47\hsize}
        \begin{center}
          \includegraphics[width=1.0\linewidth]{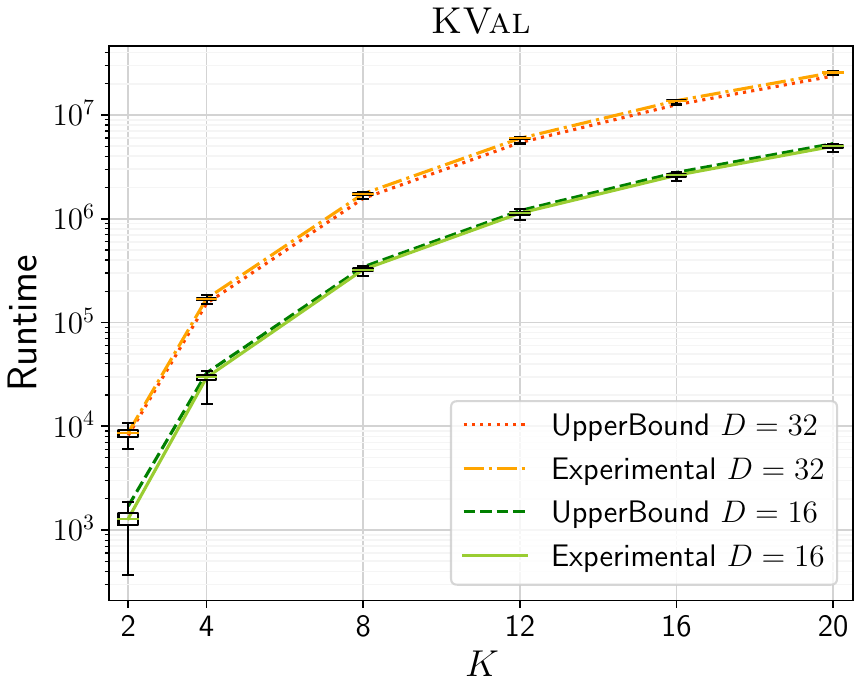}
        \end{center}
      \end{minipage}

    \end{tabular}
    \vspace{-2mm}
    \caption{The boxplots indicate \rev{experimental }runtimes of ccGA on COM~(left) and \textsc{KVal}~(right). There are one hundred runs at each point. The upper bounds on COM and \textsc{KVal} are plotted as $D K (\ln (DK))^2$ and $D^2 K^2 \ln K \ln (DK) $ \rev{by lines without boxes}, respectively.}
    \label{fig:runtime}
  \end{center}
\end{figure*}

\revdel{Theorems}\rev{The theorems} in \rrev{the} previous sections contain the constant factors in the conditions for the learning rate and the lower bound of the number of dimensions. Even though several theoretical results on \rrev{the} runtime analysis of cGA on the binary domain also contain such conditions~\citep{Domino:2018, Sudholt:2019}, cGA can optimize the target problem even when breaking such conditions slightly. In this section, we investigate whether the constant factor $c_1$ in the conditions of the learning rate is severe by setting the learning rate as the upper bound of the conditions and replacing the constant factor $c_1$ with one. We compute the runtime of the ccGA on COM and \textsc{KVal}. The detailed settings are described as follows.

\begin{itemize}
\item {Experimental settings for COM:}
Since we assume that $K$ and $\eta$ satisfy $(\eta K)^{-1} \in \mathbb{N}$ in Theorem~\ref{thm:runtime-upper-one}, we set the learning rate as \revdel{$\eta^{-1} = \lceil (\sqrt{D} K \ln (DK)) K^{-1}\rceil K$}\rev{$\eta^{-1} = \lceil \sqrt{D} \ln (DK)\rceil K$}.
We perform ccGA changing the number of categories as $K \in \{2, 4, 8, 12, 16, 20\}$ and the number of dimensions as $D= 128$ and $D = 64$. We run $100$ independent trials for each setting.

\item {Experimental settings for \textsc{KVal}:}
Similar to the setting on COM, we set \revdel{$\eta^{-1} = \lceil (D K^2 \ln K \ln (DK)) K^{-1}\rceil K$}\rev{$\eta^{-1} = \lceil D K \ln K \ln (DK)\rceil K$} so that $\del{\eta K}{}\new{(\eta K)^{-1}} \in \del{\N}{}\new{\mathbb{N}}$. We changed the number of categories as $K \in \{2, 4, 8, 12, 16, 20\}$ and the number of dimensions $D=32$ and $D=16$. We run $100$ independent trials for each setting.
\end{itemize}

Figure~\ref{fig:runtime} shows the result of the median runtime on optimizing COM and \textsc{KVal}. We also plot the upper bounds $\sqrt{D} \ln (DK) / \eta$ and $D\ln K / \eta$ for COM and \textsc{KVal}, respectively, to compare the slope of \rrev{the} actual runtime and the upper bound obtained from our runtime analysis. In all settings, all $100$ trials succeed in sampling the optimal solution, although the conditions for the constant factor in our theorems are not satisfied.
\rrevdel{When focusing}\rrev{Focusing} on the slope, we confirm that the runtime changes in the same order as the derived upper bound when the number of dimensions or categories is large enough. There is a variance in the runtime when $D$ and $K$ are small.
As a result, the actual optimization times are asymptotically close to the theoretical ones as $K$ increases.

\section{Conclusion} \label{sec:conclusion}
We investigated the runtime of ccGA, which is a probabilistic model-based algorithm with the family of the categorical distributions. We focused on two linear functions, COM and \textsc{KVal}, as representative examples of simple and \rev{relatively} difficult \revdel{objective}\rev{linear} functions. Our analysis results show that the runtime on COM is $O( \sqrt{D} \ln (DK) / \eta)$ \new{and $\Omega((\sqrt{D} + \ln K) / \eta)$} with high probability when the learning rate satisfies $\eta \in O( 1 / ( \sqrt{D} K \ln (DK) )$ and $\eta \in \Omega(1 / (DK)^c)$ for some $c >0$. The runtime on \textsc{KVal} is $\Theta( D \ln K / \eta)$ with high probability when $\eta \in O( 1 / ( D K^2 \ln K \ln (DK) )$ and $\eta \in \Omega(1 / (DK)^c)$.
These \revdel{analysis }results recover the previous results of the analysis of cGA on \textsc{KVal} as a special case $K=2$ in \citep{Domino:2018}, while there is a gap \rrev{of} $\Theta(\ln D)$ on COM compared \rrevdel{with}\rrev{to} \citep{Sudholt:2019}. We can confirm that the order of the tail bounds of the runtime on those functions is the same in $K$, unlike in $D$, when considering the same learning rate. 
The analysis of the most efficient learning rates on COM and \textsc{KVal} is necessary to show the difference of the tail bounds of the runtimes, which is left as a future work.
Our analysis results indicate the recommended setting of the learning rate of ccGA, particularly when $D$ and $K$ are large. \revdel{From another viewpoint, our result helps to formulate the optimization problem to solve easily. }As the problems on the categorical domain appear in several research fields, the analysis of search efficiency is important.

\rrev{As we delve into the conclusion of our study, it is important to introduce a new development.
A recent study by \cite{Multival:2023} tackled a mathematical runtime analysis on estimation-of-distribution algorithms (EDAs) for categorical variables, which are referred to as multi-valued variables in their work.
They showed that the multi-valued univariate marginal distribution algorithm ($r$-UMDA) solves the $r$-valued LeadingOnes problem in $O(r (\ln r)^2 n^2 \ln n)$ function evaluations, where $n$ and $r$ are the number of dimensions and categories, respectively, corresponding to $D$ and $K$ in our paper.
Although the results are mostly disjoint from our work in that they only analyze the generalized version of UMDA, they also proposed the $r$-cGA, which is the generalized version of cGA and is mostly identical to ccGA.
While the $r$-cGA is proposed from a perspective different from that of the IGO, we should probably reconsider our notation, especially in terms of the number of categories.
Furthermore, it is also important to obtain additional insights through a comparison with the analysis by Ben Jedidia et al.}

\rev{In this paper, we analyze the algorithm without the bound of the probabilities\rrev{, which corresponds to the border in cGA}. The effect of the bound is considered to be negligible since the results are derived under the condition that the probability of generating the optimal category does not become small. The analysis to deal strictly with the parameter correction by the bound is complex and is left for future work.}
\rev{In addition, our}\revdel{Our} analysis only covers the representative examples of the linear functions on the categorical domain. Therefore, the analysis of the runtime on the arbitrary linear function is one of our future works. We also consider the generalization of the sample size, particularly about the relation between the optimal sample size and $K$, as another future work.

\section*{Acknowledgments}
This work was partially supported by JSPS KAKENHI Grant Numbers JP20J23664 and JP20H04240, and a project commissioned by NEDO (JPNP18002).

\appendix

\rev{
\section{Existing Theorems and Lemmas} \label{apdx:sec:existing}
In this section, we introduce existing theorems and lemmas used in our proofs. We modify the notations to improve readability.
\begin{lemma}[Part of Lemma~3 in \rrevdel{\citep{Friedrich:2017}}\rrev{\cite{Friedrich:2017}}]
    \label{apdx:Friedrich:2017:lemma3}
    Let $0 < a < 1$ be a constant. Consider a random variable $Z = Z_1 + \cdots + Z_D$, where $Z_d$ is independently given by
    \begin{align}
        Z_d = \begin{cases}
            1 &\text{with probability} \quad p_d (1 - p_d) \\
            -1 &\text{with probability} \quad p_d (1 - p_d) \\
            0 &\text{with probability} \quad 1 - 2 p_d (1 - p_d) \\
        \end{cases}
    \end{align}
    with $a \leq p_d \leq 1$ for every $d \in \{1, \cdots, D\}$. Then, it holds $\Pr( Z = 0 ) \geq 1 / (4 \sqrt{D})$.
\end{lemma}
\begin{lemma}[Lemma~4 in \rrevdel{\citep{Droste:2006}}\rrev{\cite{Droste:2006}}]
    \label{apdx:Droste:2006:lemma4}
    Let $p_1, \cdots, p_D \in [0,1]$ and $p = \sum_{d=1}^D p_d (1 - p_d)$. Consider two samples $y$ and $y'$ generated from the Bernoulli distribution
    \begin{align}
        p_\theta(y) = \prod^D_{d=1} (p_d)^{y_d} (1 - p_d)^{( 1 - y_d)} \enspace.
    \end{align}
    Let $S_\mathrm{opt} = | \sum_{d=1}^D (y_d - y'_d) |$. Then, it holds $\E[S_\mathrm{opt}] \leq \sqrt{2 p}$.
\end{lemma}
\begin{theorem}[Theorem~3.2,~(iv) in \rrevdel{\citep{Variable:2021}}\rrev{\cite{Variable:2021}}] \label{thm:general-drift}
Consider a stochastic process $(\X, \F)_{t \in \mathbb{N}_0}$ over some state space $S \in \mathbb{R}$. Let $T$ be a stopping time defined with some $a \geq 0$ as
\begin{align}
T = \min \left\{ t \in \mathbb{N} : \X \leq a \right\} \enspace.
\end{align}
Let $g: S \to \mathbb{R}_{\geq 0}$ be a function such that $g(0) = 0$ and $g(x) \geq g(a)$ for all $x > a$. Assume $S \cap \{ x \mid x \leq a \}$ is absorbing. Then, if there exists $\lambda > 0$ and a function $\beta_l: \mathbb{N}_0 \to \mathbb{R}_{>0}$ such that
\begin{align}
    \E \left[ e^{ \lambda ( g(\X) - g(\X[t+1])) } - \beta_l(t) \mid \F \right] \indic[{ \X > a}] \leq 0
\end{align}
for all $t \in \mathbb{N}_0$, it holds, for $n > 0$ and $\X[0] > a$,
\begin{align}
    \Pr( T \leq n \mid \F[0] ) \leq \left( \prod^{n-1}_{t=0} \beta_l(t) \right) e^{ \lambda ( g(a) - g(\X[0])) } \enspace.
\end{align}
\end{theorem}
\begin{theorem}[Theorem~4.1 in \rrevdel{\citep{Wormald:1999}}\rrev{\cite{Wormald:1999}}]
    \label{apdx:Wormald:1999:theorem4}
    If, with respect to same process, $X^{(t)}$ is a super-martingale and $T$ is a stopping time, then $X^{(t \wedge T)}$ is also a super-martingale with respect to the same process, with the notation $t \wedge T = \min \{ t, T \}$.
\end{theorem}
\begin{corollary}[Corollary~4.1 in \rrevdel{\citep{Wormald:1999}}\rrev{\cite{Wormald:1999}}]
    \label{apdx:Wormald:1999:corollary4}
    Let $G^{(0)}, \cdots, G^{(\tau)}$ be a random process and $X^{(t)}$ a random variable determined by $G^{(0)}, \cdots, G^{(t)}$ for $0 \leq t \leq \tau$. Suppose that for same $b \in \R$ and constants $c_t \in \R$, $\E[ X^{(t+1)} - X^{(t)} \mid G^{(0)}, \cdots, G^{(t)} ] < b$ and $| X^{(t+1)} - X^{(t)} - b | \leq c_{t+1}$ for all $0 \leq t \leq \tau - 1$. Then, for all $\alpha > 0$,
    \begin{align}
        \Pr\left( X^{(t)} - X^{(0)} \geq tb + \alpha \quad \text{for all} \quad t \in \{0, \cdots, \tau \} \right) 
        \leq \exp \left( - \frac{\alpha^2}{2 \sum_{t=1}^\tau c_t^2} \right)
    \end{align}
\end{corollary}
\begin{theorem}[Theorem~1 in \rrevdel{\citep{Kotzing:2014}}\rrev{\cite{Kotzing:2014}}]
    \label{apdx:Kotzing:2014:theorem1}
    Consider a stochastic process $(\X, \F)_{t \in \mathbb{N}_0}$ with $\X[0] \leq 0$. Let $m \in \mathbb{R}_{>0}$ and $T = \min\{ t \in \mathbb{N}_0 : \X \geq m \}$.
    Then, for a constants $c > 0$ and $ \varepsilon \in [0, c]$,  if it holds, for all $t < T$,
    \begin{align}
    \E[ \X[t+1] - \X \mid \F ]  &\leq \varepsilon \\
    |\X[t+1] - \X | &< c \enspace,
    \end{align}
    then it holds, for all $n \leq m / (2 \varepsilon)$,
    \begin{align}
    \Pr \left( T < n \right) \leq \exp \left( - \frac{ m^2 }{8 c^2 n} \right) \enspace. 
    \end{align}
\end{theorem}
\begin{theorem}[Theorem~2 in \rrevdel{\citep{Kotzing:2014}}\rrev{\cite{Kotzing:2014}}] 
    \label{apdx:Kotzing:2014:theorem2}
    Consider a stochastic process $(\X, \F)_{t \in \mathbb{N}_0}$ with $\X[0] \geq 0$. Let $m \in \mathbb{R}_{>0}$ and $T = \min\{ t \in \mathbb{N}_0 : \X \geq m \} $.
    Then, for constants $c > 0$ and $ \varepsilon \in [0, c]$, if it holds, for all $t < T$,
    \begin{align}
    \E[ \X[t+1] - \X \mid \F ] &\geq \varepsilon \\
    |\X[t+1] - \X | &< c \enspace,
    \end{align}
    then it holds, for all $n \geq 2 m / \varepsilon$,
    \begin{align}
    \Pr \left( T \geq n \right) \leq \exp \left( - \frac{n \varepsilon^2}{8 c^2} \right) \enspace. 
    \end{align}
\end{theorem}
}

\section{Proof of Lemmas} \label{apdx:sec-lemma-proof}
\revdel{In this section, we provide the proofs of Lemma~1 and Lemma~2 in our manuscript. We also rewrite their statements to improve the readability \new{(the numbers of lemmas are consistent with the main manuscript)}.}

\revdel{
\begin{lemma}[Lemma~1 in main manuscript] 
Consider the update of \target. Assume the objective function has a unique optimal solution $x^\ast$ whose categories are the first categories in all dimensions. Let $\alpha \in [0,1]$ and 
\begin{align}
T_1 &= \min \left\{ t \in \mathbb{N}_0 : \prod^D_{d=1} \param_{d,1} \geq \alpha \right\} \\
T_2 &= \min \left\{ t \in \mathbb{N}_0 : \sum^D_{d=1} \param_{d,1} \geq D - 1 +\alpha \right\} \enspace.
\end{align}
Then, for any \new{$s, u \geq 0$} satisfying $u < \alpha / (D \eta)$, it holds
\begin{align}
\Pr( T_\mathrm{Hit} \leq s + u ) &> 1 - \Pr( T_1 \geq s )  - (1 - p)^{2u} \enspace \text{and} \\ 
\Pr( T_\mathrm{Hit} \leq s + u ) &> 1 - \Pr( T_2 \geq s )  - (1 - p)^{2u} \enspace, 
\end{align}
where $p = \alpha - \eta D u > 0$.
\end{lemma}
}

\begin{proof}[Proof of Lemma~\ref{lem:general-tail-upper}]
First, we show that the RHS in \eqref{eq:upper-tail-1} is not less than the RHS in \eqref{eq:upper-tail-2} by showing $\Pr( T_1 \geq s ) \leq \Pr( T_2 \geq s )$. By \rrev{the} Weierstrass product inequality, it holds
\begin{align}
\prod^D_{d=1} \param_{d,1} &\geq 1 - \sum^D_{d=1} (1 - \param_{d,1}) = \sum^D_{d=1} \param_{d,1} - (D - 1) \enspace.
\end{align}
Therefore, we have
\begin{align}
\sum^D_{d=1} \param_{d,1} \geq D - 1 +\alpha \enspace \Rightarrow \enspace \prod^D_{d=1} \param_{d,1} \geq \alpha \enspace,
\end{align}
which shows \rrev{$T_1 \leq T_2$ and} $\Pr( T_1 \geq s ) \leq \Pr( T_2 \geq s )$.

Further, we consider the inequality \eqref{eq:upper-tail-1}. Then, the indicator of $ T_\mathrm{Hit} > s + u $ is bounded from above as
\begin{align}
\indic[{ T_\mathrm{Hit} > s + u }] &= \indic[{ T_\mathrm{Hit} > s + u }] \left( \indic[{ T_1 \geq s }] + \indic[{ T_1 < s }] \right) \\
& \leq \indic[{ T_1 \geq s }] + \indic[{ T_\mathrm{Hit} > s + u }] \indic[{ T_1 < s }] \\
&= \indic[{ T_1 \geq s }] + \indic[{ T_\mathrm{Hit} > s + u }] \sum^{\new{\lceil s \rceil-1}}_{t=0} \indic[{ T_1 = t }] \\
&\leq \indic[{ T_1 \geq s }] + \sum^{\new{\lceil s \rceil-1}}_{t=0}  \indic[{ T_\mathrm{Hit} > t + u }]  \indic[{ T_1 = t }] \enspace.
\end{align}
Denoting the event where the optimal solution is sampled in $t$-th iteration as $\event_\mathrm{opt}$, we have
\begin{align}
\indic[{ T_\mathrm{Hit} > t + u }] = \left( \prod^{\new{t + \lfloor u \rfloor}}_{r=0}\indic[{ \bevent[r]_\mathrm{opt}  }] \right) \leq \left( \prod^{\new{t + \lfloor u \rfloor}}_{\new{r=t}}\indic[{ \bevent[r]_\mathrm{opt}  }] \right)
\end{align}
and we get
\begin{align}
\indic[{ T_\mathrm{Hit} > s + u }] &\leq \indic[{ T_1 \geq s }] + \sum^{\new{\lceil s \rceil-1}}_{t=0} \left( \prod^{\new{t + \lfloor u \rfloor}}_{\new{r=t}}\indic[{ \bevent[r]_\mathrm{opt}  }] \right)  \indic[{ T_1 = t }] \enspace. \label{eq:general-tail-upper-decomp}
\end{align}
In the following, we will consider \rrevdel{the}\rrev{an} upper bound of the expectation of second term in \eqref{eq:general-tail-upper-decomp}.
We denote $p^{(t)}_\mathrm{opt} := \prod^{D}_{d=1} \param_{d,1}$ for short. Because the sample size is two, we have
\begin{align}
\Pr( \bevent[t]_\mathrm{opt} \mid \F ) = (1 - p^{(t)}_\mathrm{opt})^2 \enspace.
\end{align}
We note that $\param_{d,1} \geq \alpha$ for any $d \in \intrange{1}{D}$ when $p^{(t)}_\mathrm{opt} \geq \alpha$. Therefore, for \new{$r \in \intrange{t}{t+u}$}, we have $\param[r]_{d,1} \new{\geq \alpha - \eta \lfloor u \rfloor} \geq \alpha - \eta u > 0$ when $T_1 = t$. 
Since \rrev{the} Weierstrass product inequality shows
\begin{align}
\prod^D_{d=1} \left( 1 - \frac{\eta u}{\param_{d,1}} \right) \geq 1 - \sum^D_{d=1} \frac{\eta u}{\param_{d,1}} \enspace,
\end{align}
we have
\begin{align}
p^{(r)}_\mathrm{opt} \indic[{ T_1 = t }]  &\geq \prod^D_{d=1} ( \param_{d,1} - \eta u) \indic[{ T_1 = t }] \\
&= \left( \prod^D_{d=1} \param_{d,1} \right) \left( \prod^D_{d=1} \left( 1 - \frac{\eta u}{\param_{d,1}} \right) \right) \indic[{ T_1 = t }] \\
&\geq \left( \prod^D_{d=1} \param_{d,1} \right) \left( 1 - \sum^D_{d=1} \frac{\eta u}{\param_{d,1}} \right) \indic[{ T_1 = t }] \\
&\geq \alpha \left( 1 - \sum^D_{d=1} \frac{\eta u}{\alpha} \right) \indic[{ T_1 = t }] \\
&= p \indic[{ T_1 = t }] \enspace.
\end{align}
Then, we obtain the following for \new{$r \in \intrange{t}{t+u}$} as
\begin{align}
\Pr( \bevent[r]_\mathrm{opt} \mid \F ) \indic[{ T_1 = t }]  \leq (1 - p)^2 \indic[{ T_1 = t }] \enspace.
\end{align}
Considering the conditional expectation under $\F[\new{t+\lfloor u \rfloor}]$, we obtain
\begin{small}
\begin{align}
\E\left[ \left( \prod^{\new{t+\lfloor u \rfloor}}_{\new{r=t}}\indic[{ \bevent[r]_\mathrm{opt}  }] \right) \indic[{ T_1 = t }]  \right] 
&= \E\left[ \E \left[ \left( \prod^{\new{t+\lfloor u \rfloor}}_{\new{r=t}} \indic[{ \bevent[r]_\mathrm{opt} }] \right) \indic[{ T_1 = t }] \mid \F[\new{t+\lfloor u \rfloor}] \right] \right] \\
&= \E\left[ \Pr( \bevent[t+u]_\mathrm{opt} \mid \F[\new{t+\lfloor u \rfloor}] ) \left( \prod^{\new{t+\lfloor u \rfloor} - 1}_{\new{r=t}} \indic[{ \bevent[r]_\mathrm{opt} }] \right) \indic[{ T_1 = t }] \right] \\
&\leq (1 - p)^2 \E\left[ \left( \prod^{\new{t+\lfloor u \rfloor} - 1}_{\new{r=t}}\indic[{ \bevent[r]_\mathrm{opt} }] \right) \indic[{ T_1 = t }]  \right] \enspace.
\end{align}
\end{small}
Similarly, considering the conditional expectation under $\F[r]$ from $r=\new{t+\lfloor u \rfloor-1}$ to $r=\new{t}$ \rrevdel{by}\rrev{in} turn, we have
\begin{align}
\E\left[ \left( \prod^{\new{t+\lfloor u \rfloor}}_{\new{r=t}}\indic[{ \bevent[r]_\mathrm{opt}  }] \right) \indic[{ T_1 = t }]  \right] & \new{ \leq  (1 - p)^{2 (\lfloor u \rfloor + 1)} \E[ \indic[{ T_1 = t }] ] } \\
&\leq  (1 - p)^{2 u} \E[ \indic[{ T_1 = t }] ] \enspace.
\end{align}

Finally, since
\begin{align}
\sum^{\new{\lceil s \rceil-1}}_{t=0} \E[ \indic[{ T_1 = t }] ] = \Pr( T_1 < s ) \leq 1\enspace,
\end{align}
the expectation of second term in \eqref{eq:general-tail-upper-decomp} is bounded from above as
\begin{align}
\E\left[ \sum^{\new{\lceil s \rceil-1}}_{t=0} \left( \prod^{\new{t+\lfloor u \rfloor}}_{\new{r=t}}\indic[{ \bevent[r]_\mathrm{opt}  }] \right) \indic[{ T_1 = t }]  \right] \leq (1 - p)^{2 u}
\end{align}
and we have
\begin{align}
\Pr( T_\mathrm{Hit} > s + u ) \leq \Pr( T_1 \geq s ) + (1 - p)^{2 u} \enspace.
\end{align}
\revdel{This is end of the proof.}{}\rev{This concludes the proof.}\yohe{proof checked: 2021/12/22}
\end{proof}

\revdel{
\begin{lemma}[Lemma~2 in main manuscript] 
Consider the update of \target. Assume the objective function has a unique optimal solution $x^\ast$ whose categories are the first categories in all dimensions. Let $\alpha \in [0,1]$ and 
\begin{align}
T_1 &= \min \left\{ t \in \mathbb{N}_0 : \prod^D_{d=1} \param_{d,1} \geq \alpha^D \right\} \\
T_2 &= \min \left\{ t \in \mathbb{N}_0 : \sum^D_{d=1} \param_{d,1} \geq \alpha D \right\} \enspace.
\end{align}
Then, for any \new{$s \geq 0$}, it holds
\begin{align}
\Pr( T_\mathrm{Hit} \geq s ) &\geq 1 - \Pr( T_1 < s )  - 2 \new{(s+1)} \alpha^D \enspace \text{and} \\ 
\Pr( T_\mathrm{Hit} \geq s ) &\geq 1 - \Pr( T_2 < s )  - 2 \new{(s+1)} \alpha^D \enspace. 
\end{align}
\end{lemma}
}

\begin{proof}[Proof of Lemma~\ref{lem:general-tail-lower}]
Because the inequality of arithmetic and geometric mean shows
\begin{align}
\sqrt[D]{ \prod^D_{d=1} \param_{d,1} } \leq \frac{1}{D} \sum^D_{d=1} \param_{d,1} \enspace,
\end{align}
we have $T_2 \leq T_1$ and $\Pr( T_1 < s ) \leq \Pr( T_2 < s )$. Therefore, \eqref{eq:lower-tail-2} is established if \eqref{eq:lower-tail-1} is established.

For the proof of the inequality \eqref{eq:lower-tail-1}, let us define a sequence of events $( \event )_{t \in \mathbb{N}_0}$ defined as
\begin{align}
\event : \enspace \prod^D_{d=1} \param_{d,1} < \alpha^D \enspace.
\end{align}
For $t \in \intrange{1}{\new{\lceil s \rceil-1}}$, we have
\begin{align}
\indic[{ T_1 \geq s }] = \prod^{\new{\lceil s \rceil-1}}_{t'=0} \indic[{ \event[t'] }] \leq \indic[{ \event }] \enspace.
\end{align}
Then, the indicator of the event $T_\mathrm{Hit} \geq s$ is bounded \rrevdel{form}\rrev{from} below as
\begin{align}
\indic[{ T_\mathrm{Hit} \geq s }] &\geq \indic[{ T_\mathrm{Hit} \geq s }] \indic[{ T_1 \geq s }] \\
&= \indic[{ T_1 \geq s }] - \indic[{ T_1 \geq s }] \sum^{\new{\lceil s \rceil-1}}_{t=0} \indic[{ T_\mathrm{Hit} = t }] \\
&\geq \indic[{ T_1 \geq s }] - \sum^{\new{\lceil s \rceil-1}}_{t=0} \indic[{ T_\mathrm{Hit} = t }] \indic[{ \event }]\enspace. \label{eq:lower-tail-exp-1} 
\end{align}
Since $\event$ is $\F$-measurable and two samples are generated in each iteration, we obtain
\begin{align}
\E \left[ \indic[{ T_\mathrm{Hit} = t }] \indic[{ \event }] \mid \F \right] < 2 \alpha^D \indic[{ \event }] \enspace.
\end{align}
Then, the expectation of second term in \eqref{eq:lower-tail-exp-1} is bounded as
\begin{align}
\sum^{\new{\lceil s \rceil-1}}_{t=0} \E \left[ \indic[{ T_\mathrm{Hit} = t }] \indic[{ \event }] \right] &= \sum^{\new{\lceil s \rceil-1}}_{t=0} \E \left[ \E \left[ \indic[{ T_\mathrm{Hit} = t }] \mid \F \right] \indic[{ \event }] \right] \\
&<  \sum^{\new{\lceil s \rceil-1}}_{t=0} 2 \alpha^D \E \left[ \indic[{ \event }] \right] \\
&\leq 2 \new{(s+1)} \alpha^D \enspace.
\end{align}
Finally, as the expectation of $\indic[{ T_1 \geq s }]$ is given by $1 - \Pr(T_1 < s )$, considering the expectation of the lower bound in \eqref{eq:lower-tail-exp-1} finishes the proof.\yohe{checked: 2021/12/22}
\end{proof}

\section{Proof of Conditional Drift Theorems} \label{apdx:sec-proof}

\revdel{In this section, we provide the proof of Theorem~1, 2, and 3 in our manuscript. Furthermore, we rewrite their statements to improve the readability \new{(the numbers of theorems are consistent with the main manuscript)}.}

\revdel{
\begin{theorem}[Theorem~1 in main manuscript] 
Consider a stochastic process $(\X, \F)_{t \in \mathbb{N}_0}$ with $\X[0] \leq 0$. Consider a sequence of events $(\event)_{t \in \mathbb{N}_0 }$, where $\event \in \F$ and $\event \supseteq \event[t+1]$. Let $m \in \mathbb{R}_{>0}$ and
\begin{align}
T = \min\{ t \in \mathbb{N}_0 : \X \geq m \} \enspace. 
\end{align}
Then, for a constants $c > 0$ and $ \varepsilon \in [0, c]$,  if it holds, for all $t < T$,
\begin{align}
\E[ \X[t+1] - \X \mid \F ] \indic &\leq \varepsilon \indic \\
|\X[t+1] - \X | \indic &< c \enspace,
\end{align}
then it holds, for all $n \leq m / (2 \varepsilon)$,
\begin{align}
\Pr \left( T < n \right) \leq \exp \left( - \frac{ m^2 }{8 c^2 n} \right) + \Pr( \bevent[n-1] ) \enspace. 
\end{align}
\end{theorem}
}

\begin{proof}[Proof of Theorem~\ref{theo:additive-d-lower}]
Let us introduce another stochastic process $(\tX, \F)$ based on $(\X, \F)$ as $\tX[0] = \X[0]$ and
\begin{align}
\tX &= \X \indic[{\event[t-1]}] + ( \tX[t-1] + \varepsilon ) \indic[{\bevent[t-1]}] \enspace \label{eq:tilde-x}
\end{align}
for $t \in \intrange{1}{n}$. We also define $\tilde{T}$, similarly to $T$ in \eqref{eq:additive-drift-d-lower-T}, as
\begin{align}
\tilde{T} = \min\{ t \in \mathbb{N}_0 : \tX \geq m \} \enspace.
\end{align}
We note $\tX$ is $\F$-measurable. Then, the difference $\tX[t+1] - \tX$ can be transformed as
\begin{align}
\tX[t+1] - \tX &= \X[t+1] \indic[{\event}] + (\tX + \varepsilon) \indic[{\bevent}] - \tX ( \indic[{\event}] + \indic[{\bevent}]) \\
&= (\X[t+1] - \X) \indic[{\event}] + \varepsilon \indic[{\bevent}] \enspace,
\end{align}
where we used the relations as
\begin{align}
\indic[{\event}] \indic[{\event[t-1]}] &= \indic[{\event}]  \\
\indic[{\event}] \indic[{\bevent[t-1]}] &= 0 \enspace.
\end{align}
Therefore, the drift of $\tX$ is bounded from above as
\begin{align}
\E[ \tX[t+1] - \tX \mid \F ] &\leq \varepsilon \enspace.
\end{align}
Moreover, since $\indic[{\event}] + \indic[{\bevent}] = 1$ and $\varepsilon \leq c$, we have
\begin{align}
| \tX[t+1] - \tX | &< c \quad \text{w.p.} \enspace 1 \enspace.
\end{align}
By \cite[Theorem~1]{Kotzing:2016}, we obtain
\begin{align}
\Pr \left( \tilde{T} < n \right) \leq \exp \left( - \frac{m^2}{8 c^2 n} \right)
\end{align}
for $n \leq m / (2\varepsilon)$. 
From the definition of $\tX$, we note
\begin{align}
\tX \indic[{\event[n-1]}] &= \left(\X \indic[{\event[t-1]}] + (\tX[t-1] + \varepsilon) \indic[{\bevent[t-1]}] \right)\indic[{\event[n-1]}] \\
&= \X \indic[{\event[n-1]}] \enspace.
\end{align}
The indicator function of $T < n$ is bounded from above as
\begin{align}
\indic[{ T < n }] &= 1 - \prod^{n-1}_{t=0} \indic[{ \X < m }] \\
&= \left( 1 - \prod^{n-1}_{t=0} \indic[{ \X < m }] \right) \left( \indic[{\event[n-1]}] + \indic[{\bevent[n-1]}] \right) \\
&\leq \left( 1 - \prod^{n-1}_{t=0} \indic[{ \X < m }] \right) \indic[{\event[n-1]}] + \indic[{\bevent[n-1]}] \\
&= \left( 1 - \prod^{n-1}_{t=0} \indic[{ \tX < m }] \right) \indic[{\event[n-1]}]  + \indic[{\bevent[n-1]}] \\
&= \indic[{ \tilde{T} < n }] \indic[{\event[n-1]}] + \indic[{\bevent[n-1]}] \\
&\leq \indic[{ \tilde{T} < n }] + \indic[{\bevent[n-1]}] \enspace. \label{eq:exp-x-upper}
\end{align}
Finally, considering the expectation finishes the proof.
\yohe{Checked: 2022/02/24}
\end{proof}

\revdel{
\begin{theorem}[Theorem~2 in main manuscript] 
Consider a stochastic process $(\X, \F)_{t \in \mathbb{N}_0}$ with $\X[0] \geq 0$. Consider a sequence of events $( \event )_{t \in \mathbb{N}_0 }$, where $\event \in \F$ and $\event \supseteq \event[t+1]$. Let $m \in \mathbb{R}_{>0}$ and
\begin{align}
T = \min\{ t \in \mathbb{N}_0 : \X \geq m \} \enspace. 
\end{align}
Then, for constants $c > 0$ and $ \varepsilon \in [0, c]$,  if it holds, for all $t < T$,
\begin{align}
\E[ \X[t+1] - \X \mid \F ] \indic &\geq \varepsilon \indic \\
|\X[t+1] - \X | \indic &< c \enspace,
\end{align}
then it holds, for all $n \geq 2 m / \varepsilon$,
\begin{align}
\Pr \left( T \geq n \right) \leq \exp \left( - \frac{n \varepsilon^2}{8 c^2} \right) + \Pr( \bevent[n-1] ) \enspace. 
\end{align}
\end{theorem}
}

\begin{proof}[Proof of Theorem~\ref{theo:additive-d-upper}]
Similarly, in the proof of Theorem~\ref{theo:additive-d-lower}, we introduce $(\tX, \F)$ defined in \eqref{eq:tilde-x} and $\tilde{T}$ as
\begin{align}
\tilde{T} = \min\{ t \in \mathbb{N}_0 : \tX \geq m \} \enspace.
\end{align}
Then, the same derivation in the proof of Theorem~\ref{theo:additive-d-lower} shows
\begin{align}
\E[ \tX[t+1] - \tX \mid \F ] &\geq \varepsilon \\
| \tX[t+1] - \tX | &< c \quad \text{w.p.} \enspace 1 \enspace.
\end{align}
By \cite[Theorem~2]{Kotzing:2016}, we obtain
\begin{align}
\Pr \left( \tilde{T} \geq n \right) \leq \exp \left( - \frac{n \varepsilon^2}{8 c^2} \right) \enspace
\end{align}
for $n \geq 2 m / \varepsilon$.
As with the proof of Theorem~\ref{theo:additive-d-lower}, the indicator of $T \geq n$ is bounded from above as
\begin{align}
\indic[{ T \geq n }] &= \prod^{n-1}_{t=0} \indic[{ \X < m }] \\
&\leq \left( \prod^{n-1}_{t=0} \indic[{ \X < m }] \right) \indic[{\event[n-1]}] + \indic[{\bevent[n-1]}] \\
&= \left( \prod^{n-1}_{t=0} \indic[{ \tX < m }] \right) \indic[{\event[n-1]}] + \indic[{\bevent[n-1]}] \\
&= \indic[{ \tilde{T} \geq n }] \indic[{\event[n-1]}] + \indic[{\bevent[n-1]}] \\
&\leq \indic[{ \tilde{T} \geq n }] + \indic[{\bevent[n-1]}] \enspace.
\end{align}
Finally, considering the expectation finishes the proof.
\yohe{Checked: 2022/02/24}
\end{proof}

\revdel{
\begin{theorem}[Theorem~3 in main manuscript] 
Consider a stochastic process $(\X, \F)_{t \in \mathbb{N}_0}$ and a sequence of events $( \event )_{t \in \mathbb{N}_0 }$, where $\event \in \F$ and $\event \supseteq \event[t+1]$.
Assume $\X$ takes the value on $\{0\} \cup [x_{\min}, x_{\max}]$ for all $t \in \mathbb{N}$, where $0 < x_{\min} < x_{\max} < \infty$, and $\X$ remains zero after once $\X$ reaches zero. Let
\begin{align}
T = \min\{ t \in \mathbb{N}_0 : \X = 0 \} \enspace.
\end{align}
Then, for a constant $\varepsilon \in [0, 1]$, if it holds
\begin{align}
\E[ \X[t+1] - \X \mid \F ] \indic \leq - \varepsilon \X \indic \enspace 
\end{align}
for all $t < T$, it holds, for any $r > 0$,
\begin{align}
\Pr \left( T > \frac{r + \ln( \X[0] / x_{\min})}{\varepsilon} \right) \leq \exp \left( - r \right) + \Pr( \bevent[n-1] ) \enspace, 
\end{align}
where $n = \lceil (r + \ln (\X[0] / x_{\min}) ) / \varepsilon \rceil$.
\end{theorem}
}

\begin{proof}[Proof of Theorem~\ref{theo:multiplicative-d}]
We consider another stochastic process $(\tX, \F)$ based on $(\X, \F)$ as $\tX[0] = \X[0]$ and
\begin{align}
\tX &= \X \indic[{\event[t-1]}] \enspace
\end{align}
for $t \in \intrange{1}{n}$.
We note $\tX$ also takes a value in $\{0 \} \cup [ x_{\min}, x_{\max} ]$ and $\tX$ also remains zero after it reaches zero once. Therefore,
\begin{align}
\indic[{ \tX[n] > 0 }] = \indic[{ \tX[n] \geq x_\mathrm{min} }] \enspace.
\end{align}
Moreover, as $a \indic[{ X \geq a }] \leq X$ for a random variable $X$ and constant value $a > 0$,
\begin{align}
\indic[{ \tX[n] \geq x_{\min} }] \leq \frac{ \X[n] }{x_{\min}} \enspace.
\end{align}
Since $\event[t-1] \supseteq \event[t]$, we have $\indic[{\event[t]}] \leq \indic[{\event[t-1]}]$. Then it holds
\begin{align}
\E[ \tX[t+1] \mid \F] &= \E[ \X[t+1] \mid \F]\indic \\
&\leq (1 - \varepsilon) \X \indic \\
&\leq (1 - \varepsilon) \X \indic[{\event[t-1]}] \\
&= (1 - \varepsilon) \tX \enspace,
\end{align}
where the first and second inequalities hold because of the assumption~\eqref{eq:multiplicative-d-drift-cond} and $\indic[{\event[t]}] \leq \indic[{\event[t-1]}]$, respectively.
By induction, we get the upper bound of $\E[\tX[n] ]$ as
\begin{align}
\E [ \tX[n] ] &= \E[ \E[ \cdots \E[ \tX[n] \mid \F[n-1] ] \cdots \mid \F[0] ] ]\\
&\leq \E[ \E[ \cdots \E[ (1 - \varepsilon) \tX[n-1] \mid \F[n-2] ] \cdots \mid \F[0]] ] \\
&\leq (1 - \varepsilon)^n \E[ \tX[0] ] \enspace.
\end{align}
Because of the inequality $1 - a \leq \exp(-a)$, we have
\begin{align}
\Pr(\tX[n] > 0) & \leq (1 - \varepsilon)^n \frac{\E[ \tX[0] ]}{x_{\min}} \\
& \leq \exp \left( - n \varepsilon \right) \frac{\X[0]}{x_{\min}} \\
& = \exp \left(- \left\lceil \frac{r+ \ln (\X[0] / x_{\min})}{\varepsilon} \right\rceil \varepsilon \right) \frac{\X[0]}{x_{\min}}\\
& \leq \exp( -r ) \enspace.
\end{align}
Since $\X$ remains zero after once it reaches zero, $T > n$ if and only if $\X[n] > 0$. Moreover, since $\tX[n] = \X[n]$ when $\event[n-1]$ occurs, we have
\begin{align}
\indic[ T > n ] &= \indic[{ \X[n] > 0 }] \\
&= \indic[{ \X[n] > 0 }] \left(\indic[{\event[n-1]}] + \indic[{\bevent[n-1]}] \right) \\
&= \indic[{ \tX[n] > 0 }] \indic[{\event[n-1]}] + \indic[{ \X[n] > 0 }] \indic[{\bevent[n-1]}] \\
&\leq \indic[{ \tX[n] > 0 }] + \indic[{ \bevent[n-1] }] \enspace, \label{eq:theo3-proof-indic-upper}
\end{align}
where the last inequality \rrevdel{is held}\rrev{holds} because $ab \leq b$ for any $a, b \in \{0, 1 \}$.
Finally, considering the expectation of \eqref{eq:theo3-proof-indic-upper} finishes the proof.\yohe{checked: 2022/03/03}
\end{proof}

\rrevdel{The statements of lemma and theorems used in the proofs are found in Appendix~\ref{apdx:sec:existing}.}\rrev{The lemmas and theorems used in the proofs are provided in Appendix~\ref{apdx:sec:existing}.}

\section{Proof of \rrev{the }Drift Theorem for Skipping Process}
\label{apdx:sec-proof-skip}
In this section, we provide the proof of Theorem~\rrevdel{4}\rrev{\ref{theo:negative-skip}}\revdel{ in our manuscript}. To prove Theorem~\rrevdel{4}\rrev{\ref{theo:negative-skip}}, we first show that the following Lemma~\ref{lem:skip-hoeffding} holds.

\begin{lemma} \label{lem:skip-hoeffding}
Assume a stochastic process $\X$ with the initial state $\X[0] = 0$ and with a filtration $\F$ for $t \in \intrange{0}{u}$. The iteration when $i$-th transition occurs is denoted as $s_i$, i.e.,  
\begin{align}
s_i = \min \left\{ t \in \rrevdel{\mathbb{N}_0}\rrev{ \llbracket0,u \rrbracket } : \sum^t_{t'=1} \indic[{ \X[t'] \neq \X[t'-1] }] \geq i \right\} \enspace.
\end{align}
Consider the first hitting time when $\X$ reaches $m > 0$, i.e.,
\begin{align}
T =  \min \left\{ t \in \rrevdel{\mathbb{N}}\rrev{ \llbracket0,u \rrbracket } : \X \geq m \right\} \enspace.
\end{align}
Assume $\X$ is \rrev{a} supermartingale. Assume $| \X[t+1]  - \X | \leq c$ for some $c > 0$ with probability $1$. Then, for any $u, r \in \mathbb{N}$,
\begin{align}
\Pr( T \leq \min\{ s_r, u \} ) \leq (r+1) \exp \left( - \frac{m^2}{2 r c^2} \right) \enspace.
\end{align}
\end{lemma}

\begin{proof}[Proof of Lemma~\ref{lem:skip-hoeffding}]
Let us denote $q_i = \min\{ s_i, u \}$ \new{and $\tX = \X[t \wedge T]$}.
\del{From \cite[Theorem 4.1]{Wormald:1999}, $\tX := \X[t \wedge T]$ is supermartingale.}{} By Markov's inequality, 
\begin{align}
\Pr( T \leq \min\{ s_r, u \} ) &= \Pr( \X[q_r \wedge T] \geq m ) \\
&= \Pr( \tX[q_r] \geq m ) \\
&= \Pr\left( \exp \left( h \tX[q_r] \right) \geq \exp(hm) \right) \\
&\leq \exp(-hm) \E\left[ \exp \left( h \tX[q_r] \right) \right] \enspace,
\end{align}
where $h > 0$ is an arbitrary positive constant.
Since $\X[t \wedge s_r]$ moves only in \rrev{the} $s_i$-th iteration for $i \in \intrange{1}{r}$, we have
\begin{align}
\tX[q_r] &= \sum^{u}_{t=1} \sum^{r}_{i=1} (\tX[t] - \tX[t-1]) \indic[{ t = s_i }] \enspace.
\end{align}
With the indicator of $s_k> u$ for any $k \in \intrange{1}{r}$, we have
\begin{align}
\tX[u] \indic[{s_k > u}] &= \sum^{u}_{t=1} \sum^{r}_{i=1} (\tX[t] - \tX[t-1]) \indic[{ t = s_i }]  \indic[{s_k > u}] \\
 &= \sum^{u}_{t=1} \sum^{k-1}_{i=1} (\tX[t] - \tX[t-1]) \indic[{ t = s_i }]  \indic[{s_k > u}] \\
&= \tX[q_{k-1}] \indic[{s_k > u}] \rrev{\enspace.}
\end{align}
Then, we have
\begin{align}
\exp \left( h \tX[q_r] \right) &= \exp \left( h \tX[s_{r}] \right) \indic[{ s_r \leq u}] + \exp \left( h \tX[u] \right) \indic[{ s_r > u}] \\
&= \sum^{u}_{t=1} \exp \left( h \tX[s_{r}] \right) \indic[{ t = s_r }] + \exp \left( h \tX[u] \right) \indic[{ s_r > u}] \\
&= \sum^{u}_{t=1} \exp \left( h \tX[s_{r}] \right) \indic[{ t = s_r }] + \exp \left( h \tX[q_{r-1}] \right) \indic[{ s_r > u}] \enspace. \label{apdx:eq:hoeff-exp-term-1}
\end{align}
Recalling $\tX[0] = 0$, the first term in \eqref{apdx:eq:hoeff-exp-term-1} can be transformed as
\begin{align}
\phi(r) := \underbrace{\sum^{u}_{t_1=1} \cdots \sum^{u}_{t_r=1}}_{\text{$r$ summations}} \prod^r_{i=1} \exp \left( h (\tX[t_i] - \tX[t_i-1]) \right) \indic[{ t_i = s_i }] \enspace.
\end{align}
Similarly, the second term in \eqref{apdx:eq:hoeff-exp-term-1} can be transformed as
\begin{align}
&\exp \left( h \tX[q_{r-1}] \right) \indic[{ s_r > u}] \notag \\
&= \exp \left( h \tX[q_{r-1}] \right) \indic[{ s_r > u}] \left( \indic[{ s_{r-1} > u}] + \indic[{ s_{r-1} \leq u}] \right)\\
&= \exp \left( h \tX[q_{r-2}] \right) \indic[{ s_{r-1} > u}] \notag \\
&\quad + \exp \left( h \tX[0] \right) \underbrace{\sum^{u}_{t_1=1} \cdots \sum^{u}_{t_{r-1}=1}}_{\text{$r-1$ summations}} \prod^{r-1}_{i=1} \exp \left( h (\tX[t_i] - \tX[t_i-1]) \right) \indic[{ t_i = s_i }] \indic[{ s_r > u}] \indic[{ s_{r-1} \leq u}] \\
&= \exp \left( h \tX[q_{r-2}] \right) \indic[{ s_{r-1} > u}] + \exp \left( h \tX[0] \right) \phi(r-1) \indic[{ s_r > u}] \indic[{ s_{r-1} \leq u}] \\
&= \cdots = \exp \left( h \tX[0] \right) \indic[{ s_1 > u}] + \exp \left( h \tX[0] \right) \sum^{r-1}_{k=1} \phi(k) \indic[{ s_{k+1} >u}] \indic[{ s_{k} \leq u}]  \\
&= \sum^{r-1}_{k=1} \phi(k) \indic[{ s_{k+1} > u}] \indic[{ s_{k} \leq u}]  + \indic[{ s_1 > u}] \enspace,
\end{align}
where the last equality \rrevdel{is held}\rrev{holds} since $\tX[0] = 0$.
Therefore, because it holds $ \indic[{ s_{k+1} > u}] \indic[{ s_{k} \leq u}] \leq 1$, we have
\begin{align}
\exp \left( h \tX[q_r] \right) \leq \sum^{r}_{k=1} \phi(k) + \indic[{ s_1 > u}] \leq \sum^{r}_{k=1} \phi(k) + 1 \enspace.
\label{eq:phi-exp-sum}
\end{align}

\rrevdel{Futher}\rrev{Further}, we consider \rrevdel{the}\rrev{an} upper bound of expectation of $\phi(k)$. Since $s_i < s_j$ for $i < j$, it holds
\begin{align}
\phi(k) &= \sum^{u}_{t_1=1} \cdots \sum^{u}_{t_k=1} \prod^k_{i=1} \exp \left( h (\tX[t_i] - \tX[t_i-1]) \right)\indic[{ t_i = s_i }]  \\
&= \sum^{u}_{t_1=1} \sum^{u}_{t_2=t_1 + 1} \cdots \sum^{u}_{t_k = t_{k-1} + 1} \prod^k_{i=1} \exp \left( h (\tX[t_i] - \tX[t_i-1]) \right)\indic[{ t_i = s_i }]  \enspace.
\end{align}
Then, \revdel{the upper bound of }the expectation of $\phi(k)$ is given by
\begin{align}
\E[ \phi(k)] &= \sum^{u}_{t_1=1} \sum^{u }_{t_2=t_1 + 1} \cdots \sum^{u }_{t_k = t_{k-1} + 1} \E \left[ \prod^k_{i=1} \exp \left( h (\tX[t_i] - \tX[t_i-1]) \right) \indic[{ t_i = s_i }]  \right] \\
&{=} \sum^{u}_{t_1=1} \sum^{u }_{t_2=t_1 + 1} \cdots \sum^{u }_{t_k = t_{k-1} + 1} \E \Biggl[ \left( \prod^{k-1}_{i=1} \exp \left( h (\tX[t_i] - \tX[t_i-1]) \right) \indic[{ t_i = s_i }] \right) \Biggr. \notag \\
&\qquad\qquad\qquad\qquad \Biggl. \E\left[ \exp \left( h (\tX[t_k] - \tX[t_k-1]) \right) \indic[{ t_k = s_k }] \mid \F[t_k-1] \right]\Biggr] \enspace\rrevdel{.}\rrev{.} \label{eq:phi-exp-decomp}
\end{align}
\rrev{where the last transformation uses the law of total expectation, i.e., $\E[ Y ] = \E[ \E[ Y \mid \F[t_k-1] ] ]$ for any random variable $Y$.}
Since $\exp(hx)$ is convex on $[-c, c]$, we have, for $x \in [-c,c]$,
\begin{align}
\exp(hx) &\leq \frac{1}{2c} \left( \exp(hc) (c + x) + \exp(-hc) (c - x) \right) \\
&= \frac{1}{2} \left( \exp(hc) + \exp(-hc) \right) + \frac{x}{2c} \left( \exp(hc) - \exp(-hc) \right) \enspace
\end{align}
and
\begin{align}
& \E\left[ \exp \left( h (\tX[t_i] - \tX[t_i-1]) \right) \indic[{ t_i = s_i }] \mid \F[t_i-1] \right] \notag\\
&\quad \leq \frac{1}{2} \left( \exp(hc) + \exp(-hc) \right) \Pr( t_i = s_i \mid \F[t_i-1] ) \notag \\
&\quad\qquad + \frac{\E[ (\tX[t_i] - \tX[t_i-1] )\indic[{ t_i = s_i }] \mid \F[t_i-1] ]}{2c} \left( \exp(hc) - \exp(-hc) \right) \enspace. \label{eq:exp-decomp-1}
\end{align}
By definition of $s_i$, when $t_i = s_i$, 
\begin{align}
\X[t_i] \neq \X[t_i-1] \qquad \text{and} \qquad \sum^{t_i - 1}_{t'=1} \indic[{ \X[t'] \neq \X[t'-1] }] = i - 1 \enspace,
\end{align}
which shows the necessary condition of $t_i = s_i$. Obviously, the above events are sufficient condition of $t_i = s_i$. Therefore, 
\begin{align}
\indic[{ t_i = s_i }] = \indic[{ \X[t_i] \neq \X[t_i-1] }] \indic[{ s_{i-1} \leq t_i - 1 < s_{i} }] \enspace.
\end{align}
\new{We note \cite[Theorem 4.1]{Wormald:1999} shows that $\tX$ is supermartingale.}
\rrev{The statement of~\cite[Theorem 4.1]{Wormald:1999} is found in Appendix~\ref{apdx:sec:existing}.}
Then, since $\indic[{ s_{i-1} \leq t_i - 1 < s_{i} }]$ is $\F[t_i-1]$-measurable, we have
\begin{align}
&\E[ (\tX[t_i] - \tX[t_i-1] )\indic[{ t_i = s_i }] \mid \F[t_i-1] ] \notag\\
&\quad = \E[ (\tX[t_i] - \tX[t_i-1] ) \indic[{ \X[t_i] \neq \X[t_i-1] }] \mid \F[t_i-1] ] \indic[{ s_{i-1} \leq t_i - 1 < s_{i} }] \\
&\quad = \E \left[ (\tX[t_i] - \tX[t_i-1] ) \left( \indic[{ \X[t_i] \neq \X[t_i-1] }] + \indic[{ \X[t_i] = \X[t_i-1] }] \right) \mid \F[t_i-1] \right] \notag \\
&\hspace{250pt} \cdot \indic[{ s_{i-1} \leq t_i - 1 < s_{i} }] \\
&\quad = \E[ \tX[t_i] - \tX[t_i-1] \mid \F[t_i-1] ] \indic[{ s_{i-1} \leq t_i - 1 < s_{i} }] \\
&\quad\leq 0 \enspace.
\end{align}
Then, from \eqref{eq:exp-decomp-1}, we have
\begin{align}
&\E\left[ \exp \left( h (\tX[t_i] - \tX[t_i-1]) \right) \indic[{ t_i = s_i }]\mid \F[t_i-1] \right] \notag \\
&\qquad\qquad\qquad\qquad \leq \frac{1}{2} \left( \exp(hc) + \exp(-hc) \right) \Pr( t_i = s_i \mid \F[t_i-1] ) \enspace.
\end{align}
Since it holds
\begin{align}
\frac{1}{2} \left( \exp(hc) + \exp(-hc) \right) = \cosh(h c) = \sum^\infty_{j=0} \frac{(hc)^{\rrrev{2}j}}{(2j)!} \leq \sum^\infty_{j=0} \frac{(hc)^{\rrrev{2}j}}{2^j j!} = \exp\left( \frac{1}{2} h^2 c^2 \right) \enspace,
\end{align}
we obtain
\begin{align}
& \E\left[ \exp \left( h (\tX[t_i] - \tX[t_i-1]) \right) \indic[{ t_i = s_i }] \mid \F[t_i-1] \right] \leq \exp\left( \frac{1}{2} h^2 c^2 \right) \Pr( t_i = s_i \mid \F[t_i-1] ) \enspace.
\end{align}
According to \eqref{eq:phi-exp-decomp}, it establishes
\begin{align}
&\E[ \phi(k)] \notag \\
&\leq \sum^{u}_{t_1=1} \sum^{u}_{t_2=t_1 + 1} \cdots \sum^{u}_{t_k = t_{k-1} + 1} \E \left[ \left( \prod^{k-1}_{i=1} \exp \left( h (\tX[t_i] - \tX[t_i-1]) \right) \indic[{ t_i = s_i }] \right) \right. \notag\\ 
&\qquad\qquad\qquad\qquad\qquad\qquad
\left. \exp\left( \frac{1}{2} h^2 c^2 \right) \Pr( t_k = s_k \mid \F[t_k-1] )\right] \\
&= \sum^{u}_{t_1=1} \sum^{u}_{t_2=t_1 + 1} \cdots \sum^{u}_{t_{k-1} = t_{k-2} + 1} \E \left[ \left( \prod^{k-1}_{i=1} \exp \left( h (\tX[t_i] - \tX[t_i-1]) \right) \indic[{ t_i = s_i }] \right) \right. \notag\\ 
&\qquad\qquad\qquad\qquad\qquad\qquad 
\left. \exp\left( \frac{1}{2} h^2 c^2 \right) \left( \sum^{u - 1}_{t_k = t_{k-1} + 1} \Pr( t_k = s_k \mid \F[t_k-1] )  \right) \right]  \enspace.
\end{align}
Since the events $i = s_k$ and $j = s_k$ are \del{mutually independent}{}\new{exclusive} for $i \neq j$, we obtain the following two inequalities as
\begin{align}
\sum^{u}_{t_k = t_{k-1} + 1} \Pr( t_k = s_k \mid \F[t_{k-1}] ) = \E\left[ \sum^{u }_{t_k = t_{k-1} + 1} \indic[{  t_k = s_k }] \mid \F[t_{k-1}] \right] \leq 1
\end{align}
and
\begin{align}
&\E[ \phi(k)] \notag \\
&\leq \underbrace{\sum^{u}_{t_1=1} \sum^{u }_{t_2=t_1 + 1} \cdots \sum^{u}_{t_{k-1} = t_{k-2} + 1}}_{\text{$k-1$ summations}} \notag \\
&\qquad\qquad\qquad\qquad\quad \E \left[ \left( \prod^{k-1}_{i=1} \exp \left( h (\tX[t_i] - \tX[t_i-1]) \right) \indic[{ t_i = s_i }] \right) \exp\left( \frac{1}{2} h^2 c^2 \right)  \right] \\
&= \new{ \sum^{u}_{t_1=1} \sum^{u }_{t_2=1} \cdots \sum^{u}_{t_{k-1} = 1} \E \left[ \left( \prod^{k-1}_{i=1} \exp \left( h (\tX[t_i] - \tX[t_i-1]) \right) \indic[{ t_i = s_i }] \right) \exp\left( \frac{1}{2} h^2 c^2 \right)  \right] } \\
&= \new{ \E \left[ \phi(k-1) \exp\left( \frac{1}{2} h^2 c^2 \right) \right] } \enspace.
\end{align}
Considering the conditional expectation w.r.t. $\F[t_i]$ \rrevdel{form}\rrev{from} $i=k$ to $i=1$ \rrevdel{by}\rrev{in} turn, we obtain
\begin{align}
\E \left[ \phi(k) \right] &\leq \exp \left( \frac{1}{2} k h^2 c^2 \right) \enspace.
\end{align}
According to \eqref{eq:phi-exp-sum}, we have
\begin{align}
\E \left[ \exp \left( h \tX[q_r] \right) \right] 
&\leq \sum^{r}_{k=1} \exp \left( \frac{1}{2} k h^2 c^2 \right) + 1 \leq (r+1) \exp \left( \frac{1}{2} r h^2 c^2 \right) \enspace.
\end{align}
Finally, \rrevdel{sitting}\rrev{setting} $h = m / (r c^2)$ shows
\begin{align}
\Pr( T < \min\{ s_r, u \} ) &\leq (r+1) \exp \left( - \frac{m^2}{r c^2} \right) \exp \left( \frac{m^2}{2 r c^2} \right) \\
&= (r+1) \exp \left( - \frac{m^2}{2 r c^2} \right)  \enspace.
\end{align}
\revdel{This is end of the proof.}{}\rev{This concludes the proof.}
\end{proof}

\rrevdel{\rev{The statement of~\cite[Theorem 4.1]{Wormald:1999} is found in Appendix~\ref{apdx:sec:existing}.}}
We note that the proof of Lemma~\ref{lem:skip-hoeffding} is similar way to the proof of \cite[Corollary 2.1]{Fan:2012}.
The \revdel{statement and }proof of Theorem~\ref{theo:negative-skip} \rrevdel{are provided follows:}\rrev{is provided as follows.}

\revdel{
\begin{theorem}[Theorem~4 in main manuscript] 
Consider a stochastic process $(\X, \F)_{t \in \mathbb{N}_0}$ with $\X[0] \leq 0$. Let $T$ be a stopping time defined as
\begin{align}
T = \min \left\{ t \in \mathbb{N} : \X \geq m \right\} \enspace.
\end{align}
Assume that, for all $t$, there are constants $0 < \varepsilon < m / 2$ and $0 < c < m$ satisfying
\begin{align}
\E[ \X[t+1] - \X \mid \F] &\leq - \varepsilon \Pr( \X[t+1] \neq \X \mid \F) \\
| \X[t+1] - \X | &\leq c \qquad \text{w.p. 1} \enspace.
\end{align}
Then, for all $n \geq 0$,
\begin{align}
\Pr( T \leq n ) \leq \frac{2m n}{\varepsilon} \exp \left( - \frac{m \varepsilon}{4 c^2} \right)  \enspace.
\end{align}
\end{theorem}
}

\begin{proof}[Proof of Theorem~\ref{theo:negative-skip}]
Consider stochastic processes $\Y_i$ with initial state $\Y[0]_i = \X[i]$, defined as
\begin{align}
\Y[s]_i = \X[s+i] + \varepsilon \sum^s_{j=1} \indic[{ \X[i+j] \neq \X[i+j-1] }] \enspace.
\end{align}
Note $\Y[s]_i$ moves if and only if $\X[s+i]$ moves.
Then
\begin{align}
&\E[ \Y[s+1]_i - \Y[s]_i \mid \F[s+i] ] \notag\\
&\qquad = \E[ \X[s+i+1] - \X[s+i] \mid \F[s+i]] + \varepsilon \Pr( \X[s+i+1] \neq \X[s+i] \mid \F[s+i]) \leq 0 \enspace.
\end{align}
Let $s_i$ be the iteration when $i$-th transition occurs, i.e., 
\begin{align}
s_i = \min \left\{ t \in \mathbb{N} : \sum^t_{t'=1} \indic[{ \X[t'] \neq \X[t'-1] }] \geq i \right\} \enspace.
\end{align}
When $\X[i] = \Y[0]_i \leq 0$, applying Lemma~\ref{lem:skip-hoeffding} \new{to the process $(\Y[s]_i, \F[s+i])$} shows
\begin{align}
\Pr( T_i \leq \min\{ n, s_{i,r} \} \mid \F[i]) \leq (r+1) \exp \left( - \frac{m^2}{2 r c^2} \right)  \enspace,
\end{align}
where
\begin{align}
r &= \lceil m / \varepsilon \rceil \label{eq:nega-def-r} \\
T_i &= \min\{ s \in \rrevdel{\mathbb{N}_0}\rrev{ \llbracket0,n \rrbracket } : \Y[s]_i \geq m \} \\
s_{i,r} &= \min \left\{ s \in \rrevdel{\mathbb{N}_0}\rrev{ \llbracket0,n \rrbracket } : \sum^s_{j=0} \indic[{ \Y[j]_i \neq \Y[j-1]_i }] \geq r \right\} \enspace.
\end{align}
Considering the condition $\varepsilon < m / 2$ and the definition of $r$ in \eqref{eq:nega-def-r}, we have $r + 1 \leq  2 m / \varepsilon$ and $- 1/r \leq - \varepsilon / (2 m)$. Then
\begin{align}
\Pr( T_i \leq \min\{ n, s_{i,r} \} \mid \F[i]) &\leq (r+1) \exp \left( - \frac{m^2}{2 r c^2} \right) \leq \frac{2m}{\varepsilon} \exp \left( - \frac{m \varepsilon}{4 c^2} \right)  \enspace.
\end{align}
\new{Further we derive an upper bound of the indicator of $T \leq n$ using the indicator of $T_i \leq \min\{ n, s_{i,r} \}$.}
\del{Here}{}\new{To simplify the notations}, we introduce $\psi_i$ and $\omega_i$ defined as
\begin{align}
\psi_i &:= \indic[{ \X[i]  \leq 0 }] \indic[{ T > i }] \indic[{ T_i \leq \min\{n, s_{i,r} \} }] \\
\omega_i &:= \indic[{ \X[i] \leq  0 }] \indic[{ T > i }] \indic[{ T \leq n}] \enspace.
\end{align}
We note $\omega_0 = \indic[{ T \leq n}]$ and $\omega_{i} = 0$ for $i \geq n-1$. Decomposing $\omega_i$ shows
\begin{align}
\omega_i &= \indic[{ \X[i] \leq  0 }] \indic[{ T > i }] \indic[{ T \leq n}] \left( \indic[{ T_i \leq \min\{n, s_{i,r} \} }] + \indic[{ T_i > \min\{n, s_{i,r} \} }]  \right) \\
&\leq \indic[{ \X[i]  \leq 0 }] \indic[{ T > i }] \indic[{ T_i \leq \min\{n, s_{i,r} \} }] \notag \\
&\qquad + \indic[{ \X[i] \leq  0 }] \indic[{ T > i }] \indic[{ T \leq n}]\indic[{ T_i > \min\{n, s_{i,r} \} }] \\
&= \indic[{ \X[i]  \leq 0 }] \indic[{ T > i }] \indic[{ T_i \leq \min\{n, s_{i,r} \} }] \notag \\
&\qquad + \indic[{ \X[i] \leq  0 }] \indic[{ T > i }] \indic[{ T \leq n}]\indic[{ T_i > \min\{n, s_{i,r} \} }] \left( \indic[{ n \leq s_{i,r}}] + \indic[{ n > s_{i,r}}]  \right)\\
&= \indic[{ \X[i]  \leq 0 }] \indic[{ T > i }] \indic[{ T_i \leq \min\{n, s_{i,r} \} }] \notag \\
&\qquad + \indic[{ \X[i] \leq  0 }] \indic[{ T > i }] \indic[{ T \leq n}] \cdot \indic[{ T_i > n  }] \indic[{ n \leq s_{i,r}}] \notag \\
&\qquad+ \indic[{ \X[i] \leq  0 }] \indic[{ T > i }] \indic[{ T \leq n}] \cdot \indic[{ T_i > s_{i,r} }] \indic[{ n > s_{i,r}}] \enspace. \label{apdx:proof:lemma-hoeff-indic-term-3}
\end{align}
The first term is given by $\psi_i$. The second term is $0$, since $T_i > n$ and $T > i$ means $T > n + i$, which is contradictory to $T \leq n$\revdel{.}\rev{, i.e.,
\begin{align}
    \omega_i = \psi_i + \indic[{ \X[i] \leq  0 }] \indic[{ T > i }] \indic[{ T \leq n}] \indic[{ T_i > s_{i,r} }] \indic[{ n > s_{i,r}}] \enspace.
    \label{apdx:proof:lemma-hoeff-indic-term-3-3}
\end{align}
} 
Moreover, when $T_i \geq s_{i,r}$\rrev{, which holds $\Y[s_{i,r}]_i < m$}, we observe $\X[s_{i,r} + i]$ becomes less than zero since
\begin{align}
\Y[s_{i,r}]_i = \X[s_{i,r} + i] + \varepsilon r \geq \X[s_{i,r} + i] + m \enspace,
\end{align}
indicating
\begin{align}
\indic[{ \X[i] \leq  0 }]\indic[{ T_i > j }]  \indic[{ s_{i,r} = j}] = \indic[{ \X[i] \leq  0 }]\indic[{ T_i > j }]\indic[{\X[i+j] \leq 0}]  \indic[{ s_{i,r} = j}] \enspace.
\end{align}
In addition, since $\Y[s]_i \geq \X[s+i]$,
\begin{align}
\indic[{ T_i > j}] &= \indic[{ T_i > j}] \left( \indic[{ T \geq i}] + \indic[{ T < i}] \right) \\
&\leq \indic[{ T_i > j}] \indic[{ T \geq i}] + \indic[{ T < i}] \\
&\leq \indic[{ T > i + j}] + \indic[{ T < i}] \enspace.
\end{align}
Therefore, we obtain an upper bound of the \revdel{third term in \eqref{apdx:proof:lemma-hoeff-indic-term-3}}\rev{second term in \eqref{apdx:proof:lemma-hoeff-indic-term-3-3}} as
\begin{align}
&\indic[{ \X[i] \leq  0 }] \indic[{ T > i }] \indic[{ T \leq n}] \sum_{j=1}^{n-1} \indic[{ T_i > j }] \indic[{ s_{i,r} = j}] \notag \\
&= \indic[{ \X[i] \leq  0 }] \indic[{ T > i }] \indic[{ T \leq n}] \sum_{j=1}^{n-1} \indic[{ \X[i + j] \leq  0 }] \indic[{ T_i > j }] \indic[{ s_{i,r} = j}] \\
&\leq \indic[{ \X[i] \leq  0 }] \indic[{ T > i }] \indic[{ T \leq n}] \sum_{j=1}^{n-1} \indic[{ \X[i + j] \leq  0 }] ( \indic[{ T> i + j }] + \indic[{ T < i }]) \indic[{ s_{i,r} = j}] \\
&= \indic[{ \X[i] \leq  0 }] \indic[{ T > i }] \indic[{ T \leq n}] \sum_{j=1}^{n-1} \indic[{ \X[i + j] \leq  0 }] \indic[{ T> i + j }] \indic[{ s_{i,r} = j}] \\
&\leq \indic[{ T \leq n}] \sum_{j=1}^{n-1} \indic[{ \X[i + j] \leq  0 }] \indic[{ T> i + j }] \indic[{ s_{i,r} = j}] \\
&= \sum_{j=1}^{n-1} \omega_{i+j} \indic[{ s_{i,r} = j}]
\end{align}
Totally,
\begin{align}
\omega_i \leq \psi_i + \sum_{j=1}^{n-1} \omega_{i+j} \indic[{ s_{i,r} = j}] \enspace.
\end{align}
Remaining $\omega_0 = \indic[{ T \leq n}]$ and $\omega_{n-1} = 0$, we have
\begin{align}
\indic[{ T \leq n}] &\leq \psi_0 + \sum_{j_1=1}^{n-1} \omega_{j_1} \indic[{ s_{0,r} = j_1}] \\
&\leq \psi_0 + \sum_{j_1=1}^{n-1} \psi_{j_1} \indic[{ s_{0,r} = j_1}] + \sum_{j_1=1}^{n-1} \sum_{j_2=1}^{n-1} \omega_{{j_1 + j_2}} \indic[{ s_{j_1,r} = j_2}] \indic[{ s_{0,r} = j_1}] \\
&\leq \psi_0 + \sum_{j_1=1}^{n-1} \psi_{j_1} \indic[{ s_{0,r} = j_1}] + \sum_{j_1=1}^{n-1} \sum_{j_2=1}^{n-1} \psi_{{j_1 + j_2}} \indic[{ s_{j_1,r} = j_2}] \indic[{ s_{0,r} = j_1}] + \cdots  \notag \\
&\qquad + \underbrace{ \sum_{j_1=1}^{n-1} \cdots \sum_{j_{l}=1}^{n-1} }_{\text{$l$ summations}}\psi_{\sum^{l}_{k=1} j_k} \prod^{l}_{k=1} \indic[{ s_{j_{k-1},r} = j_k}] + \cdots \notag \\
&\qquad + \underbrace{ \sum_{j_1=1}^{n-1} \cdots \sum_{j_{n-1}=1}^{n-1} }_{\text{$n-1$ summations}} \psi_{\sum^{n-1}_{k=1} j_k} \prod^{n-1}_{k=1} \indic[{ s_{j_{k-1},r} = j_k}] \enspace.
\end{align}
Since $\prod^{l}_{k=1} \indic[{ s_{j_{k-1},r} = j_k}]$ is $\F[\sum^{l}_{k=1} j_k]$ measurable,
\begin{align}
&\E \left[ \sum_{j_1=1}^{n-1} \cdots \sum_{j_{l}=1}^{n-1} \psi_{\sum^{l}_{k=1} j_k} \prod^{l}_{k=1} \indic[{ s_{j_{k-1},r} = j_k}] \right] \\
&= \E \left[ \sum_{j_1=1}^{n-1} \cdots \sum_{j_{l}=1}^{n-1} \E \left[ \psi_{\sum^{l}_{k=1} j_k} \mid \F[\sum^{l}_{k=1} j_k] \right] \prod^{l}_{k=1} \indic[{ s_{j_{k-1},r} = j_k}] \right] \\
&\leq\E \left[ \sum_{j_1=1}^{n-1} \cdots \sum_{j_{l}=1}^{n-1} \E \left[ \indic[{ T_{\sum^{l}_{k=1} j_k} \leq \min\{n, s_{{\sum^{l}_{k=1} j_k},r} \} }] \mid \F[\sum^{l}_{k=1} j_k] \right] \prod^{l}_{k=1} \indic[{ s_{j_{k-1},r} = j_k}] \right] \\
&\leq \frac{2m}{\varepsilon} \exp \left( - \frac{m \varepsilon}{4 c^2} \right) \E \left[ \sum_{j_1=1}^{n-1} \cdots \sum_{j_{l}=1}^{n-1}  \prod^{l}_{k=1} \indic[{ s_{j_{k-1},r} = j_k}] \right] \\
&= \frac{2m}{\varepsilon} \exp \left( - \frac{m \varepsilon}{4 c^2} \right) \E \left[ \sum_{j_1=1}^{n-1} \cdots \sum_{j_{l-1}=1}^{n-1} \prod^{l - 1}_{k=1} \indic[{ s_{j_{k-1},r} = j_k}] \underbrace{ \left( \sum_{j_{l}=1}^{n-1} \indic[{ s_{j_{l-1},r} = j_l}]\right) }_{ = \indic[{s_{j_{l-1},r} \in \intrange{1}{n-1}}] \leq 1} \right] \\
&\leq \frac{2m}{\varepsilon} \exp \left( - \frac{m \varepsilon}{4 c^2} \right) \enspace,
\end{align}
where the last inequality is obtained by bounding the summations from above by $1$ from the last summation by turn.
Since 
\begin{align}
\E[ \psi_0 ] \leq \frac{2m}{\varepsilon} \exp \left( - \frac{m \varepsilon}{4 c^2} \right) \enspace,
\end{align}
we have
\begin{align}
\Pr(T \leq n) \leq \frac{2mn}{\varepsilon} \exp \left( - \frac{m \varepsilon}{4 c^2} \right) \enspace.
\end{align}
\revdel{This is end of the proof.}{}\rev{This concludes the proof.}
\end{proof}

\bibliographystyle{apalike}
\bibliography{reference}



\end{document}